\documentclass{article}

 \PassOptionsToPackage{numbers, compress}{natbib}

\usepackage[final]{neurips_2025}

\usepackage[utf8]{inputenc} 
\usepackage[T1]{fontenc}    
\usepackage{hyperref}       
\usepackage{url}            
\usepackage{booktabs}       
\usepackage{amsfonts}       
\usepackage{nicefrac}       
\usepackage{microtype}      
\usepackage{xcolor}         

\usepackage{amsmath}
\usepackage{amssymb}
\usepackage{mathtools}
\usepackage{amsthm}

\usepackage[capitalize,noabbrev]{cleveref}

\theoremstyle{plain}
\newtheorem{theorem}{Theorem}[section]
\newtheorem{proposition}[theorem]{Proposition}
\newtheorem{lemma}[theorem]{Lemma}
\newtheorem{corollary}[theorem]{Corollary}
\theoremstyle{definition}
\newtheorem{definition}[theorem]{Definition}

\theoremstyle{remark}
\newtheorem{remark}[theorem]{Remark}

\usepackage{subcaption}
\usepackage{epsf}
\usepackage{fancyhdr}
\usepackage{graphics}
\usepackage{psfrag}

\usepackage{wrapfig}
\usepackage{float}

\usepackage{algorithm}
\usepackage{algpseudocode}

\usepackage[algo2e,linesnumbered,ruled,vlined]{algorithm2e}

\usepackage{color}

\usepackage{multirow}
\usepackage{balance}

\def\calGS{\mathcal{GS}}

\def\calOS{\mathcal{OS}}

\def\calOE{\mathcal{OE}}

\def\calW{\mathcal{W}}

\def\supp{\mathrm{supp}}

\def\OrliczSobolevPhi{W\!L_{\Phi}}
\def\OrliczSobolevPsi{W\!L_{\Psi}}

\usepackage{verbatim}

\usepackage{xr-hyper}
\usepackage{hyperref}

\makeatletter
\def\munderbar#1{\underline{\sbox\tw@{$#1$}\dp\tw@\z@\box\tw@}}
\makeatother

\newcommand{\be}{\begin{equation}}
\newcommand{\ee}{\end{equation}}
\newcommand{\bea}{\begin{equation*}\begin{aligned}}
\newcommand{\eea}{\end{aligned}\end{equation*}}

\newcommand{\R}{\mathbb{R}}

\newcommand{\G}{{\mathbb G}}

\newcommand{\calP}{{\mathcal P}}

\newcommand{\calS}{{\mathcal S}}

\newcommand{\calT}{{\mathcal T}}
\newcommand{\calI}{{\mathcal I}}

\newcommand{\calF}{{\mathcal F}}

\newcommand{\dd}{\mathrm{d}}

\newcommand{\norm}[1]{\left\lVert#1\right\rVert}

\title{An Efficient Orlicz-Sobolev Approach for Transporting Unbalanced Measures on a Graph}

\author{%
  Tam Le\thanks{: equal contribution. Correspondence to: Tam Le <\texttt{tam@ism.ac.jp}>.} \\
  Institute of Statistical Mathematics \\
  \texttt{tam@ism.ac.jp} \\
  \And
  Truyen Nguyen$^*$ \\
  University of Akron \\
  \texttt{truyennguyen3@gmail.com} \\
  \AND
  Hideitsu Hino \\
  Institute of Statistical Mathematics \\
  \texttt{hino@ism.ac.jp} \\
  \And
  Kenji Fukumizu \\
  Institute of Statistical Mathematics \\
  \texttt{fukumizu@ism.ac.jp} \\
}

\begin{document}

\maketitle

\begin{abstract}

We investigate optimal transport (OT) for measures on graph metric spaces with different total masses. To mitigate the limitations of traditional $L^p$ geometry, Orlicz-Wasserstein (OW) and generalized Sobolev transport (GST) employ \emph{Orlicz geometric structure}, leveraging convex functions to capture nuanced geometric relationships and remarkably contribute to advance certain machine learning approaches. However, both OW and GST are restricted to measures with equal total mass, limiting their applicability to real-world scenarios where mass variation is common, and input measures may have noisy supports, or outliers. To address unbalanced measures, OW can either incorporate mass constraints or marginal discrepancy penalization, but this leads to a more complex two-level optimization problem. Additionally, GST provides a scalable yet rigid framework, which poses significant challenges to extend GST to accommodate nonnegative measures. To tackle these challenges, in this work we revisit the entropy partial transport (EPT) problem. By exploiting~\citet{CM}'s insights, we develop a novel variant of EPT endowed with Orlicz geometric structure, called \emph{Orlicz-EPT}. We establish theoretical background to solve Orlicz-EPT using a binary search algorithmic approach. Especially, by leveraging  the dual EPT and the underlying graph structure, we formulate a novel regularization approach that leads to the proposed \emph{Orlicz-Sobolev transport} (OST). Notably, we demonstrate that OST can be efficiently computed by simply solving a univariate optimization problem, in stark contrast to the intensive computation needed for Orlicz-EPT. Building on this, we derive geometric structures for OST and draw its connections to other transport distances. We empirically illustrate that OST is several-order faster than Orlicz-EPT. Furthermore, we show initial evidence on the advantages of OST for measures on a graph in document classification and topological data analysis.

\end{abstract}


\section{Introduction}
\label{sec:introduction}

Orlicz-Wasserstein (OW) extends $L^p$
geometry by leveraging a specific class of convex functions for Orlicz geometric structure.
Intuitively, OW is an instance of optimal transport (OT), which utilizes Orlicz metric as its ground cost~\citep{sturm2011generalized, kell2017interpolation, GuhaHN23, altschuler2024faster, le2024generalized}. 
Building on this foundation, OW has proven instrumental in advancing certain machine learning approaches.
For example, recent works have leveraged OW to tackle challenging problems: \citet{altschuler2024faster} use OW as a metric shift for R\'enyi divergence, enabling novel differential-privacy-inspired techniques to overcome longstanding challenges for fast convergence of hypocoercive differential equations, while \citet{GuhaHN23} employ OW metric to significantly improve Bayesian contraction rates in hierarchical Bayesian nonparametric models by overcoming limitations raised from the usage of traditional OT with Euclidean ground cost.
However, OW's high computational complexity, stemming from its two-level optimization formula,
poses a significant limitation. To address this challenge, \citet{le2024generalized} introduce generalized Sobolev transport (GST), a scalable variant of OW suitable for practical application domains, especially for large-scale settings. Moreover, Orlicz geometric structure has been successfully applied to various machine learning problems, including linear regression~\citep{andoni2018subspace, song2019efficient}, scalable approaches~\citep{deng2022fast} for reinforcement learning, kernelized support vector machines, and clustering. 
Additionally, Orlicz metrics play a crucial role in deriving deviation bounds for polynomial-growth functions to approximate kernel derivatives~\citep{chamakh2020orlicz}, and have been used as regularization in OT problems~\citep{lorenz2022orlicz}. For in-depth studies on Orlicz functions, see~\citep{adams2003sobolev, rao1991theory}.

When dealing with input measures having different total masses, various approaches have been proposed in the literature to address this challenge~\citep{hanin1992kantorovich, guittet2002extended, benamou2003numerical, CM, figalli2010optimal, lellmann2014imaging, P1, P2, frogner2015learning, kondratyev2016new, Liero2018, chizat2018unbalanced, bonneel2019spot, gangbo2019unnormalized, sejourne2019sinkhorn, pmlr-v151-sejourne22a, pham2020unbalanced, sato2020fast, chapel2020partial, balaji2020robust, mukherjee2021outlier, le2021ept, fatras2021unbalanced, chapel2021unbalanced, sejourne2023unbalanced, le2023scalable, nguyen2023unbalanced, bonet2024slicing, chapel2025one, tran2025treeslicedEPT}. These approaches for unbalanced measures have proven effective in various domains, including color transfer~\citep{bonneel2019spot}, shape matching~\citep{bonneel2019spot}, image-to-image translation~\citep{tran2025treeslicedEPT}, multi-label learning~\citep{frogner2015learning}, positive-unlabeled learning~\citep{chapel2020partial}, point-cloud gradient flow~\citep{tran2025treeslicedEPT}, natural language processing~\citep{le2021ept, le2023scalable}, topological data analysis (TDA)~\citep{le2021ept, le2023scalable}, generative modeling~\citep{balaji2020robust, tran2025treeslicedEPT}, domain adaptation~\citep{balaji2020robust}, and robust approaches for handling noisy supports, outliers~\citep{frogner2015learning, balaji2020robust, mukherjee2021outlier}, or noisy ground cost~\citep{pmlr-v97-paty19a, le2024optimal}.


In this work, we focus on the OT problem with Orlicz geometric structure for unbalanced measures supported on a graph metric space. On one hand, OW naturally extends OT's flexibility to handle unbalanced measures by incorporating either mass constraints or marginal difference penalization, formulated as partial OT (POT) or unbalanced OT (UOT) respectively. However, these approaches result in a more complex two-level optimization problem, analogous to POT/UOT with Orlicz metric cost, which poses significant computational challenges. On the other hand, although GST provides a scalable alternative to the computationally intensive OW, it still assumes equal-mass input measures. Moreover, due to GST's definition as an optimization over the critic function, extending it to unbalanced measures is nontrivial. To address these limitations, we revisit the entropy partial transport (EPT) problem~\citep{le2021ept, le2023scalable, tran2025treeslicedEPT} and leverage insights from~\citet{CM} to reformulate EPT as a standard complete OT problem. Then by carefully calibrating the corresponding ground cost for its nonnegativity, we propose Orlicz-EPT and establish a theoretical foundation 
for solving it by a binary search algorithmic approach. Furthermore, by exploiting the dual EPT and underlying graph structure, we introduce a novel regularization approach, leading to Orlicz-Sobolev transport (OST), which scales Orlicz-EPT for practical applications.

\paragraph{Contribution.} In summary, our contributions are two-fold as follows:

\begin{itemize}

\item We revisit the EPT problem and  leverage~\citet{CM}'s insights to reformulate EPT as a standard complete OT, leading to the development of the proposed Orlicz-EPT. We establish its theoretical foundations, enabling a binary search algorithmic approach for its computation. Additionally, we develop a novel regularization approach, resulting in the proposed OST. We show that OST can be efficiently computed by simply solving a univariate optimization problem, unlike the computationally intensive Orlicz-EPT.

\item We derive geometric structures for OST and establish its connections to other transport distances. Our empirical results demonstrate that OST is several-order faster than Orlicz-EPT. We also provide initial evidence on the advantages of OST for document classification and TDA.


\end{itemize}

\paragraph{Organization.} In~\S\ref{sec:preliminaries}, we briefly review relevant background and notions.
We revisit EPT problem and propose Orlicz-EPT in~\S\ref{sec:OrliczEPT}. In \S\ref{sec:OST}, we introduce the computationally efficient OST. 
Then we derive geometric structures for OST and draw its connections to other transport distances in~\S\ref{sec:properties_OrliczSobolevTransport}. In 
\S\ref{sec:related_works}, we discuss related work. Empirical results are presented in~\S\ref{sec:experiments}, followed by concluding remarks in~\S\ref{sec:conclusion}. Proofs of key theoretical results and additional materials are deferred to the Appendices. Furthermore, we have released code for our proposed approaches.\footnote{The code repository is on \url{https://github.com/lttam/OST_OrliczEPT}.}


\section{Preliminaries}
\label{sec:preliminaries}

In this section, we introduce notations, and briefly review graph, and Orlicz functions.

\textbf{Graph.} We follow the graph setting as in~\citep{le2022st}. Let $V, E$ be the sets of nodes and edges respectively. We consider a connected, undirected, and physical\footnote{In the sense that $V$ is a subset of Euclidean space $\R^n$, and each edge $e \in E$ is the standard line segment in $\R^n$ connecting the two vertices  of the edge $e$.} graph $\G = (V,E)$ with positive edge lengths $\{w_e\}_{e\in E}$. For continuous graph setting, we regard $\G$ as the set of all nodes in $V$ and all points forming the edges in $E$. We equip $\G$ with graph metric $d_{\G}(x,y)$  which equals to the length of the shortest path between $x$ and $y$ in $\G$. Additionally, we assume that there exists a fixed root node $z_0 \in V$ such that the shortest path connecting $z_0$ and $x$ is unique for any $x \in \G$, i.e., the uniqueness property of the shortest paths. We denote $\calP(\G)$ (resp.$\,\calP(\G \times\G)$) as the set of all nonnegative Borel measures on $\G$ (resp.$\,\G\times\G$) with a finite mass. Let $[x, z]$ be the shortest path connecting $x$ and $z$ in $\G$. For $x \in \G$, edge $e \in E$, define the sets $\Lambda(x)$ and $\gamma_e$ as follows: 
\begin{eqnarray}\label{sub-graph}
 \Lambda(x) := \big\{y\in \G: \, x\in [z_0,y]\big\}, \qquad \gamma_e := \big\{y\in \G: \, e\subset  [z_0,y]\big\}.
\end{eqnarray}


\textbf{Functions on graph.} By a continuous function $f$ on $\G$, we mean that  $f: \G\to \R$ is continuous w.r.t.~the topology on $\G$ induced by the Euclidean distance. Henceforth, $C(\G)$ denotes the set of all continuous functions on $\G$.
Similar notation is used for continuous functions on $\G \times \G$. Given a positive scalar $b>0$, then a function $f:\G\to\R$ is called $b$-Lipschitz w.r.t.~$d_\G$ if $|f(x) - f(y)|\leq b \, d_\G(x,y)$ for every $ x$ and $y$ in $\G$.

\textbf{A family of convex functions.} We consider the collection  of  $N$-functions~\citep[\S8.2]{adams2003sobolev} which are special convex functions on $\R_+$. Hereafter, a strictly increasing and   convex function $\Phi: [0, \infty)\to [0, \infty)$ is called an $N$-function if  $\lim_{t \to 0} \frac{\Phi(t)}{t} = 0$ and $\lim_{t \to +\infty} \frac{\Phi(t)}{t} = +\infty$.

\textbf{Orlicz functional space.} Given an  $N$-function $\Phi$ and  
a nonnegative Borel measure $\omega$ on $\G$, let $L_{\Phi}(\G, \omega)$ be the linear hull of the set of all Borel measurable functions $f: \G \to \R$ satisfying $\int_{\G} \Phi(|f(x)|) \omega(\text{d}x) < \infty$. Then, $L_{\Phi}(\G, \omega)$ is a normed space with the Luxemburg norm, defined as 
\begin{equation}\label{eq:Luxemburg_norm}
\norm{f}_{L_\Phi}  \coloneqq \inf \left\{t > 0 \mid \int_{\G} \Phi\!\left(\frac{|f(x)|}{t}\right)\omega(\text{d}x) \le 1 \right\}.
\end{equation}


\section{Orlicz-EPT: Entropy Partial Transport with Orlicz Geometric Structure}
\label{sec:OrliczEPT}

In this section, we revisit the entropy partial transport (EPT) problem~\citep{le2021ept,le2023scalable, tran2025treeslicedEPT}, then develop Orlicz-EPT as a variant of EPT endowed with Orlicz geometric structure.

\subsection{Entropy Partial Transport (EPT)}

Let $\gamma_1, \gamma_2$ be the first and second marginals of $\gamma \in \calP(\G \times \G)$ respectively. For unbalanced measures $\mu, \nu \in \calP(\G)$, we consider the set $\Pi_{\leq}(\mu,\nu) := \left\{ \gamma :  \, \gamma_1\leq \mu, \, \gamma_2\leq \nu \right\}$.\footnote{$\gamma_1 \le \mu$ means that $\gamma_1(B) \le \mu(B)$ for all Borel set $B \subset \G$, similarly for $\gamma_2 \le \nu$.} Additionally, let $f_1, f_2$ be the Radon-Nikodym derivatives of $\gamma_1$ w.r.t. $\mu$ and of $\gamma_2$ w.r.t. $\nu$ respectively, i.e., $\gamma_1=f_1 \mu$ ($0\leq f_1 \leq 1$, $\mu$-a.e.) and $\gamma_2 = f_2 \nu$ ($0\leq f_2 \leq 1$, $\nu$-a.e.). 

For convex and lower semicontinuous entropy functions $F_1, \, F_2: [0,1]\to (0,\infty)$, and nonnegative weight functions $w_1, w_2:\G \to [0,\infty)$, we consider the weighted relative entropies $\calF_1(\gamma_1| \mu) := \int_\G w_1(x) F_1(f_1(x) ) \mu(\dd x)$, and $\calF_2(\gamma_2| \nu) := \int_\G w_2(x) F_2(f_2(x) ) \nu(\dd x)$. For scalar $b > 0$, scalar $m\in [0,\bar m]$ with $\bar m := \min\{\mu(\G), \nu(\G) \}$, and graph metric $d_{\G}$ as ground cost, the EPT problem is
\begin{eqnarray}\label{original}
\mathrm{W}_{m}(\mu,\nu) 
 := \inf_{\gamma \in \Pi_{\leq}(\mu,\nu), \, \gamma(\G\times \G)=m} \Big[ \calF_1(\gamma_1| \mu )  + \calF_2(\gamma_2| \nu )  + \, b \int_{\G \times \G} d_{\G}(x,y) \gamma(\dd x, \dd y) \Big].
\end{eqnarray}

Following~\citep[\S3]{le2023scalable}, by using entropy functions $F_1(s)=F_2(s) :=|s-1|$ and considering a Lagrange multiplier $\lambda\in\R$ conjugate to the constraint $\gamma(\G\times \G)=m$, we instead study the problem
\begin{eqnarray}\label{P1}
\mathrm{ET}_{\lambda}(\mu,\nu) 
= \inf_{\gamma \in \Pi_{\leq}(\mu,\nu)} \mathcal{C}_\lambda(\gamma),
\end{eqnarray}
where $\mathcal{C}_\lambda(\gamma) = \int_\G w_1 \mu(\dd x) + \int_\G  w_2  \nu(\dd x) -\int_\G  w_1 \gamma_1(\dd x) - \int_\G  w_2\gamma_2(\dd x)+ b  \int_{\G \times \G} [d_{\G}(x,y)-\lambda]\gamma(\dd x, \dd y)$.\footnote{The relationship between Problem~\eqref{original} and Problem~\eqref{P1} is established in~\citep[Theorem A.1]{le2023scalable}.} 


\textbf{EPT as a standard OT.} Following \citet{CM}'s insights, we can reformulate problem~\eqref{P1} as the standard complete OT problem. However, it is not guarantee that the corresponding standard OT has a nonnegative ground cost, e.g., see~\citep{CM, le2021ept, le2023scalable, tran2025treeslicedEPT}. Therefore, such OT reformulation may not be applicable to derive corresponding OW as in~\citep{sturm2011generalized, kell2017interpolation, GuhaHN23, altschuler2024faster, le2024generalized} since $N$-function is only defined for nonnegative domain (\S\ref{sec:preliminaries}). Therefore, it is essential to carefully calibrate the ground cost of the corresponding standard OT of EPT to ensure its nonnegativity. 

Precisely, following~\citep[Theorem 3.1]{le2023scalable},  we henceforth
 consider $\lambda \ge 0$.\footnote{The dual EPT result is the foundation for developing Orlicz-Sobolev transport in \S\ref{sec:OST}, where $\lambda$ is nonnegative.}
Then let $\hat s$ be a point outside graph $\G$, i.e., $\hat s \notin \G$, and extend graph metric cost $d_{\G}$ on $\G$ to a new \emph{nonnegative} cost function $\hat c$ with $b\lambda$-deviation on $\hat\G:= \G \cup \{\hat s\}$ as follows:
\begin{equation}\label{eq:c_hat_cost}
\hat c(x,y) :=
\left\{\begin{array}{lr}
b \,d_{\G}(x,y) \hspace{2.7 em} \mbox{ if } x,y\in \G,\\
w_1(x) + b\lambda \hspace{2.0 em} \mbox{ if }  x\in \G \mbox{ and } y=\hat s,\\
 w_2(y) + b\lambda \hspace{2.0 em}  \mbox{ if }  x=\hat s \mbox{ and } y\in \G,\\
  b\lambda \hspace{5.7 em} \mbox{ if }  x=y=\hat s.
\end{array}\right.
\end{equation}

For unbalanced measures $\mu, \nu$, we construct corresponding probability (balanced) measures $\hat\mu = \frac{\mu +\nu(\G) \delta_{\hat s}}{\mu(\G) + \nu(\G)}$ and $\hat\nu = \frac{\nu +\mu(\G) \delta_{\hat s}}{\mu(\G) + \nu(\G)}$. Let $\Pi(\hat \mu,\hat \nu) := \Big\{ \hat\gamma \in \calP( \hat\G \times \hat \G): \hat \mu(U) =\hat\gamma(U\times \hat \G),\, \hat\nu(U)= \hat\gamma(\hat \G\times U) \mbox{ for all Borel sets } U\subset \hat \G\Big\}$, then one can recast EPT~\eqref{P1} as a standard OT with cost $\hat c$.
\begin{proposition}\label{prop:ET_KT}
    Consider the standard OT $\calW_{\hat c}$ between probability measures $\hat\mu, \hat\nu$ with cost $\hat{c}$,
    \begin{equation}
    \calW_{\hat c}(\hat \mu,\hat \nu) := \inf_{\hat \gamma \in \Pi(\hat\mu,\hat \nu)}  \int_{\hat \G\times \hat \G} \hat c(x,y) \hat\gamma(\dd x, \dd y),
    \end{equation}
    then we have
    \begin{eqnarray}\label{P2}
        \mathrm{KT}(\mu, \nu) := \left(\mu(\G) + \nu(\G)\right)\left(\calW_{\hat c}(\hat \mu,\hat \nu) - b\lambda\right) = \mathrm{ET}_{\lambda}(\mu, \nu).
    \end{eqnarray}
\end{proposition}
The proof is placed in Appendix \S\ref{appsubsec:ET_KT}.


Thus, we have reformulated EPT~\eqref{P1} for unbalanced measures as a corresponding standard complete OT~\eqref{P2} with \emph{nonnegative} ground cost. Consequently, we bypass the technical challenges inherent in unbalanced settings and can leverage abundant existing results and approaches in the standard balanced setting for OT problems with unbalanced measures on a graph.

\begin{remark}[Nonnegativity]
Unlike existing approaches, e.g., as in~\citep{CM, le2021ept, le2023scalable, tran2025treeslicedEPT}, the new ground cost $\hat{c}$ of the corresponding standard OT problem~\eqref{P2} for EPT is \emph{guaranteed} to be nonnegative. Our calibration is essential for developing the associated OW from its standard OT as in~\citep{sturm2011generalized, kell2017interpolation, GuhaHN23, chewi2023optimization}.
\end{remark}

\subsection{Orlicz-EPT} 

Following the approaches in~\citep{sturm2011generalized, kell2017interpolation, GuhaHN23, chewi2023optimization}, we define \emph{Orlicz-EPT}, which is EPT endowed with an Orlicz geometric structure, based on the standard OT problem~\eqref{P2} as follows:
\begin{eqnarray}\label{eq:OrliczEPT}
\calOE_{\Phi}(\mu, \nu) := \left(\mu(\G) + \nu(\G)\right)(\calW_{\Phi}(\hat \mu,\hat \nu) - b\lambda),
\end{eqnarray}
where $\calW_{\Phi}(\hat \mu,\hat \nu) := \inf_{\tilde \gamma \in \Pi(\hat \mu, \hat \nu)} \inf \Big[ t > 0 : \int_{\hat \G \times \hat \G} \Phi\left(\frac{\hat{c}(x, y)}{t}\right) \text{d}\tilde\gamma(x, y) \le 1\Big]$.

It should be noted that, similar to OW, Orlicz-EPT~\eqref{eq:OrliczEPT} is derived from the standard OT problem~\eqref{P2}, 
which circumvents all challenges coming from the setting of unbalanced measures. 

We next show that the objective function of Orlicz-EPT is monotone non-increasing w.r.t. $t$.
\begin{proposition}[Monotonicity]\label{prop:monotonicity_OT}
Let $\Phi$ be an $N$-function, and let $\hat c$ be the cost given by
\eqref{eq:c_hat_cost}. 
For any probability measures $\hat \mu$ and $\hat \nu$ on $\hat\G$, define
\begin{equation}\label{eq:OW_t}
\mathcal{A}(t; \hat \mu, \hat \nu) := \inf_{\tilde \gamma \in \Pi(\hat \mu, \hat \nu)} \int_{\hat \G \times \hat \G} \Phi\!\left(\frac{\hat{c}(x, y)}{t}\right) \dd\tilde\gamma(x, y)\quad \mbox{for} \quad t > 0.
\end{equation}
Then the function $t\in (0, +\infty) \longmapsto \mathcal{A}(t; \hat \mu, \hat \nu)$ is monotone non-increasing.
\end{proposition}

The proof is placed in Appendix \S\ref{appsubsec:Monotonicity_OT}. 

\paragraph{Computation.} Observe that for a fixed $t$, $\mathcal{A}$ is a standard OT problem between $\hat\mu$ and $\hat\nu$ with the cost function $\Phi\!\left(\frac{\hat{c}(\cdot, \cdot)}{t}\right)$. For computational efficiency,\footnote{Entropic regularized OT reduces the computational cost of OT from super-cubic to quadratic complexity~\citep{Cuturi-2013-Sinkhorn}.} we consider its corresponding entropic regularization~\citep{Cuturi-2013-Sinkhorn}, and show that the monotonicity is preserved.
\begin{proposition}[Entropic regularization]\label{prop:monotonicity_regOT}
Define the entropic regularization of $\mathcal{A}$ as   
\begin{equation}\label{eq:regOW_t}
\mathcal{A}_{\varepsilon}(t; \hat \mu, \hat \nu) := \inf_{\tilde \gamma \in \Pi(\hat \mu, \hat \nu)} \left[ \int_{\hat \G \times \hat \G} \Phi\!\left( \frac{\hat{c}(x, y)}{t} \right) \dd\tilde\gamma(x, y) - \varepsilon H(\tilde\gamma) \right],
\end{equation}
where $\varepsilon \ge 0$ and $H$ is Shannon entropy defined by $H(\tilde\gamma) := -\int_{\hat\G \times \hat \G} (\log\tilde\gamma(x, y) - 1)\dd\tilde\gamma(x, y)$. Then the function $t\in (0, +\infty) \longmapsto \mathcal{A}_{\varepsilon}(t; \hat \mu, \hat \nu)$ is monotone non-increasing.
\end{proposition}


The proof is placed in Appendix \S\ref{appsubsec:Monotonicity_regOT}.

In addition, we obtain  the following upper and lower bounds for $\mathcal{A}_{\varepsilon}$.
\begin{proposition}[Bounds]\label{prop:limits_regOT}
Let $\supp(\cdot)$ be a set of supports of a measure, then we have
\[
\mathcal{A}_{\varepsilon} \! \left(\frac{\calW_{\hat c}(\hat \mu, \hat \nu)}{\Phi^{-1}(1 + \varepsilon\left[ H(\hat \mu) + H(\hat \nu) - 1 \right])}; \hat \mu, \hat \nu  \right) \ge  1, \quad \text{and} \quad \mathcal{A}_{\varepsilon} \! \left(\frac{L_{\hat \mu, \hat \nu}}{\Phi^{-1}(1 + \varepsilon)}; \hat \mu, \hat \nu\right) \le 1,
\]
where $L_{\hat \mu, \hat \nu} := \max_{x \in \supp(\hat \mu), y \in \supp(\hat \nu)} \hat{c}(x, y)$.
\end{proposition}

The proof is placed in Appendix \S\ref{appsubsec:limits_regOT}. 



Thanks to the monotonicity of $\mathcal{A}_{\varepsilon}$ in Proposition~\ref{prop:monotonicity_regOT} and  the limits of $\mathcal{A}_{\varepsilon}$ in Proposition~\ref{prop:limits_regOT}, we can leverage the binary search approach to compute the entropic regularized Orlicz-EPT, which corresponds to the original Orlicz-EPT~\eqref{eq:OrliczEPT}. Precisely, this entropic regularization is defined as
\begin{eqnarray}\label{eq:regOrliczEPT}
\calOE_{\Phi, \varepsilon}(\mu, \nu) := \left(\mu(\G) + \nu(\G)\right)(\calW_{\Phi, \varepsilon}(\hat \mu,\hat \nu) - b\lambda),
\end{eqnarray}
where $\calW_{\Phi, \varepsilon}(\hat \mu,\hat \nu) := \inf_{\tilde \gamma \in \Pi(\hat \mu, \hat \nu)} \inf \Big[ t > 0 : \int_{\hat \G \times \hat \G} \Phi\left(\frac{\hat{c}(x, y)}{t}\right) \text{d}\tilde\gamma(x, y) - \varepsilon H(\tilde \gamma) \le 1\Big]$.

\textbf{Discussions.} Orlicz-EPT is a novel variant of EPT that incorporates Orlicz geometric structure. Leveraging~\citet{CM}'s insights and carefully calibrating the ground cost of the corresponding standard OT to ensure its nonnegativity, we are able to bypass all challenges of unbalanced measures and derive the proposed Orlicz-EPT from the standard OT, similar to OW~\citep{sturm2011generalized}.
We note that $\calOE_{\Phi, \varepsilon}$~\eqref{eq:regOrliczEPT} performs binary search with quadratic complexity $\mathcal{A}_{\varepsilon}$~\eqref{eq:regOW_t}, instead of dealing with super-cubic complexity $\mathcal{A}$~\eqref{eq:OW_t} in $\calOE_{\Phi}$~\eqref{eq:OrliczEPT}.
Unfortunately, the two-level optimization structure of $\calOE_{\Phi, \varepsilon}$ still retains significant complexity, severely limiting its practical applications, particularly in large-scale settings.  To address this computational challenge, in the next section we exploit the dual EPT and graph structure to develop a \emph{novel regularization} approach, resulting in the proposed Orlicz-Sobolev transport. This approach adopts the Orlicz geometric structure used in Orlicz-EPT, but offers a much more efficient computation.
  
\section{Orlicz-Sobolev Transport: A Scalable Variant of Orlicz-EPT}
\label{sec:OST}

In this section, we leverage the dual EPT and underlying graph structure to develop a novel regularization approach, resulting in the proposed \emph{Orlicz-Sobolev transport} (OST).


\textbf{Dual EPT.} For $b$-Lipschitz $w_1, w_2$ (w.r.t. $d_\G$), from~\citep[Corollary 3.2]{le2023scalable}, the dual EPT is
\begin{eqnarray}\label{equ:ETlambda}
\mathrm{ET}_\lambda(\mu,\nu) = \sup_{f\in \mathbb{U}} \int_\G f (\mu - \nu) - \frac{b\lambda}{2}\big[ \mu(\G) +  \nu(\G)\big],
\end{eqnarray}
where $ \mathbb{U} := \big\{f\in C(\G) :
 -w_2 - \frac{b\lambda}{2}\leq f \leq w_1  + \frac{b\lambda}{2}, \, |f(x)-f(y)|\leq b \, d_\G(x,y)\big\}$.




Let $\Psi$ be the complement $N$-function of $\Phi$ and $\omega$  be a nonnegative Borel measure on $\G$. Then let $\OrliczSobolevPsi(\G, \omega)$ be the graph-based Orlicz-Sobolev space~\citep[Definition 3.1]{le2024generalized} associated to $\Psi$ and $\omega$. Inspired by the approach of GST~\citep{le2024generalized}, we consider the critic function $f \in \mathbb{U}$ within $\OrliczSobolevPsi(\G, \omega)$. Consequently, the $b$-Lipschitz constraint on the critic function $f \in \mathbb{U}$ is replaced by $\norm{f'}_{L_{\Psi}} \le b$. 


For $f \in \OrliczSobolevPsi(\G, \omega)$, we have $f(x) = f(z_0) + \int_{[z_0,x]} f'(y) \omega(\mathrm{d}y),\forall x\in \G$. Let $\mathbf{1}$ be the indicator function. Then by using the generalized H\"older inequality~\citep[\S8.11]{adams2003sobolev} and $\norm{f'}_{L_{\Psi}} \le b$, we can control the integral part over the generalized graph derivative $f'$ for $f(x)$ as follows:
\begin{equation}\label{eq:Bound_IntegralGrad}
\int_{[z_0,x]} f'(y) \omega(\mathrm{d}y) 
\le 2 \norm{f'}_{L_{\Psi}}\norm{\mathbf{1}_{[z_0,x]}}_{L_{\Phi}}
\le 2b\norm{\mathbf{1}_{[z_0,x]}}_{L_{\Phi}} \le \frac{2b}{ \Phi^{-1}\!\left(1/\omega(\G) \right)},
\end{equation}
where the last inequality is due to the increasing property of $N$-function $\Phi$. Therefore, instead of the bounded constraint on the critic function $f$ in $\mathbb{U}$, we constraint only on $f(z_0)$.     




\begin{definition}[Orlicz-Sobolev transport (OST)]\label{def:OST}
For $\alpha \in [0, \frac{1}{2}(b\lambda + w_1(z_0) + w_2(z_0))]$, let $\calI_{\alpha} := \left[-w_2(z_0) - \frac{b\lambda}{2} + \alpha, w_1(z_0)  + \frac{b\lambda}{2} - \alpha \right]$. The Orlicz-Sobolev transport for $\mu, \nu\in \calP(\G)$ is defined 
\begin{eqnarray}\label{equ:OST}
\calOS_{\Phi, \alpha}(\mu,\nu ) :=  \sup_{f\in \mathbb{U}_{\Psi, \alpha}} \left[ \int_\G  f(x)\mu(\dd x) - \int_\G  f(x)\nu(\dd x) \right],
\end{eqnarray}
where $\mathbb{U}_{\Psi, \alpha}  :=  \big\{f  \in  \OrliczSobolevPsi(\G, \omega):\, \norm{f'}_{L_{\Psi}}  \le  b, f(z_0) \in  \calI_{\alpha}\big\}$.
\end{definition}
Intuitively, $\mathbb{U}_{\Psi, \alpha}$ is the collection of all functions $f$ expressed by $f(x) = s + \int_{[z_0, x]} h(y) \omega(\dd y), \forall x \in \G$, where $s \in \calI_{\alpha}$, and $\norm{h}_{L_{\Psi}} \le b$. The upper bound constraint on $\alpha$ is to ensure that $\calI_{\alpha}$ is nonempty. When $\alpha = 0$, $\calI_{\alpha}$ is the largest interval. Also, OST is an instance of the integral probability metric~\citep{muller1997integral}.



\paragraph{Computation.} Given unbalanced measures $\mu, \nu \in \calP(\G)$, for brevity let us define
\begin{align}\label{def:Theta}
    \Theta :=  \left\{\begin{array}{lr}
     w_1(z_0) + \frac{b\lambda}{2} -\alpha \hspace{1.2em} &\mbox{if}\quad\mu(\G)\geq \nu(\G),\\
    w_2(z_0) + \frac{b\lambda}{2} -\alpha \hspace{1.2em} &\mbox{if}\quad\mu(\G)< \nu(\G).
    \end{array}\right.
\end{align}

\begin{theorem}[Univariate optimization problem for OST]\label{thm:OST_computation} OST can be computed as follows
\begin{eqnarray}\label{eq:OrliczSobolevFunction}
    \calOS_{\Phi, \alpha}(\mu,\nu ) =  \Theta |\mu(\G)-\nu(\G)| + \inf_{k > 0} \frac{1}{k}\left( 1 + \int_{\G} \Phi\left(kb \left| \mu(\Lambda(x)) - \nu(\Lambda(x)) \right|\right) \omega(\text{d}x) \right). 
\end{eqnarray}
\end{theorem}

The proof is placed in Appendix \S\ref{appsubsec:thm:OST_computation}.




We derive the discrete case for OST which 
provides an explicit expression for the integral in~\eqref{eq:OrliczSobolevFunction}.
\begin{corollary}[Discrete case]\label{cor:OST_1d_optimization_discrete}
Let $\omega$ be the length measure of graph $\G$, and assume that input measures $\mu,\nu$ are supported on nodes in $V$ of graph $\G$.\footnote{It can be extended for measures supported in $\G$ (see \S\ref{appsubsec:further_discussion}).} Then, we have \begin{eqnarray}\label{eq:OST_1d_opt_discrete}
    \calOS_{\Phi, \alpha}(\mu,\nu ) =  \Theta |\mu(\G)-\nu(\G)| + \inf_{k > 0} \frac{1}{k}\left( 1 + \sum_{e \in E} w_e \Phi\!\left(kb \left| \mu(\gamma_{e}) - \nu(\gamma_{e}) \right|\right) \right).
\end{eqnarray}
\end{corollary}
The proof is placed in Appendix \S\ref{app:subsec:cor:OST_1d_optimization_discrete}. 


Therefore, OST can be efficiently computed by simply solving the \emph{univariate} optimization problem~\eqref{eq:OST_1d_opt_discrete}, thanks to the proposed novel regularization for critic functions in $\mathbb{U}_{\Psi, \alpha}$.

\begin{remark}[Non-physical graph]
In \S\ref{sec:preliminaries}, $\G$ is assumed to be a physical graph. Corollary~\ref{cor:OST_1d_optimization_discrete} implies that OST only depends on graph structure $(V, E)$ and edge weights $w_e$ when input measures are supported on nodes in $V$ of $\G$. Hence, OST is applicable for non-physical graph $\G$ for such cases.  
\end{remark}

\begin{remark}[Complementary pairs of $N$-functions]
Corollary~\ref{cor:OST_1d_optimization_discrete} also implies that one can compute OST with $N$-function $\Phi$ without involving its complementary $N$-function $\Psi$~\eqref{eq:OST_1d_opt_discrete}, unlike its definition~\eqref{equ:OST}. The univariate optimization formula~\eqref{eq:OST_1d_opt_discrete}) for OST requires that $\Psi$ is finite-valued, which is satisfied for any $N$-function $\Phi$ as it grows faster than linear.
\end{remark}

\textbf{Preprocessing for $\gamma_e$.} Similar to the GST computation~\citep{le2024generalized}, we precompute set $\gamma_e$~\eqref{sub-graph} for all edge $e$ in $\G$. More concretely, we apply the Dijkstra algorithm to recompute the shortest paths from $z_0$ to all other vertices in $V$ with complexity $\mathcal{O}(|E| + |V| \log{|V|})$, where $|\cdot|$ denotes the set cardinality. 



\textbf{Sparsity.} Observe that for every $x \in \text{supp}(\mu)$, its mass is gathered into $\mu(\gamma_e)$ if and only if $e \subset [z_0, x]$~\citep{le2024generalized}. Let $E_{\mu, \nu} \hspace{-0.2em}:=\hspace{-0.2em} \left\{e \! \in \! E \mid \exists z \! \in \! (\text{supp}(\mu) \cup \text{supp}(\nu)), e \subset [z_0, z] \right\} \subset E$. Then it suffices to compute the summation only over edges $e \in E_{\mu, \nu}$ in~\eqref{eq:OST_1d_opt_discrete} for OST, i.e., screen out all edges $e \in E \setminus E_{\mu, \nu}$.

\section{Theoretical Properties}
\label{sec:properties_OrliczSobolevTransport}

In this section, we leverage the computational efficiency of OST to derive its geometric structure and explore its connections to other transport distances.



\textbf{Geometric structures of OST.}
\begin{proposition}[Geometric structure]\label{prop:OST_geodesic_space} 
Let $0\leq \alpha< \frac{b\lambda}{2} +\min\{w_1(z_0), w_2(z_0)\}$~and~$\mu, \nu, \sigma \in \calP(\G)$.

i) $\calOS_{\Phi, \alpha}(\mu +\sigma,\nu +\sigma) = \calOS_{\Phi, \alpha}(\mu,\nu)$.

ii) $\calOS_{\Phi, \alpha}$ is a divergence,\footnote{$\calOS_{\Phi, \alpha}(\mu,\nu) \geq 0$, and $\calOS_{\Phi, \alpha}(\mu,\nu) = 0$ if and only if $\mu=\nu$.} and
$\calOS_{\Phi, \alpha}(\mu,\nu)\leq \calOS_{\Phi, \alpha}(\mu,\sigma) + \calOS_{\Phi, \alpha}(\sigma, \nu)$.

iii) With an additional assumption $w_1(z_0)=w_2(z_0)$, then $\calOS_{\Phi, \alpha}$ is a metric.

\end{proposition}


We next establish connections of OST with other transport distances, including GST~\citep{le2024generalized}, Sobolev transport (ST)~\citep{le2022st}, unbalanced Sobolev transport (UST)~\citep{le2023scalable}.




\textbf{Connection of OST with GST.} Denote $\mathcal{GS}_{\Phi}$ for the GST with $N$-function $\Phi$. 
\begin{proposition}\label{prop:relation_OST_GST}
For $\mu(\G) = \nu(\G)$, $b=1$, then $\calOS_{\Phi, \alpha}(\mu,\nu) = \mathcal{GS}_{\Phi}(\mu,\nu)$.
\end{proposition}

\textbf{Connection of OST with ST.} Denote $\mathcal{S}_p$ for the $p$-order ST, for $1 < p < \infty$. 
\begin{proposition}\label{prop:relation_OST_ST}
For $\mu(\G) = \nu(\G)$, $b=1$, and $\Phi(t) = \frac{(p-1)^{p-1}}{p^p} t^p$, then $\calOS_{\Phi, \alpha}(\mu,\nu) = \mathcal{S}_{p}(\mu,\nu)$.
\end{proposition}

\textbf{Connection of OST with UST.} Denote $\mathcal{US}_{p, \alpha}$ for UST, for $1 < p < \infty$. 
\begin{proposition}\label{prop:relation_OST_UST}
For $N$-function $\Phi(t) = \frac{(p-1)^{p-1}}{p^p} t^p$, then $\calOS_{\Phi, \alpha}(\mu,\nu) = \mathcal{US}_{p, \alpha}(\mu,\nu)$.
\end{proposition}


Additionally, we investigate the limit case for $N$-function, i.e., $\Phi(t) = t$,\footnote{Notice that $\Phi(t) = t$ is not an $N$-function due to its linear growth. It can be considered as the limit $p \to 1^+$ of the $N$-function $\Phi(t) = t^p$ with $p>1$.} for OST and Orlicz-EPT.

\begin{proposition}[Limit case for OST]\label{prop:limit_OST}
    For $\Phi(t) = t$, and with the same assumptions as in Corollary~\ref{cor:OST_1d_optimization_discrete}, then OST yields a closed-form expression as follows:
\begin{eqnarray}\label{eq:OST_1d_opt_discrete_limit}
    \calOS_{\Phi, \alpha}(\mu,\nu ) = b \sum_{e \in E} w_e \left| \mu(\gamma_{e}) - \nu(\gamma_{e}) \right| + \Theta |\mu(\G) - \nu(\G)|. 
\end{eqnarray}
\end{proposition}


    
\begin{proposition}[Limit case for Orlicz-EPT]\label{prop:limit_OrliczEPT}
    For $\Phi(t) = t$, then we have $\calOE_{\Phi}(\mu, \nu) = \mathrm{KT}(\mu, \nu)$ for every $\mu, \nu \in \calP(\G)$.
\end{proposition}

\begin{proposition}[Relation of OST and Orlicz-EPT]\label{prop:limit_OST_OrliczEPT}
For $\Phi(t) = t$, length measure $\omega$ on $\G$, $b$-Lipschitz $w_1, w_2$ (w.r.t. $d_{\G}$), $\alpha = 0$, and $p = 1$, then $\calOS_{\Phi, \alpha}(\mu,\nu ) \ge \calOE_{\Phi}(\mu, \nu) + \frac{b\lambda}{2}(\mu(\G) + \nu(\G))$. 
\end{proposition}


The proofs for these theoretical results (in \S\ref{sec:properties_OrliczSobolevTransport}) are respectively placed in \S\ref{app:subsec:prop:OST_geodesic_space}--\S\ref{app:subsec:prop:limit_OST_OrliczEPT}. 








\section{Related Works and Discussions}
\label{sec:related_works}

In this section, we discuss relations between our proposals with related works in the literature.

\textbf{GST~\citep{le2024generalized}.} Proposition~\ref{prop:relation_OST_GST} shows that OST provably generalizes GST~\citep{le2024generalized} for unbalanced measures. We emphasize that GST is restricted for balanced measures and is defined as an optimization over the critic function, making it nontrivial to directly extend it to accommodate unbalanced measures.

\textbf{EPT~\citep{le2021ept, le2023scalable, tran2025treeslicedEPT}.} Orlicz-EPT and OST are developed from the primal and dual EPT respectively. Notably, the corresponding standard OT following~\citep{CM} is not guarantee nonnegativity for ground cost, see~\citep{CM, le2021ept, le2023scalable}. Additionally, \citet{CM}'s insights may not be applicable to certain other UOT formulations, such as those proposed in~\citep{benamou2003numerical, frogner2015learning, chizat2018unbalanced, sejourne2019sinkhorn, pmlr-v151-sejourne22a, gangbo2019unnormalized, balaji2020robust, mukherjee2021outlier, nguyen2023unbalanced}. The calibration is essential to guarantee nonnegativity for ground cost of the corresponding standard OT for EPT, paving ways to develop Orlicz-EPT. Similar to OW, Orlicz-EPT is derived from standard OT, thereby circumventing the challenges associated with unbalanced measures. Furthermore, we derive a novel regularization, resulting in the proposed OST with an efficient computation.

\textbf{UST~\citep{le2023scalable} and ST~\citep{le2022st}} Proposition~\ref{prop:relation_OST_UST} shows that OST provably generalizes UST to a more general collection of $N$-functions. Consequently, OST also provably generalizes ST to unbalanced measures, and to a more general set of $N$-functions (see Proposition~\ref{prop:relation_OST_ST}).

\textbf{Measures on a graph.} We study OT problem between \emph{two unbalanced measures} supported on the \emph{same} graph, a setting also explored in~\citep{le2023scalable}. One should distinguish our considered problem with the research lines on computing either distances/discrepancies~\citep{petric2019got, xu2019gromov, xu2019scalable, dong2020copt, brogat2022learning, ma2023fused} or kernels~\citep{borgwardt2020graph, kriege2020survey, nikolentzos2021graph, sando2025TWWL} between \emph{two (different) input graphs}. 


\section{Experiments}
\label{sec:experiments}

In this section, we illustrate that the computation of Orlicz-EPT is costly. Especially, OST is several-order faster than Orlicz-EPT. Following the problem setups in~\citep{le2023scalable}, we evaluate OST for \emph{unbalanced measures supported a given graph},\footnote{One should distinguish the considered problem, i.e., compare two \emph{input unbalanced measures} supported in the \emph{same} graph, with either OT or Gromov-Wasserstein problem between \emph{two different input graphs} (\S\ref{sec:related_works}).} and show initial evidences on its advantages for document classification and topological data analysis (TDA). 

\textbf{Document classification.} We use 4 real-world document datasets: \texttt{TWITTER}, \texttt{RECIPE}, \texttt{CLASSIC}, and \texttt{AMAZON} as in~\citep{le2023scalable}, and summarize their characteristics in Figure~\ref{fg:DOC_SLE_10K}. By regarding each word in a document as a support with a unit mass, we represent each document as a nonnegative measure. Consequently, the representations of documents with different lengths are \emph{measures with different total mass}. We apply the same word embedding procedure in~\citep{le2023scalable} to map words into vectors in~$\R^{300}$. 

\textbf{TDA.} We consider orbit recognition on \texttt{Orbit} dataset~\citep{adams2017persistence}, and object shape classification on \texttt{MPEG7} dataset~\citep{latecki2000shape} as in \citep{le2023scalable}. We summarize these dataset characteristics in Figure~\ref{fg:TDA_SLE_10K1K}. We use persistence diagrams (PD), a multiset of $2$-dimensional data points summarized topological features, to represent objects of interest. We then consider each data point in PD as a support with a unit-mass, and represent PD as nonnegative measures. As a result, PD having different numbers of topological features are presented as \emph{measures with different total mass}.\footnote{We distinguish our problem setup with~\citep{le2024generalized}, where objects are represented as \emph{probability measures} instead.}


\textbf{Graph.} Following~\citep{le2023scalable}, we use the graphs $\G_{\text{Log}}$ and $\G_{\text{Sqrt}}$~\citep[\S5]{le2022st} for our experiments,\footnote{Due to the space limitation, corresponding experimental results for graph $\G_{\text{Log}}$ are placed in \S\ref{appsubsec:further_empirical_results}.} which empirically satisfy the assumptions in \S\ref{sec:preliminaries}. Additionally, we set $M=10^4$ for the number of nodes for these graphs, except experiments on \texttt{MPEG7} dataset with $M=10^3$ due to its small size.

\textbf{$N$-function.} Following~\citep{le2024generalized}, we consider two $N$-functions: $\Phi_1(t) = \exp(t) - t - 1$, and $\Phi_2(t) = \exp(t^2) - 1$, and the limit case of $N$-functions, i.e., $\Phi_0(t) = t$.

\textbf{Parameters.} For simplicity, we follow the experimental setup in~\citep{le2023scalable}. We set $\lambda = b = 1$, $\alpha=0$, and consider the weight functions $w_1(x) \!=\! w_2(x) \!=\! a_1 d_{\G}(z_0, x) + a_0$ where $a_1\!=\! b$ and $a_0\!=\!1$. The entropic regularization $\varepsilon$ is chosen from $\left\{0.01, 0.1, 1, 10\right\}$, typically via cross validation.

\textbf{Optimization algorithm.} For OST, we use \texttt{fmincon} MATLAB solver with trust-region-reflective algorithm, for solving the \emph{univariate} optimization problem~\eqref{eq:OST_1d_opt_discrete}. 

\textbf{SVM classification.} For document classification and TDA, we use support vector machine (SVM) with kernels $\exp(-\bar{t} \bar{d}(\cdot, \cdot))$, where $\bar{d}$ is a distance/discrepancy (e.g., OST, Orlicz-EPT) for unbalanced measures on a graph, and $\bar{t} > 0$. We regularize Gram matrices by adding a sufficiently large diagonal term for indefinite kernels~\citep{Cuturi-2013-Sinkhorn}. Additionally, we note that there are more than $29$M pairs for \texttt{AMAZON} which we need to evaluate distances/discrepancies for SVM in each run to illustrate the experiment scale.\footnote{See Table~\ref{tb:numpairs} in \S\ref{appsubsec:further_discussion} for the details.}

\textbf{Set up.} We randomly split each dataset into $70\%/30\%$ for training and test, and use $10$ repeats. We basically choose hyper-parameters via cross validation. More concretely, we choose kernel hyperparameter from $\{1/q_{s}, 0.5/q_{s}, 0.2/q_{s}\}$ with $s = 10, 20, \dotsc, 90$, where $q_s$ is the $s\%$ quantile of a subset of distances observed on a training set; SVM regularization hyperparameter from $\left\{0.01, 0.1, 1, 10\right\}$; root node $z_0$ from a random $10$-root-node subset of $V$ in graph $\G$. Note that reported time consumption includes all preprocessing procedures, e.g., preprocessing for $\gamma_e$ for OST.


\subsection{Computation}\label{subsec:computation}

\begin{figure}[h]
  \vspace{-10pt}
  \begin{center}
    \includegraphics[width=0.6\textwidth]{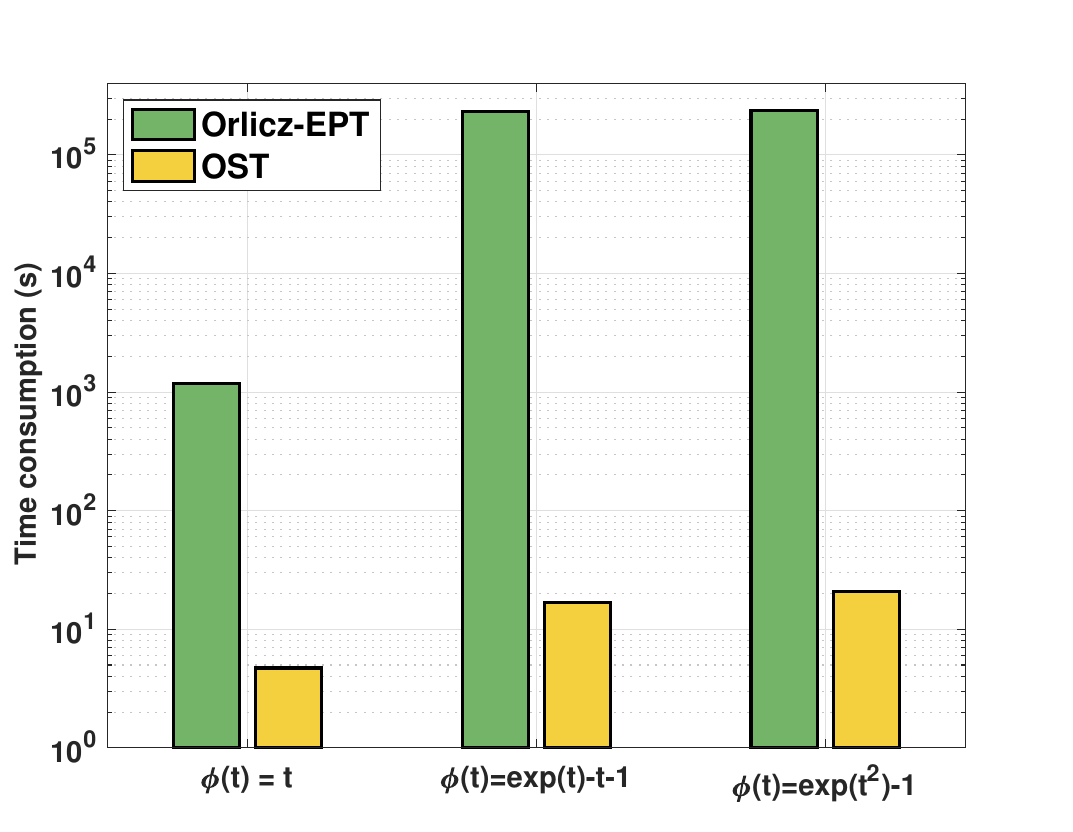}
  \end{center}
  \vspace{-11pt}
  \caption{Time consumption.}
  \label{fg:Time_OST_OrliczEPT_10K_SLE}
 \vspace{-6pt}
\end{figure}

We compare the time consumption of OST and Orlicz-EPT with $\Phi_1, \Phi_2$, and with the limit case $\Phi_0$.

\textbf{Set up.} We randomly sample $10^4$ pairs of nonnegative measures on \texttt{AMAZON} dataset for evaluation. We consider $M=10^3$ for graphs, and $\varepsilon = 0.1$ for Orlicz-EPT.

\textbf{Results.} We illustrate the time consumptions on $\G_{\text{Sqrt}}$ in Figure~\ref{fg:Time_OST_OrliczEPT_10K_SLE}. OST is several-order faster than Orlicz-EPT, i.e., at least $250\times, 13800\times, 11200\times$ for $\Phi_0, \Phi_1, \Phi_2$ respectively. Notably, for $N$-functions $\Phi_1, \Phi_2$, Orlicz-EPT takes at least \emph{$2.6$ days}, while OST takes less than \emph{$21$ seconds}. Note that for the limit case $\Phi_0$, Orlicz-EPT is equal to EPT on a graph (Proposition~\ref{prop:limit_OrliczEPT}), and OST admits a closed-form expression (Proposition~\ref{prop:limit_OST}) for a fast computation. Consequently, Orlicz-EPT and OST with $\Phi_0$ is more computationally efficient than those with $\Phi_1, \Phi_2$. 

\subsection{Document Classification}

\begin{figure*}[ht]
  \vspace{-6pt}
  \begin{center}
    \includegraphics[width=\textwidth]{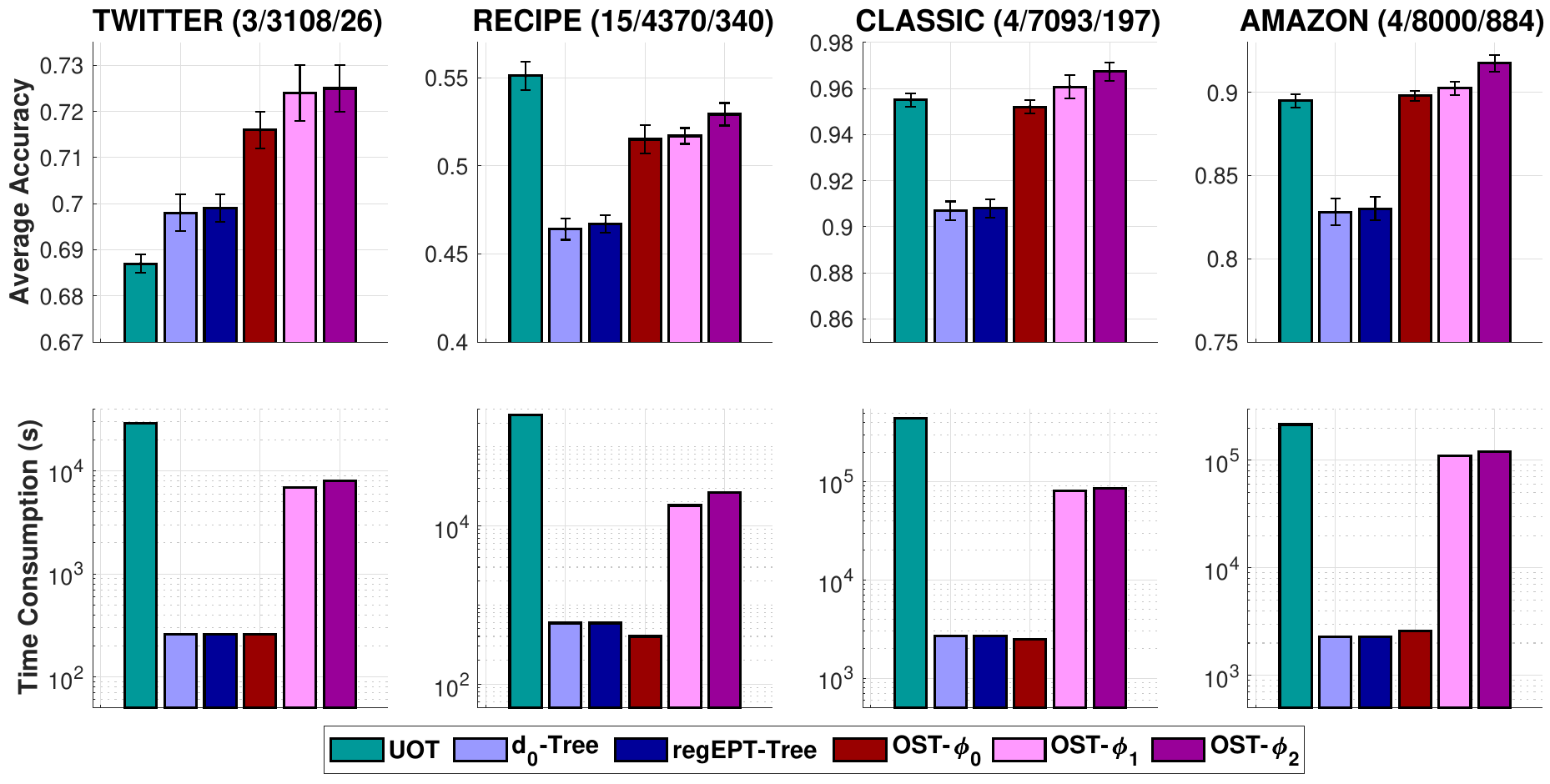}
  \end{center}
  \vspace{-10pt}
  \caption{Document classification on graph $\G_{\text{Sqrt}}$. For each dataset, the numbers in the parenthesis are respectively the number of classes; the number of documents; and the maximum number of unique words for each document.}
  \label{fg:DOC_SLE_10K}
 \vspace{-6pt}
\end{figure*}

\textbf{Set up.} We evaluate OST with $\Phi_0, \Phi_1, \Phi_2$ (\S\ref{subsec:computation}), denote them as OST-$\Phi_i$ for $i = 0, 1, 2$. We exclude Orlicz-EPT due to their heavy computations (\S\ref{subsec:computation}). Additionally, following~\citep{le2023scalable}, we consider UOT~\citep{frogner2015learning, sejourne2019sinkhorn} with ground cost $d_{\G}$,\footnote{\citet{sejourne2019sinkhorn, sejourne2023unbalanced} derived a debiased version for (Sinkhorn-based) UOT
which may be helpful in applications~\citep[\S B.3.3]{le2023scalable}. It has the same computational complexity as UOT, and is also empirically indefinite.} and special cases with tree-structure graph. More concretely, we randomly sample a tree from the given graph $\G$, then consider the regularized EPT and $d_{0}$, denoted as $d_0$-Tree and regEPT-Tree~\citep[Proposition 3.8, Equation (9)]{le2021ept}.



\textbf{Results.} We show SVM results and time consumptions of kernels on $\G_{\text{Sqrt}}$ in Figure~\ref{fg:DOC_SLE_10K}. The performances of OST with all $\Phi$ functions are comparable to UOT, but the computation of UOT is more costly than OST. Additionally, OST outperforms $d_0$-Tree and regEPT-Tree. However, the computations of OST-$\Phi_1$, OST-$\Phi_2$ are more expensive while the computation of OST-$\Phi_0$ is comparative to those fast-computational variants of UOT on tree (i.e., $d_0$-Tree and regEPT-Tree). Moreover, OST-$\Phi_1$ and OST-$\Phi_2$ improve performances of OST-$\Phi_0$, but their computational time is several-order higher, which may imply that Orlicz geometric structure in OST may be helpful for document classification. The performances of UOT also agree with observations in~\citep{le2023scalable}.


\subsection{Topological Data Analysis (TDA)}

\begin{figure}[h]
  \vspace{-6pt}
  \begin{center}
    \includegraphics[width=0.7\textwidth]{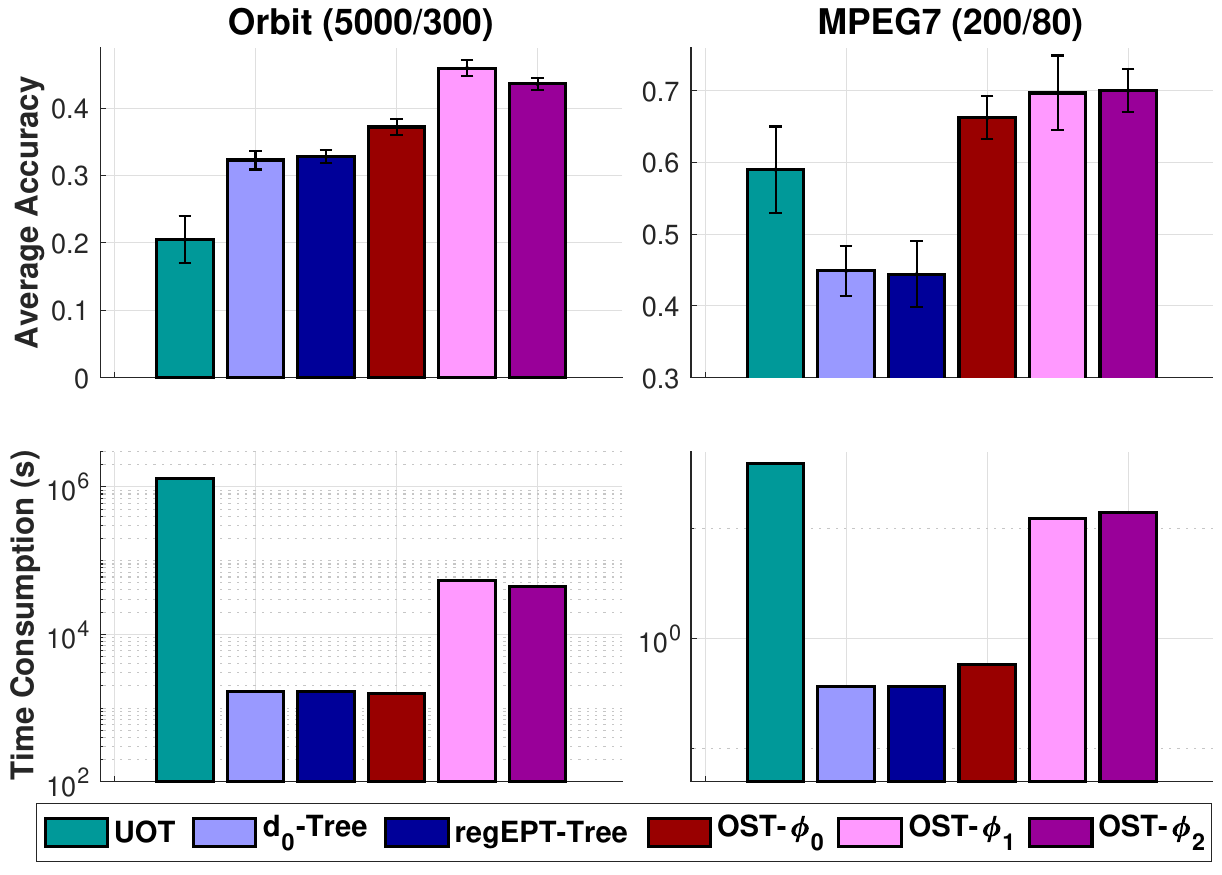}
  \end{center}
  \vspace{-10pt}
  \caption{TDA on graph $\G_{\text{Sqrt}}$. For each dataset, the numbers in the parenthesis are respectively the number of PD; and the maximum number of points in PD.}
  \label{fg:TDA_SLE_10K1K}
 \vspace{-6pt}
\end{figure}

\textbf{Set up.} Similarly, we evaluate OST-$\Phi_0$, OST-$\Phi_1$, OST-$\Phi_2$, UOT, $d_0$-Tree, and regEPT-Tree for TDA.

\textbf{Results.} We illustrate SVM results and time consumptions of kernels on $\G_{\text{Sqrt}}$ in Figure~\ref{fg:TDA_SLE_10K1K}. The performances of OST with all $\Phi$ functions compare favorably with other transport distance approaches. Especially, the performances of OST-$\Phi_1$ and OST-$\Phi_2$ compare favorably with those of OST-$\Phi_0$, but it comes with higher computational cost (i.e., OST-$\Phi_0$ has a closed-form expression (Proposition~\ref{prop:limit_OST})), which may imply that Orlicz geometric structure may be also helpful for TDA tasks.

\section{Conclusion}
\label{sec:conclusion}

In this work, we propose novel approaches to extend OW/GST for unbalanced measures on a graph. Building on the EPT problem and leveraging~\citet{CM}'s insights, we derive Orlicz-EPT by recasting it as a standard OT with a carefully calibrated ground cost, thereby bypassing challenges raised from unbalanced measures. Furthermore, by exploiting dual EPT and the underlying geometric structure, we formulate a novel regularization, resulting in the proposed OST, which is efficient in computation.
It provably suffices to compute OST by simply solving a univariate optimization problem, unlike the computationally intensive Orlicz-EPT. Moreover, we illustrate empirical evidence on the advantages of OST in document classification and topological data analysis.


\begin{ack}
We thank the area chairs and anonymous reviewers for their comments. KF has been supported in part by Grant-in-Aid for Transformative Research Areas (A) 22H05106 and JST CREST JPMJCR2015. HH acknowledges the support of JSPS KAKENHI JP25H01494 and JP23K24909. TL gratefully acknowledges the support of JSPS KAKENHI Grant number 23K11243, and Mitsui Knowledge Industry Co., Ltd. grant.
\end{ack}

\bibliography{bibSobolev25}

\begin{thebibliography}{74}
\providecommand{\natexlab}[1]{#1}
\providecommand{\url}[1]{\texttt{#1}}
\expandafter\ifx\csname urlstyle\endcsname\relax
  \providecommand{\doi}[1]{doi: #1}\else
  \providecommand{\doi}{doi: \begingroup \urlstyle{rm}\Url}\fi

\bibitem[Adams et~al.(2017)Adams, Emerson, Kirby, Neville, Peterson, Shipman,
  Chepushtanova, Hanson, Motta, and Ziegelmeier]{adams2017persistence}
Adams, H., Emerson, T., Kirby, M., Neville, R., Peterson, C., Shipman, P.,
  Chepushtanova, S., Hanson, E., Motta, F., and Ziegelmeier, L.
\newblock Persistence images: A stable vector representation of persistent
  homology.
\newblock \emph{Journal of Machine Learning Research}, 18\penalty0
  (1):\penalty0 218--252, 2017.

\bibitem[Adams \& Fournier(2003)Adams and Fournier]{adams2003sobolev}
Adams, R.~A. and Fournier, J.~J.
\newblock \emph{Sobolev spaces}.
\newblock Elsevier, 2003.

\bibitem[Altschuler \& Chewi(2024)Altschuler and Chewi]{altschuler2024faster}
Altschuler, J.~M. and Chewi, S.
\newblock Faster high-accuracy log-concave sampling via algorithmic warm
  starts.
\newblock \emph{Journal of the ACM}, 71\penalty0 (3):\penalty0 1--55, 2024.

\bibitem[Andoni et~al.(2018)Andoni, Lin, Sheng, Zhong, and
  Zhong]{andoni2018subspace}
Andoni, A., Lin, C., Sheng, Y., Zhong, P., and Zhong, R.
\newblock Subspace embedding and linear regression with {O}rlicz norm.
\newblock In \emph{International Conference on Machine Learning}, pp.\
  224--233. PMLR, 2018.

\bibitem[Balaji et~al.(2020)Balaji, Chellappa, and Feizi]{balaji2020robust}
Balaji, Y., Chellappa, R., and Feizi, S.
\newblock Robust optimal transport with applications in generative modeling and
  domain adaptation.
\newblock \emph{Advances in Neural Information Processing Systems},
  33:\penalty0 12934--12944, 2020.

\bibitem[Benamou(2003)]{benamou2003numerical}
Benamou, J.-D.
\newblock Numerical resolution of an “unbalanced” mass transport problem.
\newblock \emph{ESAIM: Mathematical Modelling and Numerical
  Analysis-Mod{\'e}lisation Math{\'e}matique et Analyse Num{\'e}rique},
  37\penalty0 (5):\penalty0 851--868, 2003.

\bibitem[Bonet et~al.(2024)Bonet, Nadjahi, Sejourne, Fatras, and
  Courty]{bonet2024slicing}
Bonet, C., Nadjahi, K., Sejourne, T., Fatras, K., and Courty, N.
\newblock Slicing unbalanced optimal transport.
\newblock \emph{Transactions on Machine Learning Research}, 2024.

\bibitem[Bonneel \& Coeurjolly(2019)Bonneel and Coeurjolly]{bonneel2019spot}
Bonneel, N. and Coeurjolly, D.
\newblock {SPOT}: sliced partial optimal transport.
\newblock \emph{ACM Transactions on Graphics (TOG)}, 38\penalty0 (4):\penalty0
  1--13, 2019.

\bibitem[Bonneel et~al.(2015)Bonneel, Rabin, Peyr{\'e}, and
  Pfister]{bonneel2015sliced}
Bonneel, N., Rabin, J., Peyr{\'e}, G., and Pfister, H.
\newblock Sliced and {R}adon {W}asserstein barycenters of measures.
\newblock \emph{Journal of Mathematical Imaging and Vision}, 51\penalty0
  (1):\penalty0 22--45, 2015.

\bibitem[Borgwardt et~al.(2020)Borgwardt, Ghisu, Llinares-L{\'o}pez, O’Bray,
  Rieck, et~al.]{borgwardt2020graph}
Borgwardt, K., Ghisu, E., Llinares-L{\'o}pez, F., O’Bray, L., Rieck, B.,
  et~al.
\newblock Graph kernels: State-of-the-art and future challenges.
\newblock \emph{Foundations and Trends{\textregistered} in Machine Learning},
  13\penalty0 (5-6):\penalty0 531--712, 2020.

\bibitem[Brogat-Motte et~al.(2022)Brogat-Motte, Flamary, Brouard, Rousu, and
  d’Alch{\'e} Buc]{brogat2022learning}
Brogat-Motte, L., Flamary, R., Brouard, C., Rousu, J., and d’Alch{\'e} Buc,
  F.
\newblock Learning to predict graphs with fused gromov-wasserstein barycenters.
\newblock In \emph{International Conference on Machine Learning}, pp.\
  2321--2335. PMLR, 2022.

\bibitem[Caffarelli \& McCann(2010)Caffarelli and McCann]{CM}
Caffarelli, L.~A. and McCann, R.~J.
\newblock Free boundaries in optimal transport and {M}onge-{A}mpere obstacle
  problems.
\newblock \emph{Annals of mathematics}, pp.\  673--730, 2010.

\bibitem[Chamakh et~al.(2020)Chamakh, Gobet, and Szab{\'o}]{chamakh2020orlicz}
Chamakh, L., Gobet, E., and Szab{\'o}, Z.
\newblock {O}rlicz random {F}ourier features.
\newblock \emph{The Journal of Machine Learning Research}, 21\penalty0
  (1):\penalty0 5739--5775, 2020.

\bibitem[Chapel \& Tavenard(2025)Chapel and Tavenard]{chapel2025one}
Chapel, L. and Tavenard, R.
\newblock One for all and all for one: Efficient computation of partial
  {W}asserstein distances on the line.
\newblock In \emph{The Thirteenth International Conference on Learning
  Representations}, 2025.

\bibitem[Chapel et~al.(2020)Chapel, Alaya, and Gasso]{chapel2020partial}
Chapel, L., Alaya, M.~Z., and Gasso, G.
\newblock Partial optimal tranport with applications on positive-unlabeled
  learning.
\newblock \emph{Advances in Neural Information Processing Systems},
  33:\penalty0 2903--2913, 2020.

\bibitem[Chapel et~al.(2021)Chapel, Flamary, Wu, F{\'e}votte, and
  Gasso]{chapel2021unbalanced}
Chapel, L., Flamary, R., Wu, H., F{\'e}votte, C., and Gasso, G.
\newblock Unbalanced optimal transport through non-negative penalized linear
  regression.
\newblock \emph{Advances in Neural Information Processing Systems},
  34:\penalty0 23270--23282, 2021.

\bibitem[Chewi(2023)]{chewi2023optimization}
Chewi, S.
\newblock \emph{An optimization perspective on log-concave sampling and
  beyond}.
\newblock PhD thesis, Massachusetts Institute of Technology, 2023.

\bibitem[Chizat et~al.(2018)Chizat, Peyr{\'e}, Schmitzer, and
  Vialard]{chizat2018unbalanced}
Chizat, L., Peyr{\'e}, G., Schmitzer, B., and Vialard, F.-X.
\newblock Unbalanced optimal transport: Dynamic and {K}antorovich formulations.
\newblock \emph{Journal of Functional Analysis}, 274\penalty0 (11):\penalty0
  3090--3123, 2018.

\bibitem[Cover \& Thomas(1999)Cover and Thomas]{cover1999elements}
Cover, T.~M. and Thomas, J.~A.
\newblock \emph{Elements of information theory}.
\newblock John Wiley \& Sons, 1999.

\bibitem[Cuturi(2013)]{Cuturi-2013-Sinkhorn}
Cuturi, M.
\newblock Sinkhorn distances: {L}ightspeed computation of optimal transport.
\newblock In \emph{Advances in Neural Information Processing Systems}, pp.\
  2292--2300, 2013.

\bibitem[Deng et~al.(2022)Deng, Song, Weinstein, and Zhang]{deng2022fast}
Deng, Y., Song, Z., Weinstein, O., and Zhang, R.
\newblock Fast distance oracles for any symmetric norm.
\newblock \emph{Advances in Neural Information Processing Systems},
  35:\penalty0 7304--7317, 2022.

\bibitem[Dong \& Sawin(2020)Dong and Sawin]{dong2020copt}
Dong, Y. and Sawin, W.
\newblock {COPT}: {C}oordinated optimal transport on graphs.
\newblock \emph{Advances in Neural Information Processing Systems},
  33:\penalty0 19327--19338, 2020.

\bibitem[Edelsbrunner \& Harer(2008)Edelsbrunner and
  Harer]{edelsbrunner2008persistent}
Edelsbrunner, H. and Harer, J.
\newblock Persistent homology - {A} survey.
\newblock \emph{Contemporary Mathematics}, 453:\penalty0 257--282, 2008.

\bibitem[Fatras et~al.(2021)Fatras, S{\'e}journ{\'e}, Flamary, and
  Courty]{fatras2021unbalanced}
Fatras, K., S{\'e}journ{\'e}, T., Flamary, R., and Courty, N.
\newblock Unbalanced minibatch optimal transport; applications to domain
  adaptation.
\newblock In \emph{International Conference on Machine Learning}, pp.\
  3186--3197. PMLR, 2021.

\bibitem[Figalli(2010)]{figalli2010optimal}
Figalli, A.
\newblock The optimal partial transport problem.
\newblock \emph{Archive for rational mechanics and analysis}, 195\penalty0
  (2):\penalty0 533--560, 2010.

\bibitem[Frogner et~al.(2015)Frogner, Zhang, Mobahi, Araya, and
  Poggio]{frogner2015learning}
Frogner, C., Zhang, C., Mobahi, H., Araya, M., and Poggio, T.~A.
\newblock Learning with a {W}asserstein loss.
\newblock In \emph{Advances in neural information processing systems}, pp.\
  2053--2061, 2015.

\bibitem[Gangbo et~al.(2019)Gangbo, Li, Osher, and
  Puthawala]{gangbo2019unnormalized}
Gangbo, W., Li, W., Osher, S., and Puthawala, M.
\newblock Unnormalized optimal transport.
\newblock \emph{Journal of Computational Physics}, 399:\penalty0 108940, 2019.

\bibitem[Guha et~al.(2023)Guha, Ho, and Nguyen]{GuhaHN23}
Guha, A., Ho, N., and Nguyen, X.
\newblock On excess mass behavior in {G}aussian mixture models with
  {O}rlicz-{W}asserstein distances.
\newblock In \emph{International Conference on Machine Learning, {ICML}},
  volume 202, pp.\  11847--11870. {PMLR}, 2023.

\bibitem[Guittet(2002)]{guittet2002extended}
Guittet, K.
\newblock Extended {K}antorovich norms: a tool for optimization.
\newblock \emph{INRIA report}, 2002.

\bibitem[Hanin(1992)]{hanin1992kantorovich}
Hanin, L.~G.
\newblock Kantorovich-{R}ubinstein norm and its application in the theory of
  {L}ipschitz spaces.
\newblock \emph{Proceedings of the American Mathematical Society}, 115\penalty0
  (2):\penalty0 345--352, 1992.

\bibitem[Hertzsch et~al.(2007)Hertzsch, Sturman, and Wiggins]{hertzsch2007dna}
Hertzsch, J.-M., Sturman, R., and Wiggins, S.
\newblock {DNA} microarrays: {D}esign principles for maximizing ergodic,
  chaotic mixing.
\newblock \emph{Small}, 3\penalty0 (2):\penalty0 202--218, 2007.

\bibitem[Kell(2017)]{kell2017interpolation}
Kell, M.
\newblock On interpolation and curvature via {W}asserstein geodesics.
\newblock \emph{Advances in Calculus of Variations}, 10\penalty0 (2):\penalty0
  125--167, 2017.

\bibitem[Kondratyev et~al.(2016)Kondratyev, Monsaingeon, and
  Vorotnikov]{kondratyev2016new}
Kondratyev, S., Monsaingeon, L., and Vorotnikov, D.
\newblock A new optimal transport distance on the space of finite {R}adon
  measures.
\newblock \emph{Advances in Differential Equations}, 21\penalty0
  (11/12):\penalty0 1117--1164, 2016.

\bibitem[Kriege et~al.(2020)Kriege, Johansson, and Morris]{kriege2020survey}
Kriege, N.~M., Johansson, F.~D., and Morris, C.
\newblock A survey on graph kernels.
\newblock \emph{Applied Network Science}, 5:\penalty0 1--42, 2020.

\bibitem[Latecki et~al.(2000)Latecki, Lakamper, and Eckhardt]{latecki2000shape}
Latecki, L.~J., Lakamper, R., and Eckhardt, T.
\newblock Shape descriptors for non-rigid shapes with a single closed contour.
\newblock In \emph{Proceedings of the IEEE Conference on Computer Vision and
  Pattern Recognition (CVPR)}, volume~1, pp.\  424--429, 2000.

\bibitem[Le \& Nguyen(2021)Le and Nguyen]{le2021ept}
Le, T. and Nguyen, T.
\newblock Entropy partial transport with tree metrics: Theory and practice.
\newblock In \emph{International Conference on Artificial Intelligence and
  Statistics (AISTATS)}, pp.\  3835--3843, 2021.

\bibitem[Le et~al.(2019)Le, Yamada, Fukumizu, and Cuturi]{LYFC}
Le, T., Yamada, M., Fukumizu, K., and Cuturi, M.
\newblock Tree-sliced variants of {W}asserstein distances.
\newblock In \emph{Advances in neural information processing systems}, pp.\
  12283--12294, 2019.

\bibitem[Le et~al.(2021)Le, Ho, and Yamada]{le2021fba}
Le, T., Ho, N., and Yamada, M.
\newblock Flow-based alignment approaches for probability measures in different
  spaces.
\newblock In \emph{International Conference on Artificial Intelligence and
  Statistics (AISTATS)}. 2021.

\bibitem[Le et~al.(2022)Le, Nguyen, Phung, and Nguyen]{le2022st}
Le, T., Nguyen, T., Phung, D., and Nguyen, V.~A.
\newblock Sobolev transport: A scalable metric for probability measures with
  graph metrics.
\newblock In \emph{International Conference on Artificial Intelligence and
  Statistics}, pp.\  9844--9868, 2022.

\bibitem[Le et~al.(2023)Le, Nguyen, and Fukumizu]{le2023scalable}
Le, T., Nguyen, T., and Fukumizu, K.
\newblock Scalable unbalanced {S}obolev transport for measures on a graph.
\newblock In \emph{International Conference on Artificial Intelligence and
  Statistics}, pp.\  8521--8560, 2023.

\bibitem[Le et~al.(2024{\natexlab{a}})Le, Nguyen, and
  Fukumizu]{le2024generalized}
Le, T., Nguyen, T., and Fukumizu, K.
\newblock Generalized {S}obolev transport for probability measures on a graph.
\newblock In \emph{Forty-first International Conference on Machine Learning},
  2024{\natexlab{a}}.

\bibitem[Le et~al.(2024{\natexlab{b}})Le, Nguyen, and Fukumizu]{le2024optimal}
Le, T., Nguyen, T., and Fukumizu, K.
\newblock Optimal transport for measures with noisy tree metric.
\newblock In \emph{International Conference on Artificial Intelligence and
  Statistics}, pp.\  3115--3123, 2024{\natexlab{b}}.

\bibitem[Le et~al.(2025)Le, Nguyen, Hino, and Fukumizu]{le2025scalable}
Le, T., Nguyen, T., Hino, H., and Fukumizu, K.
\newblock Scalable {S}obolev {IPM} for probability measures on a graph.
\newblock In \emph{Forty-second International Conference on Machine Learning},
  2025.

\bibitem[Lellmann et~al.(2014)Lellmann, Lorenz, Schonlieb, and
  Valkonen]{lellmann2014imaging}
Lellmann, J., Lorenz, D.~A., Schonlieb, C., and Valkonen, T.
\newblock Imaging with {K}antorovich--{R}ubinstein discrepancy.
\newblock \emph{SIAM Journal on Imaging Sciences}, 7\penalty0 (4):\penalty0
  2833--2859, 2014.

\bibitem[Liero et~al.(2018)Liero, Mielke, and Savar{\'e}]{Liero2018}
Liero, M., Mielke, A., and Savar{\'e}, G.
\newblock Optimal entropy-transport problems and a new
  {H}ellinger--{K}antorovich distance between positive measures.
\newblock \emph{Inventiones mathematicae}, 211\penalty0 (3):\penalty0
  969--1117, 2018.

\bibitem[Lorenz \& Mahler(2022)Lorenz and Mahler]{lorenz2022orlicz}
Lorenz, D. and Mahler, H.
\newblock {O}rlicz space regularization of continuous optimal transport
  problems.
\newblock \emph{Applied Mathematics \& Optimization}, 85\penalty0 (2):\penalty0
  14, 2022.

\bibitem[Ma et~al.(2023)Ma, Chu, Wang, Lin, Zhao, Ma, and Zhu]{ma2023fused}
Ma, X., Chu, X., Wang, Y., Lin, Y., Zhao, J., Ma, L., and Zhu, W.
\newblock Fused {G}romov-{W}asserstein graph mixup for graph-level
  classifications.
\newblock \emph{Advances in Neural Information Processing Systems},
  36:\penalty0 15252--15276, 2023.

\bibitem[Mukherjee et~al.(2021)Mukherjee, Guha, Solomon, Sun, and
  Yurochkin]{mukherjee2021outlier}
Mukherjee, D., Guha, A., Solomon, J.~M., Sun, Y., and Yurochkin, M.
\newblock Outlier-robust optimal transport.
\newblock In \emph{International Conference on Machine Learning}, pp.\
  7850--7860. PMLR, 2021.

\bibitem[M{\"u}ller(1997)]{muller1997integral}
M{\"u}ller, A.
\newblock Integral probability metrics and their generating classes of
  functions.
\newblock \emph{Advances in Applied Probability}, 29\penalty0 (2):\penalty0
  429--443, 1997.

\bibitem[Musielak(2006)]{musielak2006orlicz}
Musielak, J.
\newblock \emph{Orlicz spaces and modular spaces}, volume 1034.
\newblock Springer, 2006.

\bibitem[Nguyen et~al.(2024)Nguyen, Zhang, Le, and Ho]{nguyen2024randompathSW}
Nguyen, K., Zhang, S., Le, T., and Ho, N.
\newblock Sliced {W}asserstein with random-path projecting directions.
\newblock In \emph{Forty-first International Conference on Machine Learning},
  2024.

\bibitem[Nguyen et~al.(2023)Nguyen, Nguyen, Zhou, and
  Nguyen]{nguyen2023unbalanced}
Nguyen, Q.~M., Nguyen, H.~H., Zhou, Y., and Nguyen, L.~M.
\newblock On unbalanced optimal transport: Gradient methods, sparsity and
  approximation error.
\newblock \emph{The Journal of Machine Learning Research}, 2023.

\bibitem[Nikolentzos et~al.(2021)Nikolentzos, Siglidis, and
  Vazirgiannis]{nikolentzos2021graph}
Nikolentzos, G., Siglidis, G., and Vazirgiannis, M.
\newblock Graph kernels: A survey.
\newblock \emph{Journal of Artificial Intelligence Research}, 72:\penalty0
  943--1027, 2021.

\bibitem[Paty \& Cuturi(2019)Paty and Cuturi]{pmlr-v97-paty19a}
Paty, F.-P. and Cuturi, M.
\newblock Subspace robust {W}asserstein distances.
\newblock In \emph{Proceedings of the 36th International Conference on Machine
  Learning}, pp.\  5072--5081, 2019.

\bibitem[Petric~Maretic et~al.(2019)Petric~Maretic, El~Gheche, Chierchia, and
  Frossard]{petric2019got}
Petric~Maretic, H., El~Gheche, M., Chierchia, G., and Frossard, P.
\newblock {GOT}: {A}n optimal transport framework for graph comparison.
\newblock \emph{Advances in Neural Information Processing Systems}, 32, 2019.

\bibitem[Pham et~al.(2020)Pham, Le, Ho, Pham, and Bui]{pham2020unbalanced}
Pham, K., Le, K., Ho, N., Pham, T., and Bui, H.
\newblock On unbalanced optimal transport: An analysis of {S}inkhorn algorithm.
\newblock In \emph{Proceedings of the International Conference on Machine
  Learning}, 2020.

\bibitem[Piccoli \& Rossi(2014)Piccoli and Rossi]{P1}
Piccoli, B. and Rossi, F.
\newblock Generalized {W}asserstein distance and its application to transport
  equations with source.
\newblock \emph{Archive for Rational Mechanics and Analysis}, 211\penalty0
  (1):\penalty0 335--358, 2014.

\bibitem[Piccoli \& Rossi(2016)Piccoli and Rossi]{P2}
Piccoli, B. and Rossi, F.
\newblock On properties of the generalized {W}asserstein distance.
\newblock \emph{Archive for Rational Mechanics and Analysis}, 222\penalty0
  (3):\penalty0 1339--1365, 2016.

\bibitem[Rabin et~al.(2011)Rabin, Peyr{\'e}, Delon, and
  Bernot]{rabin2011wasserstein}
Rabin, J., Peyr{\'e}, G., Delon, J., and Bernot, M.
\newblock Wasserstein barycenter and its application to texture mixing.
\newblock In \emph{International Conference on Scale Space and Variational
  Methods in Computer Vision}, pp.\  435--446, 2011.

\bibitem[Rao \& Ren(1991)Rao and Ren]{rao1991theory}
Rao, M.~M. and Ren, Z.~D.
\newblock Theory of {O}rlicz spaces.
\newblock \emph{Marcel Dekker}, 1991.

\bibitem[Sando et~al.(2025)Sando, Le, and Hino]{sando2025TWWL}
Sando, K., Le, T., and Hino, H.
\newblock Tree structure for the categorical {W}asserstein
  {W}eisfeiler-{L}ehman graph kernel.
\newblock \emph{Transactions on Machine Learning Research (TMLR)}, 2025.

\bibitem[Sato et~al.(2020)Sato, Yamada, and Kashima]{sato2020fast}
Sato, R., Yamada, M., and Kashima, H.
\newblock Fast unbalanced optimal transport on tree.
\newblock In \emph{Advances in neural information processing systems}, 2020.

\bibitem[S{\'e}journ{\'e} et~al.(2019)S{\'e}journ{\'e}, Feydy, Vialard,
  Trouv{\'e}, and Peyr{\'e}]{sejourne2019sinkhorn}
S{\'e}journ{\'e}, T., Feydy, J., Vialard, F.-X., Trouv{\'e}, A., and Peyr{\'e},
  G.
\newblock Sinkhorn divergences for unbalanced optimal transport.
\newblock \emph{arXiv preprint arXiv:1910.12958}, 2019.

\bibitem[S{\'e}journ{\'e} et~al.(2022)S{\'e}journ{\'e}, Vialard, and
  Peyr\'e]{pmlr-v151-sejourne22a}
S{\'e}journ{\'e}, T., Vialard, F.-X., and Peyr\'e, G.
\newblock Faster unbalanced optimal transport: Translation invariant {S}inkhorn
  and {1-D} {F}rank-{W}olfe.
\newblock In \emph{Proceedings of The 25th International Conference on
  Artificial Intelligence and Statistics}, volume 151, pp.\  4995--5021. PMLR,
  2022.

\bibitem[S{\'e}journ{\'e} et~al.(2023)S{\'e}journ{\'e}, Peyr{\'e}, and
  Vialard]{sejourne2023unbalanced}
S{\'e}journ{\'e}, T., Peyr{\'e}, G., and Vialard, F.-X.
\newblock Unbalanced optimal transport, from theory to numerics.
\newblock \emph{Handbook of Numerical Analysis}, 24:\penalty0 407--471, 2023.

\bibitem[Song et~al.(2019)Song, Wang, Yang, Zhang, and
  Zhong]{song2019efficient}
Song, Z., Wang, R., Yang, L., Zhang, H., and Zhong, P.
\newblock Efficient symmetric norm regression via linear sketching.
\newblock \emph{Advances in Neural Information Processing Systems}, 32, 2019.

\bibitem[Sturm(2011)]{sturm2011generalized}
Sturm, K.-T.
\newblock Generalized {O}rlicz spaces and {W}asserstein distances for
  convex--concave scale functions.
\newblock \emph{Bulletin des sciences math{\'e}matiques}, 135\penalty0
  (6-7):\penalty0 795--802, 2011.

\bibitem[Tran et~al.(2025{\natexlab{a}})Tran, Tran, Chu, Pham, Ghaoui, Le, and
  Nguyen]{tran2025nonlinear}
Tran, T., Tran, V.-H., Chu, T., Pham, T., Ghaoui, L.~E., Le, T., and Nguyen, T.
\newblock Tree-sliced {W}asserstein distance with nonlinear projection.
\newblock In \emph{Forty-second International Conference on Machine Learning},
  2025{\natexlab{a}}.

\bibitem[Tran et~al.(2025{\natexlab{b}})Tran, Chu, Nguyen, Pham, Le, and
  Nguyen]{tran2025spherical}
Tran, V.-H., Chu, T., Nguyen, K., Pham, T., Le, T., and Nguyen, T.
\newblock Spherical tree-sliced {W}asserstein distance.
\newblock In \emph{The Thirteenth International Conference on Learning
  Representations}, 2025{\natexlab{b}}.

\bibitem[Tran et~al.(2025{\natexlab{c}})Tran, Nguyen, Pham, Chu, Le, and
  Nguyen]{tran2025distancebased}
Tran, V.-H., Nguyen, K., Pham, T., Chu, T., Le, T., and Nguyen, T.
\newblock Distance-based tree-sliced {W}asserstein distance.
\newblock In \emph{The Thirteenth International Conference on Learning
  Representations}, 2025{\natexlab{c}}.

\bibitem[Tran et~al.(2025{\natexlab{d}})Tran, Pham, Tran, Nguyen, Chu, Le, and
  Nguyen]{tran2025geometric}
Tran, V.-H., Pham, T., Tran, T., Nguyen, K., Chu, T., Le, T., and Nguyen, T.
\newblock Tree-sliced {W}asserstein distance: A geometric perspective.
\newblock In \emph{Forty-second International Conference on Machine Learning},
  2025{\natexlab{d}}.

\bibitem[Tran et~al.(2025{\natexlab{e}})Tran, Tran, Chu, Le, and
  Nguyen]{tran2025treeslicedEPT}
Tran, V.-H., Tran, T., Chu, T., Le, T., and Nguyen, T.
\newblock Tree-sliced {E}ntropy {P}artial {T}ransport.
\newblock In \emph{The Thirty-ninth Annual Conference on Neural Information
  Processing Systems}, 2025{\natexlab{e}}.

\bibitem[Xu et~al.(2019{\natexlab{a}})Xu, Luo, and Carin]{xu2019scalable}
Xu, H., Luo, D., and Carin, L.
\newblock Scalable {G}romov-{W}asserstein learning for graph partitioning and
  matching.
\newblock \emph{Advances in neural information processing systems}, 32,
  2019{\natexlab{a}}.

\bibitem[Xu et~al.(2019{\natexlab{b}})Xu, Luo, Zha, and Duke]{xu2019gromov}
Xu, H., Luo, D., Zha, H., and Duke, L.~C.
\newblock {G}romov-{W}asserstein learning for graph matching and node
  embedding.
\newblock In \emph{International conference on machine learning}, pp.\
  6932--6941. PMLR, 2019{\natexlab{b}}.

\end{thebibliography}
\bibliographystyle{icml2025}


%


\clearpage
\appendix

\begin{center}
{\bf{\Large{\textit{Supplement to}  ``An Efficient Orlicz-Sobolev Approach for
Transporting Unbalanced Measures on a Graph''}}}
\end{center}

In this appendix, we provide further theoretical results and detailed proofs in \S\ref{appsec:detailed_proofs_theoretical_results}. Additionally, we give brief reviews on related notions used in our work, together with further discussions, and empirical results in \S\ref{appsec:further_results_discussions}.


\section{Detailed Proofs and Further Theoretical Results}\label{appsec:detailed_proofs_theoretical_results}

In this section, we provide further theoretical results, and detailed proofs for all the theoretical results.

\subsection{Further Theoretical Results}\label{appsubsec:further_theoretical_results}

We investigate special cases for OST, alternative upper limit for $\mathcal{A}_{\varepsilon}$, and the limit case of $N$-function for entropic regularized Orlicz-EPT.

\subsubsection{Special Cases for OST}

We exam the special cases of OST when graph $\G$ is a tree.

\begin{proposition}[Relation of OST and a variant of regularized EPT]\label{prop:limit_OST_dalpha}
    Under the same assumptions as in Proposition~\ref{prop:limit_OST}, and assume in addition that graph $\G$ is a tree, then 
    \[
    \calOS_{\Phi, \alpha}(\mu,\nu ) = d_{\alpha}(\mu, \nu),
    \]
    where $d_{\alpha}$ is a variant of the regularized EPT in~\citep[Equation (9)]{le2021ept}. 
\end{proposition}

The proof is placed in Appendix~\S\ref{app:subsec:prop:limit_OST_dalpha}.

\begin{proposition}[Relation of OST and standard OT]\label{prop:limit_OST_OT}
Under the same assumptions as in Proposition~\ref{prop:limit_OST_dalpha}, and assume in addition that $\mu(\G) = \nu(\G)$ and $b=1$, then 
\[
\calOS_{\Phi, \alpha}(\mu,\nu ) = \calW_{d_{\G}}(\mu, \nu),
\]
where $\calW_{d_{\G}}$ is the standard OT with graph metric ground cost $d_{\G}$.
\end{proposition}

The proof is placed in Appendix~\S\ref{app:subsec:prop:limit_OST_OT}.

\subsubsection{Upper Limit of $\mathcal{A}_{\varepsilon}$ w.r.t. Entropic Regularized OT}\label{app:subsec:upper_bound_over_entropicOT}

With a technical assumption that entropic regularized input is nonnegative for $N$-function $\Phi$,\footnote{The technical assumption is specified in the proof (\S\ref{appsubsec:rm:upperlimit_regOT}).} we derive an alternative upper limit of $\mathcal{A}_{\varepsilon}$ w.r.t. entropic regularized OT as summarized in the following proposition. 
\begin{proposition}[Upper bound w.r.t. entropic regularized OT]\label{rm:upperlimit_regOT}
We have
\[
\mathcal{A}_{\varepsilon} \! \left( \frac{\mathcal{W}_{\hat{c},\varepsilon}(\hat\mu, \hat\nu) + \frac{\varepsilon} {2}\left(H(\hat\mu) + H(\hat\nu)\right)}{\Phi^{-1}(1 + \varepsilon\left[ H(\hat \mu) + H(\hat \nu) - 1 \right])}; \hat \mu, \hat \nu \right) \ge 1,
\]
where $\calW_{\hat{c},\varepsilon}(\hat \mu,\hat \nu)$ is the entropic regularized OT between probability measures $\hat \mu,\hat \nu$ with ground cost $\hat{c}$, defined as
\begin{equation}\label{eq:Wc_eps}
\calW_{\hat{c},\varepsilon}(\hat \mu,\hat \nu) := \inf_{\tilde \gamma \in \Pi(\hat\mu,\hat \nu)} \left[  \int_{\hat \G\times \hat \G} \hat c(x,y) \tilde\gamma(\dd x, \dd y) - \varepsilon H(\tilde \gamma) \right].
\end{equation}
\end{proposition}

The proof is placed in Appendix \S\ref{appsubsec:rm:upperlimit_regOT}.

\subsubsection{Limit Case for Entropic Regularized Orlicz-EPT}\label{app:subsec:limit_case_entropicOrliczEPT}

We consider the limit case for $N$-function, i.e., $\Phi(t) = t$, for $\calOE_{\Phi, \varepsilon}$, similar to Proposition~\ref{prop:limit_OrliczEPT} for the original Orlicz-EPT.
\begin{proposition}[Limit case for entropic regularized Orlicz-EPT]\label{prop:limit_regOrliczEPT}
    For $\Phi(t) = t$, and $\mu, \nu \in \calP(\G)$, we have
    \begin{eqnarray}\label{eq:regOrliczEPT_limit}
    \calOE_{\Phi, \varepsilon}(\mu, \nu) = \left(\mu(\G) + \nu(\G)\right)(\calW_{\hat{c},\varepsilon}(\hat \mu,\hat \nu) - b\lambda),
    \end{eqnarray}
    where $\calW_{\hat{c}, \varepsilon}$ is the entropic regularized optimal transport (see Equation~\eqref{eq:Wc_eps}).
\end{proposition}

The proof is placed in Appendix~\S\ref{app:subsec:prop:limit_regOrliczEPT}.

\subsection{Detailed Proofs}\label{appsubsec:detailed_proofs}

\subsubsection{Proof for Proposition~\ref{prop:ET_KT}}\label{appsubsec:ET_KT}
\begin{proof}
Consider the cost function $\tilde c$ on $\hat\G$ as follows:
\begin{equation}\label{eq:tildec_hat_cost}
\tilde c(x, y) := 
\left\{\begin{array}{lr}
\!\!b(d_{\G}(x,y) - \lambda) \hspace{1.1 em} \mbox{ if } x,y\in \G,\\
\!\!w_1(x) \hspace{5.0 em} \mbox{ if }  x\in \G \mbox{ and } y=\hat s,\\
 \!\! w_2(y) \hspace{5.0 em}  \mbox{ if }  x=\hat s \mbox{ and } y\in \G,\\
  \!\! 0 \hspace{7.0 em} \mbox{ if }  x=y=\hat s.
\end{array}\right.
\end{equation}
Additionally, for unbalanced measures $\mu, \nu$, we construct corresponding balanced measures $\tilde\mu := {\mu +\nu(\G) \delta_{\hat s}}$ and $\tilde\nu := {\nu +\mu(\G) \delta_{\hat s}}$ where measures $\tilde \mu, \tilde \nu$ have the same total mass $(\mu(\G) + \nu(\G))$. 
Let $\widetilde\Pi(\tilde \mu,\tilde \nu) := \Big\{ \tilde\gamma \in \mathcal{P}(\hat G \times \hat G) : \tilde \mu(U) =\tilde\gamma(U\times \hat \G),\, \tilde\nu(U)= \tilde\gamma(\hat \G\times U) \mbox{ for all Borel sets } U\subset \hat \G\Big\}$, then following~\citep[Lemma A.8]{le2023scalable}, we have
\begin{eqnarray}\label{eq:ET_W_tildec}
    \mathrm{ET}_{\lambda}(\mu, \nu) = \calW_{\tilde c}(\tilde \mu,\tilde \nu),
\end{eqnarray}
where $\calW_{\tilde c}(\tilde \mu,\tilde \nu)$ is a standard complete OT between two balanced measures $\tilde \mu$ and $\tilde \nu$ (i.e., having the same total mass $\mu(\G) + \nu(\G)$), with cost $\tilde c$, defined as
\[
\calW_{\tilde c}(\tilde \mu,\tilde \nu) := \inf_{\tilde \gamma \in \widetilde\Pi(\tilde\mu,\tilde \nu)}  \int_{\hat \G\times \hat \G} \tilde c(x,y) \tilde\gamma(\dd x, \dd y).
\]
Moreover, from Equation~\eqref{eq:ET_W_tildec}, we have
\begin{eqnarray}
    \mathrm{ET}_{\lambda}(\mu, \nu) &&= \calW_{\tilde c}(\tilde \mu,\tilde \nu) \nonumber\\
    &&= (\mu(\G) + \nu(\G)) \, \calW_{\tilde c}(\hat \mu,\hat \nu) \\
    &&= (\mu(\G) + \nu(\G)) \, (\calW_{\hat c}(\hat \mu,\hat \nu - b\lambda) \label{eq:tmp1_ET_KT}\\
    &&= \mathrm{KT}(\mu, \nu), \nonumber
\end{eqnarray}
where the equality in Equation~\eqref{eq:tmp1_ET_KT} is due to $\tilde c(x, y) = \hat c(x,y) - b\lambda$ for all $x, y \in \hat \G$.

Hence, the proof is completed.
\end{proof}

\subsubsection{Proof for Proposition~\ref{prop:monotonicity_OT}}\label{appsubsec:Monotonicity_OT}
\begin{proof}
    From the definition, we have
    \begin{equation}
\mathcal{A}(t; \hat \mu, \hat \nu) := \inf_{\tilde \gamma \in \Pi(\hat \mu, \hat \nu)} \int_{\hat \G \times \hat \G} \Phi\left(\frac{\hat{c}(x, y)}{t}\right) \dd\tilde\gamma(x, y).
\end{equation}
Let $0 < t_1 \le t_2 < \infty$, denote $\widetilde{\gamma^*_{t_1}}, \widetilde{\gamma^*_{t_2}}$ as the optimal transport plans of $\mathcal{A}(t_1; \hat \mu, \hat \nu), \mathcal{A}(t_2; \hat \mu, \hat \nu)$ respectively. Then, we have
\begin{align*}
    \mathcal{A}(t_2; \hat \mu, \hat \nu) &=  \int_{\hat \G \times \hat \G} \Phi\left(\frac{\hat{c}(x, y)}{t_2}\right) \dd\widetilde{\gamma^*_{t_2}}(x, y) \\
    &\le \int_{\hat \G \times \hat \G} \Phi\left(\frac{\hat{c}(x, y)}{t_2}\right) \dd\widetilde{\gamma^*_{t_1}}(x, y) \\
    &\le \int_{\hat \G \times \hat \G} \Phi\left(\frac{\hat{c}(x, y)}{t_1}\right) \dd\widetilde{\gamma^*_{t_1}}(x, y) \\
    &= \mathcal{A}(t_1; \hat \mu, \hat \nu),
\end{align*}
where the second inequality is due to the strictly increasing property of the $N$-function $\Phi$. 

Hence, the proof is completed.
\end{proof}

\subsubsection{Proof for Proposition~\ref{prop:monotonicity_regOT}}\label{appsubsec:Monotonicity_regOT}

\begin{proof}
    
The result is followed by the same reasoning as in the proof for Proposition~\ref{prop:monotonicity_OT} where we leverage the strictly increasing property of the $N$-function $\Phi$ and the optimal transport plans for $\mathcal{A}_{\varepsilon}$. 

More concretely, let $0 < t_1 \le t_2 < \infty$, denote $\widetilde{\gamma^*_{t_1}}, \widetilde{\gamma^*_{t_2}}$ as the optimal transport plans of $\mathcal{A}_{\varepsilon}(t_1; \hat \mu, \hat \nu), \mathcal{A}_{\varepsilon}(t_2; \hat \mu, \hat \nu)$ respectively. Then, we have
\begin{align*}
    \mathcal{A}_{\varepsilon}(t_2; \hat \mu, \hat \nu) &=  \int_{\hat \G \times \hat \G} \Phi\left(\frac{\hat{c}(x, y)}{t_2}\right) \dd\widetilde{\gamma^*_{t_2}}(x, y) - \varepsilon H(\widetilde{\gamma^*_{t_2}}) \\
    &\le \int_{\hat \G \times \hat \G} \Phi\left(\frac{\hat{c}(x, y)}{t_2}\right) \dd\widetilde{\gamma^*_{t_1}}(x, y) - \varepsilon H(\widetilde{\gamma^*_{t_1}}) \\
    &\le \int_{\hat \G \times \hat \G} \Phi\left(\frac{\hat{c}(x, y)}{t_1}\right) \dd\widetilde{\gamma^*_{t_1}}(x, y) - \varepsilon H(\widetilde{\gamma^*_{t_1}}) \\
    &= \mathcal{A}_{\varepsilon}(t_1; \hat \mu, \hat \nu).
\end{align*}
Hence, the proof is completed.
\end{proof}

\subsubsection{Proof for Proposition~\ref{prop:limits_regOT}}\label{appsubsec:limits_regOT}

\begin{proof}
We provide the proof for the lower and upper limits for $\mathcal{A}_{\varepsilon}$ as follows:

\textbf{For lower limit.} From the definition in Equation~\eqref{eq:regOW_t}, we have
\begin{equation*}
\mathcal{A}_{\varepsilon}\left(\frac{L_{\hat \mu, \hat \nu}}{\Phi^{-1}(1 + \varepsilon)}; \hat \mu, \hat \nu \right) = \inf_{\tilde \gamma \in \Pi(\hat \mu, \hat \nu)} \left[ \int_{\hat \G \times \hat \G} \Phi\left(\frac{\hat{c}(x, y)}{\frac{L_{\hat \mu, \hat \nu}}{\Phi^{-1}(1 + \varepsilon)}}\right) \dd\tilde\gamma(x, y) - \varepsilon H(\tilde\gamma) \right].
\end{equation*}
Additionally, since $N$-function $\Phi$ is strictly increasing, we have
\begin{equation*}
\Phi\left(\frac{\hat{c}(x, y)}{\frac{L_{\hat \mu, \hat \nu}}{\Phi^{-1}(1 + \varepsilon)}}\right) \le \Phi(\Phi^{-1}(1 + \varepsilon)) = 1 + \varepsilon.
\end{equation*}
For convenience, given any $\tilde \gamma \in \Pi(\hat \mu, \hat \nu)$, we define
\begin{equation}\label{eq:entropy_def}
\bar{\mathcal{H}}(\gamma) := -\int_{\hat\G \times \hat \G} \log\tilde\gamma(x, y)\dd\tilde\gamma(x, y).
\end{equation}

From the definition of $H$ in Proposition~\ref{prop:monotonicity_regOT}, for any $\tilde \gamma \in \Pi(\hat \mu, \hat \nu)$, we have
\begin{equation*}
    H(\tilde \gamma) = -\int_{\hat\G \times \hat \G} \log\tilde\gamma(x, y)\dd\tilde\gamma(x, y) + 1 \ge 1,
\end{equation*}
where the inequality is followed by using \citep[Lemma 2.1.1]{cover1999elements} (i.e., $\bar{\mathcal{H}}(\gamma) \ge 0$). 

Thus, we have
\begin{equation}
\mathcal{A}_{\varepsilon}\left(\frac{L_{\hat \mu, \hat \nu}}{\Phi^{-1}(1 + \varepsilon)}; \hat \mu, \hat \nu \right) \le (1 + \varepsilon) - \varepsilon \le 1.
\end{equation}
The proof for the lower limit is completed.

\paragraph{For upper limit.} For any $\tilde \gamma \in \Pi(\hat \mu, \hat \nu)$, we have
\begin{align*}
\mathcal{T} &:= \int_{\hat \G \times \hat \G} \Phi\left(\frac{\hat{c}(x, y)}{t}\right) \dd\tilde\gamma(x, y) - \varepsilon H(\tilde\gamma) \\
&\ge \Phi \left( \int_{\hat \G \times \hat \G} \left(\frac{\hat{c}(x, y)}{t}\right) \dd\tilde\gamma(x, y)\right) - \varepsilon H(\tilde\gamma) \\
&= \Phi \left( \frac{1}{t} \int_{\hat \G \times \hat \G} \hat{c}(x, y) \dd\tilde\gamma(x, y) \right) - \varepsilon H(\tilde\gamma),
\end{align*}
where we use the Jensen's inequality for the second row.

Additionally, for any $\tilde \gamma \in \Pi(\hat \mu, \hat \nu)$, we have
\begin{align*}
 H(\tilde\gamma) &= \bar{\mathcal{H}}(\tilde\gamma) + 1  \\
 &\le \bar{\mathcal{H}}(\hat\mu) + \bar{\mathcal{H}}(\hat\nu) + 1 \\
 &= H(\hat \mu) + H(\hat \nu) - 1,
\end{align*}
where we apply \citep[Theorem 2.2.1 and Theorem 2.6.5]{cover1999elements} for the inequality in the second row.

Thus, we have
\begin{equation}\label{eq:upperbound_objFunc_OT}
\mathcal{T} \ge \Phi \left( \frac{1}{t} \int_{\hat \G \times \hat \G} \hat{c}(x, y) \dd\tilde\gamma(x, y) \right) - \varepsilon \left( H(\hat \mu) + H(\hat \nu) - 1 \right)
\end{equation}
Taking the infimum of $\tilde \gamma$ in $\Pi(\hat \mu, \hat \nu)$, we obtain
\begin{align}
\mathcal{A}_{\varepsilon}\left(t; \hat \mu, \hat \nu \right) \ge \Phi \left( \frac{1}{t} \mathcal{W}_{\hat c}(\hat\mu, \hat\nu) \right) - \varepsilon \left( H(\hat \mu) + H(\hat \nu) - 1 \right)
\end{align}

Therefore, by choosing $t = \frac{\calW_{\hat c}(\hat \mu, \hat \nu)}{\Phi^{-1}(1 + \varepsilon\left[ H(\hat \mu) + H(\hat \nu) - 1 \right])}$, then we have
\[
\mathcal{A}_{\varepsilon} \! \left(\frac{\calW_{\hat c}(\hat \mu, \hat \nu)}{\Phi^{-1}(1 + \varepsilon\left[ H(\hat \mu) + H(\hat \nu) - 1 \right])}; \hat \mu, \hat \nu\right) \ge 1.
\]
The proof for the upper limit is completed.

\end{proof}

\subsubsection{Proof for Theorem~\ref{thm:OST_computation}}\label{appsubsec:thm:OST_computation}

\begin{proof}
    For $f \in \OrliczSobolevPhi(\G, \omega)$, as in Equation~\eqref{eq:OrliczSobolevFunction}, we have
    \[
    f(x) = f(z_0) + \int_{[z_0,x]} f'(y) \omega(\mathrm{d}y),  \quad \forall x\in \G.
    \]
    Thus, following the Definition~\ref{def:OST}, we have
    \begin{eqnarray}
    &\hspace{-16em} \calOS_{\Phi, \alpha}(\mu,\nu ) = \sup_{f(z_0) \in \calI_{\alpha}} f(z_0) (\mu(\G) - \nu(\G)) + \nonumber \\
     &\hspace{8em} \sup_{f \in \OrliczSobolevPsi(\G, \omega), \norm{f'}_{L_{\Psi}} \le b} \int_{\G} \left( \int_{[z_0,x]} f'(y) \omega(\mathrm{d}y) \right) \left(\mu(x) - \nu(x)\right) \dd x \nonumber
    \end{eqnarray}
    Thus, we can rewrite $\calOS_{\Phi, \alpha}(\mu,\nu )$ as follows:
    \begin{align}\label{equ:OST_tmp1}
    \hspace{-1em} \calOS_{\Phi, \alpha}(\mu,\nu ) &= \sup_{f \in \OrliczSobolevPsi(\G, \omega), \norm{f'}_{L_{\Psi}} \le b} \int_{\G} \left( \int_{[z_0,x]} f'(y) \omega(\mathrm{d}y) \right) \left(\mu(x) - \nu(x)\right) \dd x + \Theta |\mu(\G)-\nu(\G)| \nonumber \\
    &= \sup_{f \in \OrliczSobolevPsi(\G, \omega), \norm{f'}_{L_{\Psi}} \le b} \int_{\G} \left( \int_{[z_0,x]} f'(y) \omega(\mathrm{d}y) \right) \left(\mu(x) - \nu(x)\right) \dd x + \Theta |\mu(\G)-\nu(\G)|,
    \end{align}
    where recall that $\Theta$ is defined in Equation~\eqref{def:Theta}. 
    
    Additionally, recall that the indicator function of the shortest path $[z_0,x]$ is as follows:
\begin{equation}\label{eq:Indicator}
{\bf{1}}_{[z_0,x]}(y) = 
  \begin{cases} 
   1 & \text{if } y\in [z_0,x] \\
   0 & \text{otherwise}.
  \end{cases}
\end{equation}
We rewrite the objective function for the first term of $\calOS_{\Phi, \alpha}(\mu,\nu )$ in Equation~\eqref{equ:OST_tmp1} as follows:
\begin{align}
    \int_{\G} \left( \int_{[z_0,x]} f'(y) \omega(\mathrm{d}y) \right) \left(\mu(x) - \nu(x)\right) \dd x \nonumber &= \int_\G  \int_{\G}  {\bf{1}}_{[z_0,x]}(y) \, f'(y) \left(\mu(x) - \nu(x)\right) \omega(\mathrm{d}y) \mathrm{d}x \\
    &= \int_\G  \left[ \int_{\G}  {\bf{1}}_{[z_0,x]}(y) \,  \left(\mu(x) - \nu(x)\right)  \mathrm{d}x \right] f'(y)\omega(\mathrm{d}y) \\
    &= \int_\G  \left[ \mu(\Lambda(y)) -  \nu(\Lambda(y)) \right] f'(y)\omega(\mathrm{d}y),
\end{align}
where we apply the Fubini's theorem to interchange the order of integration for the second row, and use the definition of $\Lambda$ in Equation~\eqref{sub-graph} for the last row. Consequently, following~\citep[Proposition 10, pp.81]{rao1991theory} and notice that $\norm{bf'}_{L_{\Psi}} = b \norm{f'}_{L_{\Psi}}$ for $b>0$, we have
\begin{align}\label{eq:OST_1st_term}
    & \hspace{-4em} \sup_{f \in \OrliczSobolevPsi(\G, \omega), \norm{f'}_{L_{\Psi}} \le b} \int_{\G} \left( \int_{[z_0,x]} f'(y) \omega(\mathrm{d}y) \right) \left(\mu(x) - \nu(x)\right) \dd x \nonumber \\
    &\hspace{4em} = \sup_{f \in \OrliczSobolevPsi(\G, \omega), \norm{\frac{1}{b}f'}_{L_{\Psi}} \le 1} \int_\G  b \left[ \mu(\Lambda(y)) -  \nu(\Lambda(y)) \right] \left[\frac{1}{b}f'(y) \right]\omega(\mathrm{d}y) \nonumber \\
    &\hspace{4em} = \norm{\tilde{f}}_{\Phi},
\end{align}
where $\tilde{f}(x) := b\left(\mu(\Lambda(x)) - \nu(\Lambda(x))\right), \forall x \in \G$, and we write $\norm{\tilde{f}}_{\Phi}$ for the Orlicz norm of $\tilde{f}$ with $N$-function $\Phi$~\citep[Definition~2, pp.58]{rao1991theory} (i.e., see a review in Equation~\eqref{eq:OrliczNorm} in \S\ref{appsubsec:Orlicz_functions}). 

Moreover, following~\citep[Theorem 13, pp.69]{rao1991theory}, we also have 
\begin{equation}\label{eq:Orlicz_Amemiyanorm}
\|\tilde{f}\|_{\Phi}  =  
\inf_{k > 0} \frac{1}{k}\left( 1 + \int_{\G} \Phi\left(k \left| \tilde{f}(x) \right|\right) \omega(\text{d}x) \right).
\end{equation}

Hence, putting these Equations~\eqref{equ:OST_tmp1},~\eqref{eq:OST_1st_term},~\eqref{eq:Orlicz_Amemiyanorm} together, we have
\begin{eqnarray}
    \calOS_{\Phi, \alpha}(\mu,\nu ) = \inf_{k > 0} \frac{1}{k}\left( 1 + \int_{\G} \Phi\left(k b \left| \mu(\Lambda(x)) - \nu(\Lambda(x)) \right|\right) \omega(\text{d}x) \right) + \Theta |\mu(\G)-\nu(\G)|. 
\end{eqnarray}
The proof is completed.

\end{proof}

\subsubsection{Proof for Corollary~\ref{cor:OST_1d_optimization_discrete}}\label{app:subsec:cor:OST_1d_optimization_discrete}
\begin{proof}
Following Theorem~\ref{thm:OST_computation}, we have
\begin{eqnarray}\label{eq:fromTheorem_OST_computation}
    \calOS_{\Phi, \alpha}(\mu,\nu ) =  \Theta |\mu(\G)-\nu(\G)| + \inf_{k > 0} \frac{1}{k}\left( 1 + \hspace{-0.3em} \int_{\G} \hspace{-0.3em} \Phi\left(kb \left| \mu(\Lambda(x)) - \nu(\Lambda(x)) \right|\right) \omega(\text{d}x) \right), 
\end{eqnarray}

We next follow the same reasoning as in~\citep[Corollary 3.4]{le2024generalized} to compute the integral in~\eqref{eq:fromTheorem_OST_computation} by an explicit expression. 

For an edge $e$ between two nodes $u, v \in V$ of graph $\G$, then $u, v$ are also two data points in $\R^n$ as $\G$ is a physical graph. For convenience, denote $\langle u,  v\rangle$ as the line segment in $\R^n$ connecting the two data points $u, v$, and $(u, v)$ as the same line segment but without its two end-points. Therefore, we have $e=\langle u,v\rangle$. 

Additionally, for any $x\in (u,v)$, we have $y\in \G\setminus (u,v)$ belongs to $\Lambda(x)$ if and only if $y\in \gamma_e$ (see Equation~\eqref{sub-graph} for the definitions of $\Lambda(x)$ and $\gamma_e$). Thus, we have 
\begin{equation}\label{eq:relation_Lambda_gamma}
\Lambda(x)\setminus (u,v) =\gamma_e.
\end{equation}

Consider the case where $\omega$ is the length measure of graph $\G$, we have $\omega(\{x\})= 0$ for every $x\in\G$. Consequently,
\begin{equation}\label{eq:int_sum_edges}
\int_{\G} \hspace{-0.3em} \Phi\left(kb \left| \mu(\Lambda(x)) - \nu(\Lambda(x)) \right|\right) \omega(\text{d}x) = \sum_{e=\langle u,v\rangle\in E}   \int_{(u,v)} \hspace{-0.3em} \Phi\left(kb \left| \mu(\Lambda(x)) - \nu(\Lambda(x)) \right|\right) \omega(\text{d}x).
\end{equation}

Additionally, for measures $\mu, \nu$ supported on nodes $V$ of $\G$, and using Equation~\eqref{eq:relation_Lambda_gamma}, then we have
\[
\left| \mu(\Lambda(x)) -  \nu(\Lambda(x)) \right| = \left| \mu(\Lambda(x)\setminus (u,v)) -  \nu(\Lambda(x)\setminus (u,v)) \right| = \left| \mu(\gamma_e) - \nu(\gamma_e) \right|,
\]
for every edge $e=\langle u,v\rangle\in E$ of graph $\G$.

Therefore, we can rewrite the identity~\eqref{eq:int_sum_edges} as follows:
\begin{align*}
\int_{\G} \hspace{-0.3em} \Phi\left(kb \left| \mu(\Lambda(x)) - \nu(\Lambda(x)) \right|\right) \omega(\text{d}x) &= \sum_{e=\langle u,v\rangle\in E}   \int_{(u,v)} \hspace{-0.3em} \Phi\left(kb \left| \mu(\gamma_e) - \nu(\gamma_e) \right|\right) \omega(\text{d}x). \\
&= \sum_{e=\langle u,v\rangle\in E} \Phi\left(kb \left| \mu(\gamma_e) - \nu(\gamma_e) \right|\right) \int_{(u,v)} \omega(\text{d}x)\\
&= \sum_{e \in E} w_e \, \Phi\left(kb \left| \mu(\gamma_e) - \nu(\gamma_e) \right|\right). \label{app:eq:sum_edge_on_graph}
\end{align*}
By combining it with \eqref{eq:fromTheorem_OST_computation}, we obtain
\begin{equation*}
    \calOS_{\Phi, \alpha}(\mu,\nu ) =  \Theta |\mu(\G)-\nu(\G)| + \sum_{e \in E} w_e \, \Phi\left(kb \left| \mu(\gamma_e) - \nu(\gamma_e) \right|\right).
\end{equation*}
Hence, the proof is completed. 
\end{proof}

\subsubsection{Proof for Proposition~\ref{prop:OST_geodesic_space}}\label{app:subsec:prop:OST_geodesic_space}

\begin{proof}
The proof for each property on geometric structure of OST is as follows:

i) The result is directly followed from Equation~\eqref{equ:OST} in Theorem~\ref{thm:OST_computation} with the observation that 
\[
|\mu(\G)-\nu(\G)| = |(\mu + \sigma)(\G)-(\nu + \sigma)(\G)|
\]
and 
\[
\left| \mu(\Lambda(x)) - \nu(\Lambda(x)) \right| = \left| (\mu + \sigma)(\Lambda(x)) - (\nu + \sigma)(\Lambda(x)) \right|.
\] 

ii) From Definition~\ref{def:OST}, choosing $f=0$, then $f \in \mathbb{U}_{\Psi, \alpha}$, and for any $\mu, \nu \in \calP(\G)$, we have that 
\[
\calOS_{\Phi, \alpha}(\mu,\nu) \ge 0.
\]

Assume that $\calOS_{\Phi, \alpha}(\mu,\nu)(\mu,\nu)=0$. Then, from Theorem~\ref{thm:OST_computation}, we obtain
\[
 \Theta |\mu(\G)-\nu(\G)| + 
    \inf_{k > 0} \frac{1}{k}\left( 1 + \hspace{-0.3em} \int_{\G} \hspace{-0.3em} \Phi\left(kb \left| \mu(\Lambda(x)) - \nu(\Lambda(x)) \right|\right) \omega(\text{d}x) \right)  = 0.
\]
Additionally, for $0\leq \alpha< \frac{b\lambda}{2} +\min\{w_1(z_0), w_2(z_0)\}$, we have $\Theta > 0$. Consequently, we must have 
\[
\mu(\G) =\nu(\G)  \quad\mbox{and}\quad  \inf_{k > 0} \frac{1}{k}\left( 1 + \hspace{-0.3em} \int_{\G} \hspace{-0.3em} \Phi\left(kb \left| \mu(\Lambda(x)) - \nu(\Lambda(x)) \right|\right) \omega(\text{d}x) \right)
 =0.   
 \]
Thus, $\mu(\Lambda(x)) = \nu(\Lambda(x)), \forall x\in\G$. 

By applying~\citep[Lemma A.9]{le2023scalable},\footnote{In \S\ref{appsubsec:UST} (Lemma~\ref{lem:equal-measure}), we review the Lemma A.9 in~\citet{le2023scalable}.}  it leads to $\mu=\nu$. 

Moreover, from Definition~\ref{def:OST}, we also have $\calOS_{\Phi, \alpha}(\mu,\mu) = 0$.

Furthermore, for any feasible function $f \in \mathbb{U}_{\Psi, \alpha}$, we have
\begin{align*}
\int_\G f(x) \mu(\mathrm{d}x) - \int_\G f(x) \nu(\mathrm{d}x) &= \Big[ \int_\G f(x) \mu(\mathrm{d}x) - \int_\G f(x) \sigma(\mathrm{d}x)\Big]  + \\
& \hspace{12em} \Big[\int_\G f(x) \sigma(\mathrm{d}x) - \int_\G f(x) \nu(\mathrm{d}x)\Big]\\
&\leq \calOS_{\Phi, \alpha}(\mu,\sigma ) + \calOS_{\Phi, \alpha}(\sigma,\nu ).
\end{align*}
Therefore, by taking the infimum for $f \in \mathbb{U}_{\Psi, \alpha}$, it implies that
\[
\calOS_{\Phi, \alpha}(\mu,\nu )\leq \calOS_{\Phi, \alpha}(\mu,\sigma ) + \calOS_{\Phi, \alpha}(\sigma,\nu ).
\]
Hence, $\calOS_{\Phi, \alpha}$ satisfies the triangle inequality.

iii) With an additional assumption $w_1(z_0) = w_2(z_0)$, then for any function $f \in \mathbb{U}_{\Psi, \alpha}$, we also have $(-f) \in \mathbb{U}_{\Psi, \alpha}$.

Therefore, from Definition~\ref{def:OST}, we obtain $\calOS_{\Phi, \alpha}(\mu,\nu ) = \calOS_{\Phi, \alpha}(\nu,\mu)$. 

Thus, together with results in ii), we have $\calOS_{\Phi, \alpha}$ is a metric.


\end{proof}

\subsubsection{Proof for Proposition~\ref{prop:relation_OST_GST}}\label{app:subsec:prop:relation_OST_GST}

\begin{proof}
    For $\mu(\G) = \nu(\G)$ and $b = 1$, then following Theorem~\ref{thm:OST_computation} for OST and~\citep[Theorem 3.3]{le2024generalized} for GST, we have
    \[
    \calOS_{\Phi, \alpha}(\mu,\nu) = \inf_{k > 0} \frac{1}{k}\left( 1 + \hspace{-0.3em} \int_{\G} \hspace{-0.3em} \Phi\left(k \left| \mu(\Lambda(x)) - \nu(\Lambda(x)) \right|\right) \omega(\text{d}x) \right) =  \mathcal{GS}_{\Phi}(\mu,\nu).
    \]
\end{proof}
The proof is completed.

\subsubsection{Proof for Proposition~\ref{prop:relation_OST_ST}}\label{app:subsec:prop:relation_OST_ST}

\begin{proof}
For $\mu(\G) = \nu(\G)$, $b=1$, by applying Proposition~\ref{prop:relation_OST_GST}, we have
\begin{equation}\label{eq:OST-GST}
\calOS_{\Phi, \alpha}(\mu,\nu) =  \mathcal{GS}_{\Phi}(\mu,\nu),
\end{equation}
where we recall that $\mathcal{GS}_{\Phi}$ is the GST for balanced measures on a graph.

Additionally, for $1 < p < \infty$ and $N$-function $\Phi(t) = \frac{(p-1)^{p-1}}{p^p} t^p$, by leveraging~\citep[Proposition 4.4]{le2024generalized} for the connection between GST and ST, we have
\begin{equation}\label{eq:GST-ST}
\mathcal{GS}_{\Phi}(\mu,\nu) = \calS_{p}(\mu,\nu).
\end{equation}
Therefore, by combining Equations~\eqref{eq:OST-GST} and~\eqref{eq:GST-ST}, we obtain
\[
\calOS_{\Phi, \alpha}(\mu,\nu) = \calS_{p}(\mu,\nu).
\]
The proof is completed.
\end{proof}

\subsubsection{Proof for Proposition~\ref{prop:relation_OST_UST}}\label{app:subsec:prop:relation_OST_UST}

\begin{proof}
For $N$-function $\Phi(t) = \frac{(p-1)^{p-1}}{p^p} t^p$ with $1 < p < \infty$, from Theorem~\ref{thm:OST_computation}, we have
\begin{eqnarray}\label{eq:limT}
     &\hspace{-20em} \calOS_{\Phi, \alpha}(\mu,\nu ) =  \Theta |\mu(\G)-\nu(\G)| + \nonumber \\
     &\hspace{8em} \inf_{k > 0} \frac{1}{k}\left( 1 + \int_{\G} \frac{(p-1)^{p-1}}{p^p} k^p b^p \left| \mu(\Lambda(x)) - \nu(\Lambda(x)) \right|^p  \omega(\text{d}x) \right). 
\end{eqnarray}

For convenience, for $k>0$, let 
\[
T(k) := \frac{1}{k} + \frac{(p-1)^{p-1}}{p^p} k^{p-1} b^p \int_{\G} \left| \mu(\Lambda(x)) - \nu(\Lambda(x)) \right|^p \omega(\text{d}x),
\]
i.e., the objective function of the univariate optimization problem for $\calOS_{\Phi, \alpha}$.

We next consider two cases:

{\bf Case 1:} $\int_{\G} \left| \mu(\Lambda(x)) - \nu(\Lambda(x)) \right|^p \omega(\text{d}x) = 0$. Then, we have
\[
\inf_{k > 0} T(k) = \inf_{k > 0} \frac{1}{k} = 0.
\]
Consequently, from Equation~\eqref{eq:limT}, we have
\begin{align*}
\calOS_{\Phi, \alpha}(\mu,\nu ) &=  \Theta |\mu(\G)-\nu(\G)| = b\, \Big[\int_{\G} | \mu(\Lambda(x)) -  \nu(\Lambda(x))|^p \, \omega(\dd x)\Big]^\frac{1}{p} + \, \Theta |\mu(\G)-\nu(\G)| \\
&= \mathcal{US}_{p, \alpha}(\mu, \nu).
\end{align*}

{\bf Case 2:} $\int_{\G} \left| h(x) \right|^p \omega(\text{d}x) \neq 0$.
Then, we have  
\[
\lim_{k\to 0^+} T(k) = \lim_{k\to +\infty} T(k) = +\infty.
\]
Therefore, Equation~\eqref{eq:limT}, we have
\begin{align}\label{eq:OST_k0}
    \calOS_{\Phi, \alpha}(\mu,\nu ) =  \Theta |\mu(\G)-\nu(\G)| + T(k_0),
\end{align}
for some finite number $k_0 \in (0, +\infty)$ satisfying $T'(k_0) =0$.

Additionally, we have
\[
T'(k)= -\frac{1}{k^2} + \Big(\frac{p-1}{p}\Big)^p k^{p-2} b^p \int_{\G} \left| \mu(\Lambda(x)) - \nu(\Lambda(x)) \right|^p \omega(\text{d}x).
\]
Consequently, by solving the equation $T'(k_0) =0$ w.r.t. $k_0$, we obtain
\begin{align*}\label{eq:k_opt}
k_0 = \frac{1}{ \frac{p-1}{p} b \left(\int_{\G} \left| \mu(\Lambda(x)) - \nu(\Lambda(x)) \right|^p \omega(\text{d}x)\right)^{\frac{1}{p}} }. 
\end{align*}

Therefore, by plugging this value of $k_0$ into $T$, we have
\begin{align*}
\hspace{-1em} T(k_0) &= \frac{1}{k_0}\left( 1  + \frac{(p-1)^{p-1}}{p^p} k_0^p b^p \int_{\G} \left| \mu(\Lambda(x)) - \nu(\Lambda(x)) \right|^p \omega(\text{d}x) \right)\\
&= \frac{p-1}{p} b \left(\int_{\G} \left| \mu(\Lambda(x)) - \nu(\Lambda(x)) \right|^p \omega(\text{d}x)\right)^{\frac{1}{p}} \times \\
& \hspace{0.2em} \left( 1 + \frac{(p-1)^{p-1}}{p^p} \frac{1}{ \frac{(p-1)^p}{p^p}  b^p \left(\int_{\G} \left| \mu(\Lambda(x)) - \nu(\Lambda(x)) \right|^p \omega(\text{d}x)\right) } b^p \int_{\G} \left| \mu(\Lambda(x)) - \nu(\Lambda(x)) \right|^p \omega(\text{d}x) \right) \\
&= b \left(\int_{\G} \left| \mu(\Lambda(x)) - \nu(\Lambda(x)) \right|^p \omega(\text{d}x)\right)^{\frac{1}{p}}.
\end{align*}
Thus, by plugging this value of $T(k_0)$ into Equation~\eqref{eq:OST_k0}, we obtain
\begin{align*}
\calOS_{\Phi, \alpha}(\mu,\nu ) &=  \Theta |\mu(\G)-\nu(\G)| + b \left(\int_{\G} \left| \mu(\Lambda(x)) - \nu(\Lambda(x)) \right|^p \omega(\text{d}x)\right)^{\frac{1}{p}} \\
&= \mathcal{US}_{p, \alpha}(\mu, \nu).
\end{align*}

Hence, we have shown that $\calOS_{\Phi, \alpha}(\mu,\nu )  = \mathcal{US}_{p, \alpha}(\mu, \nu)$ in both cases. 

The proof is completed.

\end{proof}

\subsubsection{Proof for Proposition~\ref{prop:limit_OST}}\label{app:subsec:prop:limit_OST}

\begin{proof}
Following Corollary~\ref{cor:OST_1d_optimization_discrete}, we have
\begin{eqnarray*}
    \calOS_{\Phi, \alpha}(\mu,\nu ) =  \Theta |\mu(\G)-\nu(\G)| + \inf_{k > 0} \frac{1}{k}\left( 1 + \hspace{-0.3em} \sum_{e \in E} \hspace{-0.1em} w_e \Phi\!\left(kb \left| \mu(\gamma_{e}) - \nu(\gamma_{e}) \right|\right) \right). 
\end{eqnarray*}
For $\Phi(t) = t$, then we have
\begin{align*}
    \calOS_{\Phi, \alpha}(\mu,\nu ) &=  \Theta |\mu(\G)-\nu(\G)| + \inf_{k > 0} \frac{1}{k}\left( 1 + \hspace{-0.3em} \sum_{e \in E} \hspace{-0.1em} w_e kb \left| \mu(\gamma_{e}) - \nu(\gamma_{e}) \right|\right) \\
    &=  \Theta |\mu(\G)-\nu(\G)| + \inf_{k > 0} \frac{1}{k} + \sum_{e \in E} w_e b \left| \mu(\gamma_{e}) - \nu(\gamma_{e}) \right| \\
    &=  \Theta |\mu(\G)-\nu(\G)| + b\sum_{e \in E} w_e \left| \mu(\gamma_{e}) - \nu(\gamma_{e}) \right|.
\end{align*}

Hence, the proof is completed.
\end{proof}

\subsubsection{Proof for Proposition~\ref{prop:limit_OrliczEPT}}\label{app:subsec:prop:limit_OrliczEPT}

\begin{proof}
From Equation~\eqref{eq:OrliczEPT}, we have
\begin{eqnarray*}
\calOE_{\Phi}(\mu, \nu) = \left(\mu(\G) + \nu(\G)\right)(\calW_{\Phi}(\hat \mu,\hat \nu) - b\lambda).
\end{eqnarray*}
For $\Phi(t) = t$, we further have
\[
\calW_{\Phi}(\hat \mu,\hat \nu) = \inf_{\tilde \gamma \in \Pi(\hat \mu, \hat \nu)} \inf \Big[ t > 0 : \int_{\hat \G \times \hat \G} \left(\frac{\hat{c}(x, y)}{t}\right) \text{d}\tilde\gamma(x, y) \le 1\Big]
\]
Then, the infimum $(t^*, \tilde{\gamma}^*)$ satisfies
\[
\int_{\hat \G \times \hat \G} \left(\frac{\hat{c}(x, y)}{t^*}\right) \text{d}\tilde\gamma^{*}(x, y) = 1.
\]
Therefore, we obtain $t^* = \int_{\hat \G \times \hat \G} \hat{c}(x, y) \text{d}\tilde\gamma^{*}(x, y) = \calW_{\hat c}(\hat \mu,\hat \nu)$.

Hence, we have
\begin{align*}
\calOE_{\Phi}(\mu, \nu) &= \left(\mu(\G) + \nu(\G)\right)\left(\calW_{\hat c}(\hat \mu,\hat \nu) - b\lambda\right) \\
&= \mathrm{KT}(\mu, \nu).
\end{align*}

The proof is completed.

\end{proof}

\subsubsection{Proof for Proposition~\ref{prop:limit_OST_OrliczEPT}}\label{app:subsec:prop:limit_OST_OrliczEPT}

\begin{proof}

For $\Phi(t) = t$, $p = 1$, from Theorem~\ref{thm:OST_computation}, we have
\begin{align*}
    \calOS_{\Phi, \alpha}(\mu,\nu ) &=  \Theta |\mu(\G)-\nu(\G)| + \inf_{k > 0} \frac{1}{k}\left( 1 +  \int_{\G}  kb \left| \mu(\Lambda(x)) - \nu(\Lambda(x)) \right| \omega(\text{d}x) \right) \\
    &=  \Theta |\mu(\G)-\nu(\G)| + \inf_{k > 0} \frac{1}{k} + b \int_{\G}  \left| \mu(\Lambda(x)) - \nu(\Lambda(x)) \right| \omega(\text{d}x) \\
    &=  \Theta |\mu(\G)-\nu(\G)| + b \int_{\G}  \left| \mu(\Lambda(x)) - \nu(\Lambda(x)) \right| \omega(\text{d}x) \\
    &= \mathcal{US}_{p, \alpha}(\mu, \nu).
\end{align*}
Additionally, for $\Phi(t) = t$, from Proposition~\ref{prop:limit_OrliczEPT}, we have
\[
 \calOE_{\Phi}(\mu, \nu) = \mathrm{KT}(\mu, \nu).
\]
With additional assumptions that $\lambda \geq 0$ and the nonnegative weight functions $w_1, w_2$ are $b$-Lipschitz w.r.t. $d_\G$, then by applying~\citep[Corollary 3.2]{le2023scalable}, we have
\[
 \calOE_{\Phi}(\mu, \nu) = \mathrm{KT}(\mu, \nu) = \mathrm{ET}_\lambda(\mu,\nu).
\]
Consequently, for $\alpha = 0$, and the length measure $\omega$ on $\G$, then following~\citep[Proposition 5.2]{le2023scalable}, we have
\[
\calOS_{\Phi, \alpha}(\mu,\nu ) \ge \calOE_{\Phi}(\mu, \nu) + \frac{b\lambda}{2}(\mu(\G) + \nu(\G)). 
\]

The proof is completed.
\end{proof}

\subsubsection{Proof for Proposition~\ref{prop:limit_OST_dalpha}}\label{app:subsec:prop:limit_OST_dalpha}

\begin{proof}
From Proposition~\ref{prop:limit_OST}, we have
\begin{align}\label{eq:OST_UST_1}
    \calOS_{\Phi, \alpha}(\mu,\nu ) &=  \Theta |\mu(\G)-\nu(\G)| + b\sum_{e \in E} w_e \left| \mu(\gamma_{e}) - \nu(\gamma_{e}) \right| \\
    &= \mathcal{US}_{1, \alpha}(\mu, \nu).
\end{align}
For the case when $\G$ is a tree, then following~\citep[Proposition 5.3 i)]{le2023scalable}, we further have
\begin{align}\label{eq:UST_dalpha_1}
\mathcal{US}_{1, \alpha}(\mu, \nu) = d_{\alpha}(\mu, \nu).
\end{align}
Thus, from Equations~\eqref{eq:OST_UST_1} and~\eqref{eq:UST_dalpha_1}, we have
\[
\calOS_{\Phi, \alpha}(\mu,\nu ) = d_{\alpha}(\mu, \nu).
\]

The proof is completed.

\end{proof}

\subsubsection{Proof for Proposition~\ref{prop:limit_OST_OT}}\label{app:subsec:prop:limit_OST_OT}

\begin{proof}
From Equation~\eqref{eq:OST_UST_1} in the proof of Proposition~\ref{prop:limit_OST_dalpha}, we have
\begin{align}\label{eq:OST_UST_1_new}
    \calOS_{\Phi, \alpha}(\mu,\nu ) = \mathcal{US}_{1, \alpha}(\mu, \nu).
\end{align}
Additionally, when $\G$ is a tree, and with an additional assumption that $\mu(\G) = \nu(\G)$, by applying~\citep[Proposition 5.3 ii)]{le2023scalable}, and notice that $p=1$ and $b=1$, we obtain
\begin{equation}\label{eq:UST_OT_1}
  \mathcal{US}_{1, \alpha}(\mu, \nu) = \calW_{d_{\G}}(\mu, \nu),
\end{equation}
where recall that $\calW_{d_{\G}}$ is the standard optimal transport with graph metric ground cost $d_{\G}$.

Hence, from Equations~\eqref{eq:OST_UST_1_new} and~\eqref{eq:UST_OT_1}, we get
\[
\calOS_{\Phi, \alpha}(\mu,\nu ) = \calW_{d_{\G}}(\mu, \nu).
\]

The proof is completed.

\end{proof}

\subsubsection{Proof for Proposition~\ref{rm:upperlimit_regOT}}\label{appsubsec:rm:upperlimit_regOT}

\begin{proof}
Following~\citep[Theorem 2.2.1]{cover1999elements} and definition of conditional entropy~\citep[Equation 2.10]{cover1999elements}, for any $\tilde \gamma \in \Pi(\hat \mu, \hat \nu)$, we have
\begin{align}
\bar{\mathcal{H}}(\tilde\gamma) &\ge \frac{1}{2}(\bar{\mathcal{H}}(\hat\mu) + \bar{\mathcal{H}}(\hat\nu)) \\
\bar{\mathcal{H}}(\tilde\gamma) + 1 &\ge \frac{1}{2}(\bar{\mathcal{H}}(\hat\mu) + \bar{\mathcal{H}}(\hat\nu)) + 1 \\
H(\tilde\gamma) &\ge \frac{1}{2}(H(\hat\mu) + H(\hat\nu)),
\end{align}
where we recall that $\bar{\mathcal{H}}$ and $H$ are defined in Equation~\eqref{eq:entropy_def} and in Proposition~\ref{prop:monotonicity_regOT} respectively.

Therefore, as in the proof for Proposition~\ref{prop:limits_regOT} in \S\ref{appsubsec:limits_regOT}, from Equation~\eqref{eq:upperbound_objFunc_OT}, we have
\begin{align}
    \mathcal{T} &\ge \Phi \left( \frac{1}{t} \int_{\hat \G \times \hat \G} \hat{c}(x, y) \dd\tilde\gamma(x, y) \right) - \varepsilon \left( H(\hat \mu) + H(\hat \nu) - 1 \right) \\
    &\ge \Phi \left( \frac{1}{t} \left[ \int_{\hat \G \times \hat \G} \hat{c}(x, y) \dd\tilde\gamma(x, y) - \varepsilon H(\tilde\gamma) + \frac{\varepsilon}{2}\left(H(\hat\mu) + H(\hat\nu)\right) \right] \right) - \varepsilon \left( H(\hat \mu) + H(\hat \nu) - 1 \right) \label{eq:assumption_nonnegativity_Phi}
\end{align}
where we assume that the entropic regularized input of $N$-function $\Phi$ is nonnegative in the second row (Equation~\eqref{eq:assumption_nonnegativity_Phi}), i.e., 
\[
\int_{\hat \G \times \hat \G} \hat{c}(x, y) \dd\tilde\gamma(x, y) - \varepsilon H(\tilde\gamma) + \frac{\varepsilon}{2}\left(H(\hat\mu) + H(\hat\nu)\right) \ge 0,
\]
for any $\tilde \gamma \in \Pi(\hat \mu, \hat \nu)$.

Taking the infimum of $\tilde \gamma$ in $\Pi(\hat \mu, \hat \nu)$, we obtain
\begin{align}
\mathcal{A}_{\varepsilon}\left(t; \hat \mu, \hat \nu \right) \ge \Phi \left( \frac{1}{t} \left[ \mathcal{W}_{\varepsilon}(\hat\mu, \hat\nu) + \frac{\varepsilon} {2}\left(H(\hat\mu) + H(\hat\nu)\right) \right] \right) - \varepsilon \left( H(\hat \mu) + H(\hat \nu) - 1 \right)
\end{align}

Therefore, by choosing $t = \frac{\mathcal{W}_{\varepsilon}(\hat\mu, \hat\nu) + \frac{\varepsilon} {2}\left(H(\hat\mu) + H(\hat\nu)\right)}{\Phi^{-1}(1 + \varepsilon\left[ H(\hat \mu) + H(\hat \nu) - 1 \right])}$, then we have
\[
\mathcal{A}_{\varepsilon} \! \left(\frac{\mathcal{W}_{\varepsilon}(\hat\mu, \hat\nu) + \frac{\varepsilon} {2}\left(H(\hat\mu) + H(\hat\nu)\right)}{\Phi^{-1}(1 + \varepsilon\left[ H(\hat \mu) + H(\hat \nu) - 1 \right])}; \hat \mu, \hat \nu\right) \ge 1.
\]

The proof is completed.
\end{proof}

\subsubsection{Proof for Proposition~\ref{prop:limit_regOrliczEPT}}\label{app:subsec:prop:limit_regOrliczEPT}

\begin{proof}
We use the same reason as in the proof for Proposition~\ref{prop:limit_OrliczEPT}. From Equation~\eqref{eq:regOrliczEPT}, we have
\begin{eqnarray*}
\calOE_{\Phi, \varepsilon}(\mu, \nu) := \left(\mu(\G) + \nu(\G)\right)(\calW_{\Phi, \varepsilon}(\hat \mu,\hat \nu) - b\lambda).
\end{eqnarray*}
For $\Phi(t) = t$, we further have
\[
\calW_{\Phi, \varepsilon}(\hat \mu,\hat \nu) = \inf_{\tilde \gamma \in \Pi(\hat \mu, \hat \nu)} \inf \left[ t > 0 : \int_{\hat \G \times \hat \G} \left(\frac{\hat{c}(x, y)}{t}\right) \text{d}\tilde\gamma(x, y) - \varepsilon H(\tilde \gamma) \le 1\right]
\]
Then, let $\tilde{\gamma}^*_{\varepsilon}$ is the optimal solution for the entropic regularized OT
 \[
 \calW_{\hat{c},\varepsilon}(\hat \mu,\hat \nu) = \inf_{\tilde \gamma \in \Pi(\hat\mu,\hat \nu)} \left[  \int_{\hat \G\times \hat \G} \hat c(x,y) \tilde\gamma(\dd x, \dd y) - \varepsilon H(\tilde \gamma) \right].
 \]
 Thus, for the infimum $(t^*, \tilde{\gamma}^{*}_{\varepsilon})$, we have
\[
\int_{\hat \G \times \hat \G} \left(\frac{\hat{c}(x, y)}{t^*}\right) \text{d}\tilde\gamma^{*}_{\varepsilon}(x, y) = 1.
\]
Therefore, we obtain 
\[
t^* = \int_{\hat \G \times \hat \G} \hat{c}(x, y) \text{d}\tilde\gamma^{*}_{\varepsilon}(x, y) = \calW_{\hat c, \varepsilon}(\hat \mu,\hat \nu).
\]

Hence, we have
\[
\calOE_{\Phi, \varepsilon}(\mu, \nu) = \left(\mu(\G) + \nu(\G)\right)\left(\calW_{\hat c, \varepsilon}(\hat \mu,\hat \nu) - b\lambda\right).
\]

The proof is completed.
\end{proof}

\section{Brief Reviews, Further Discussions and Empirical Results}\label{appsec:further_results_discussions}

In this section, we give brief reviews on important related notions to our proposed approaches. We next give further discussions on several aspects, and provide further empirical results.

\subsection{Brief Reviews}\label{appsubsec:brief_reviews}

We provide brief reviews on important related notions to our proposed approaches.

\subsubsection{Sobolev Transport (ST)}\label{appsubsec:ST}

We briefly review main notions for Sobolev transport (ST)~\citep{le2022st} for probability measures on a graph.

\paragraph{$\boldsymbol{L^{p}}$ functional space.} For a nonnegative Borel measure $\omega$ on $\G$, denote $L^p( \G, \omega)$ as the space of all Borel measurable functions $f:\G\to \R$ such that $\int_\G |f(y)|^p \omega(\mathrm{d}y) <\infty$. For $p=\infty$, we instead assume that $f$ is bounded $\omega$-a.e. Then, $L^p( \G, \omega)$ is a normed space with the norm defined by
\[
\|f\|_{L^p(\G, \omega)} := \left(\int_\G |f(y)|^p \omega(\dd y)\right)^\frac{1}{p} \text{ for } 1\leq p < \infty,
\]
and for $p = \infty$,
\[
\|f\|_{L^{\infty}(\G, \omega)} := \inf\left\{t \in \R:\, |f(x)|\leq t \mbox{ for $\omega$-a.e. } x\in\G\right\}.
\]

Functions $f_1, f_2 \in L^p( \G, \omega)$ are considered to be the same if $f_1(x) =f_2(x)$ for $\omega$-a.e. $x\in\G$. 

\paragraph{Graph-based Sobolev space~\citep{le2022st}.} Let $\omega$ be a nonnegative Borel measure on $\G$, and let  $1\leq p\leq \infty$. A continuous function $f: \G \to \R$ is said to belong to the Sobolev space $W^{1,p}(\G, \omega)$ if there exists a  function $h\in L^p( \G, \omega) $ satisfying 
\begin{equation}\label{FTC}
f(x) -f(z_0) =\int_{[z_0,x]} h(y) \omega(\mathrm{d}y),  \quad \forall x\in \G.
\end{equation}
Such function  $h$ is unique in $L^p(\G, \omega) $ and is called the generalized graph derivative of $f$ w.r.t.~the measure $\omega$. The generalized graph derivative of $f \in W^{1,p}(\G, \omega)$ is denoted $f' \in L^p( \G, \omega)$.

\paragraph{Sobolev transport~\citep{le2022st}.} Let $\omega$ be a nonnegative Borel measure on $\G$. Given $1\leq p\leq \infty$, and let  $p'$ be its conjugate, i.e., the number $p'\in [1,\infty]$ satisfying $\frac1p +\frac{1}{p'}=1$. For probability measures $\mu, \nu$ supported on graph $\G$, the $p$-order Sobolev transport (ST)~\citep[Definition 3.2]{le2022st} is defined as 
\begin{equation} \label{eq:distance}
 \calS_p(\mu,\nu ) \! \coloneqq \! \left\{
\begin{array}{cl}
 \sup \left[\int_\G f(x) \mu(\mathrm{d}x) - \int_\G f(x) \nu(\mathrm{d}x)\right] \\
 ~~ \mathrm{s.t.} \, f \hspace{-0.2em} \in W^{1,p'} \hspace{-0.3em} (\G, \omega),  \, \|f'\|_{L^{p'}\hspace{-0.2em} (\G, \omega)}\leq 1,
\end{array}
\right.
\end{equation}
where we write $f'$ for the generalized graph derivative of $f$, $W^{1,p'} \hspace{-0.3em} (\G, \omega)$ for the graph-based Sobolev space on $\G$, and $L^{p'}\hspace{-0.2em} (\G, \omega)$ for the $L^p$ functional space on $\G$.

\subsubsection{Length measure}

We briefly review the length measure on graph $\G$ in~\citep{le2022st}.

\begin{definition}[Length measure~\citep{le2022st}] \label{def:measure} 
Let $ \omega^*$ be the unique Borel measure on $\G$ such that the restriction of $\omega^*$ on any edge is the length measure of that edge. That is, $\omega^*$  satisfies:
\begin{enumerate}
\item[i)] For  any edge $e$ connecting two nodes $u$ and $v$, we have 
 $\omega^*(\langle x,y\rangle) = (t-s) w_e$ 
 whenever $x = (1-s) u + s v$ and $y = (1-t)u + t v$ for $s,t \in [0,1)$ with $s \leq t$, where recall that $\langle x,y\rangle$ denotes the line segment in $e$ connecting $x$ and $y$.
 \item[ii)] For any Borel set $G \subset \G$, we have
 \[
 \omega^*(G) = \sum_{e\in E} \omega^*(G\cap e).
 \]
\end{enumerate}
\end{definition}

\begin{lemma}[$\omega^*$ is the length measure on graph~\citep{le2022st}] \label{lem:length-measure}
Suppose that $\G$ has no short cuts, i.e., any edge $e$ is a shortest path connecting its two end-points. Then, $\omega^*$ is a length measure in the sense that
\[
\omega^*([x,y]) = d_\G(x,y)
\]
for  any  shortest path   $[x,y]$ connecting $x, y$. Particularly, $\omega^*$ has no atom in the sense that $\omega^*(\{x\})=0$ for every $x \in \G$. 
\end{lemma}

\subsubsection{Orlicz functions}\label{appsubsec:Orlicz_functions}

We describe a brief review on Orlicz functions as summarized in~\citep{le2024generalized} for completeness. Please see~\citep{adams2003sobolev, rao1991theory}, for in-depth studies on Orlicz functions.

\paragraph{A family of convex functions.} We consider the collection  of  $N$-functions~\citep[\S8.2]{adams2003sobolev} which are special convex functions on $\R_+$. Hereafter, a strictly increasing and   convex function $\Phi: [0, \infty)\to [0, \infty)$ is called an $N$-function if  $\lim_{t \to 0} \frac{\Phi(t)}{t} = 0$ and $\lim_{t \to +\infty} \frac{\Phi(t)}{t} = +\infty$.

\paragraph{Examples of $N$-functions.} Some popular examples for $N$-functions~\citep[\S8.2]{adams2003sobolev} are 
\begin{enumerate}
\item $\Phi(t) = t^p$ with $1 < p < \infty$.
\item $\Phi(t) = \exp(t) - t- 1$.
\item $\Phi(t) = \exp(t^p) - 1$ with $1 < p < \infty$.
\item $\Phi(t) = (1+t) \log(1+t) - t$.
\end{enumerate}

\paragraph{For Luxemburg norm.} For Luxemburg norm (see Equation~\eqref{eq:Luxemburg_norm}) for Orlicz functional space, the infimum in its definition is attained~\citep[\S8.9]{adams2003sobolev}.

\paragraph{Complementary function.}
For $N$-function $\Phi$, its complementary function $\Psi : \R_+ \to \R_+$~\citep[\S8.3]{adams2003sobolev} is the $N$-function, defined as follows
\begin{equation}\label{eq:complementary_func}
\Psi(t) = \sup \left[at - \Phi(a) \, \mid \, a \ge 0  \right], \quad\mbox{for}\,\,\, t\geq 0.
\end{equation}

\paragraph{Examples of complementary pairs of $N$-functions.} Some popular complementary pairs of $N$-functions~\citep[\S8.3]{adams2003sobolev},~\citep[\S 2.2]{rao1991theory} are as follows:
\begin{enumerate}
\item $\Phi(t) = \frac{t^p}{p}$ and $\Psi(t) = \frac{t^q}{q}$ where $q$ is the conjugate of $p$, i.e., $\frac{1}{p} + \frac{1}{q} = 1$ and $1 < p < \infty$.
\item $\Phi(t) = \exp(t) - t - 1$ and $\Psi(t) = (1+t)\log(1+t) - t$.
\item For the $N$-function $\Phi(t) = \exp(t^p) - 1$ with $1 < p < \infty$, its complementary $N$-function yields an explicit for, but not simple~\citep[\S 2.2]{rao1991theory}, see~\citep[\S A.8]{le2024generalized} for the detailed derivation of the complementary $N$-function.
\end{enumerate}

\paragraph{Young inequality.} Let $\Phi, \Psi$ be a pair of complementary $N$-functions, then we have
\[
	st \le \Psi(s) + \Phi(t).
\]

\paragraph{Orlicz norm.} Besides the Luxemburg norm, the Orlicz norm~\citep[\S3.3, Definition 2]{rao1991theory} is also a popular norm for $L_{\Phi}(\G, \omega)$, defined as
\begin{equation}\label{eq:OrliczNorm}
\|f\|_{\Phi} := \sup{\Big\{ \int_{\G} | f(x) g(x)| \omega(\text{d}x) \, \mid \, \int_{\G} \Psi(|g(x)|) \omega(\text{d}x) \leq 1\Big\} }, 
\end{equation}
where $\Psi$ is the complementary $N$-function of $\Phi$.

\paragraph{Computation for Orlicz norm.} By applying~\citep[\S3.3, Theorem 13]{rao1991theory}, we can rewrite the Orlicz norm as follows: 
\begin{align*}
\|f\|_{\Phi}  =  
\inf_{k > 0} \frac{1}{k}\left( 1 + \int_{\G} \Phi(k \left| f(x) \right|) \omega(\text{d}x) \right).
\end{align*}

Therefore, one can use any second-order method, e.g., \texttt{fmincon} solver in MATLAB (with trust region reflective algorithm), for solving the \emph{univariate} optimization problem. 


\paragraph{Equivalence~\citep[\S8.17]{adams2003sobolev}~\citep[\S13.11]{musielak2006orlicz}.} The Luxemburg norm is equivalent to the Orlicz norm. In fact, we have
\begin{equation}\label{eq:LuxemburgOrlicz}
\norm{f}_{L_\Phi} \le \norm{f}_{\Phi} \le 2 \norm{f}_{L_\Phi}.
\end{equation}


\paragraph{Connection between $\boldsymbol{L^{p}}$ and $\boldsymbol{L_{\Phi}}$ functional spaces.} When the convex function $\Phi(t) = t^p$, for $1 < p < \infty$, we have 
\[
L^{p}(\G, \omega) = L_{\Phi}(\G, \omega).
\]

\paragraph{Generalized H\"older inequality.} Let $\Phi, \Psi$ be a pair of complementary $N$-functions, then generalized H\"older inequality w.r.t. Luxemburg norm~\citep[\S8.11]{adams2003sobolev} is as follows:
\begin{equation}\label{eq:Holder-Luxemburg2}
\left| \int_{\G} f(x)g(x) \omega(dx) \right| \le 2 \norm{f}_{L_\Phi} \norm{g}_{L_\Psi}.
\end{equation}
Additionally, we have the generalized H\"older inequality w.r.t. Luxemburg norm and Orlicz norm~\citep[\S13.13]{musielak2006orlicz} is as follows:
\begin{equation}\label{eq:Holder-LuxemburgOrlicz}
\left| \int_{\G} f(x)g(x) \omega(dx) \right| \le \norm{f}_{L_\Phi} \norm{g}_{\Psi}.
\end{equation}

\subsubsection{Wasserstein Distance and Orlicz Wasserstein (OW)}

We briefly review $p$-Wasserstein distance with graph metric cost, and the Orlicz Wasserstein (OW) for probability measures on a graph.

\paragraph{Wasserstein distance with graph metric cost.} Given $1\leq p <\infty$, and probability measures  $\mu$ and $\nu$ supported on graph $\G$, then the $p$-order Wasserstein distance is defined as follows:
\begin{align*}
\calW_p(\mu,\nu) 
&= \left( \inf_{\gamma \in \Pi(\mu,\nu)}\int_{\G\times\G} d_\G(x,y)^p \gamma(\dd x, \dd y) \right)^{\frac{1}{p}},
\end{align*}
where $\Pi(\mu,\nu) := \Big\{ \gamma \in \calP(\G \times \G): \, \gamma_1= \mu, \, \gamma_2= \nu \Big\}$, and $\gamma_1, \gamma_2$ are the first and second marginals of $\gamma$ respectively.

\paragraph{Orlicz Wasserstein (OW).} Following~\citet[Definition 3.2]{GuhaHN23}, the OW with the $N$-function $\Phi$ for probability measures $\mu, \nu$ supported on graph $\G$ is defined as follows:
\begin{eqnarray}\label{eq:OrliczWasserstein}
W_{\Phi}(\mu, \nu) = \inf_{\pi \in \Pi(\mu, \nu)} \inf \Big[ t > 0 : \int_{\G \times \G} \Phi\left(\frac{d_{\G}(x, z)}{t}\right) \text{d}\pi(x, z) \le 1\Big],
\end{eqnarray}
where recall that $\Pi(\mu, \nu)$ is the set of all  couplings between $\mu$ and $\nu$.

\subsubsection{Generalized Sobolev transport (GST)}\label{appsubsec:GST}

We briefly review main results on generalized Sobolev transport (GST)~\citep{le2024generalized} for probability measures on a graph.

\paragraph{Graph-based Orlicz-Sobolev space~\citep{le2024generalized}.} Let $\Phi$ be an  $N$-function and $\omega$ be a nonnegative Borel measure on graph $\G$. A continuous function $f: \G \to \R$ is said to belong to the graph-based Orlicz-Sobolev space $\OrliczSobolevPhi(\G, \omega)$ if there exists a function $h\in L_{\Phi}( \G, \omega) $ satisfying 
\begin{equation}\label{eq:OrliczSobolevFunction}
f(x) - f(z_0) =\int_{[z_0,x]} h(y) \omega(\mathrm{d}y),  \quad \forall x\in \G.
\end{equation}
Such function $h$ is unique in $L_{\Phi}(\G, \omega)$ and is called the generalized graph derivative of $f$ w.r.t.~the measure $\omega$. This generalized graph derivative of $f$ is denoted as $f'$.

\paragraph{Generalized Sobolev transport (GST)~\citep{le2024generalized}.} Let $\Phi$ be an $N$-function and $\omega$ be a nonnegative Borel measure on $\G$. For probability measures $\mu, \nu$ on a graph $\G$, the generalized Sobolev transport (GST) is defined as follows:
\begin{equation} \notag \label{eq:distance_gst}
\calGS_{\Phi}(\mu,\nu) \! \coloneqq \! \left\{
\begin{array}{cl}
\sup &  \Big| \int_\G f(x) \mu(\mathrm{d}x) - \int_\G f(x) \nu(\mathrm{d}x)  \Big| \\
\mathrm{s.t.} & f \in {\OrliczSobolevPsi}(\G, \omega),  \, \|f'\|_{L_{\Psi}}\leq 1,
\end{array}
\right.
\end{equation}
where $\Psi$ is the complementary function of $\Phi$ (see \eqref{eq:complementary_func}).

\subsubsection{Unbalanced Sobolev transport (UST)}\label{appsubsec:UST}

We give a brief review on main results of unbalanced Sobolev transport (UST)~\citep{le2023scalable} for measures on a graph, possibly having different total masses.
\paragraph{The regularized set $\mathbb U_{p'}^\alpha$ for critic function~\citep{le2023scalable}.} For $1\leq p\leq \infty$ and $0\leq \alpha\leq \frac12 [b\lambda + w_1(z_0) + w_2(z_0)]$,  let  $\mathbb U_{p'}^{\alpha}$ be  the collection of all functions $f\in W^{1,p'}(\G, \omega)$ satisfying 
\[
f(z_0) \in I_\alpha = \Big[  -w_2(z_0)- \frac{b\lambda}{2}+\alpha, w_1(z_0) + \frac{b\lambda}{2} -\alpha\Big]
\]
and 
\[
\|f'\|_{L^{p'}(\G, \omega)}\leq b.
\]
Equivalently,  $\mathbb U_{p'}^{\alpha}$ is  the collection of all functions $f$ of the form
 \begin{equation}\label{alternative_representation}
  f(x)= s +  \int_{[z_0,x]} h(y) \omega(\dd y)
 \end{equation}
with $s\in I_\alpha$ and with $h:\G \to \R$ being some function satisfying 
\[
\|h\|_{L^{p'}(\G, \omega)}\leq b.
\]

\paragraph{Unbalanced Sobolev transport (UST)~\citep{le2023scalable}.} Let $\omega$ be a nonnegative Borel measure on graph $\G$. Given $1\leq p\leq \infty$ and $0\leq \alpha\leq \frac12 [b\lambda + w_1(z_0) + w_2(z_0)]$. For unbalanced measures $\mu, \nu\in \calP(\G)$, the unbalanced Sobolev transport (UST) is defined as follows 
\[
\mathcal{US}_{p,\alpha}(\mu,\nu ) :=
 \sup_{f\in \mathbb U_{p'}^{\alpha}} \left[\int_\G f(x) \mu(\mathrm{d}x) - \int_\G f(x) \nu(\mathrm{d}x)\right]. 
\]
For simplicity, we also use $\mathcal{US}_{p}$ for the $p$-order UST when the context for $\alpha$ is clear.

\paragraph{Equal measures on a graph~\citep{le2023scalable}.}
\begin{lemma}[Lemma A.9 in~\citep{le2023scalable}]\label{lem:equal-measure}
For unbalanced measures $\mu,\nu\in \calP(\G)$, then $\mu=\nu$ if and only if  $\mu(\Lambda(x)) = \nu(\Lambda(x))$ for every $x$ in $\G$.
\end{lemma}

\subsubsection{Regularized EPT and Distance $d_{\alpha}$}\label{appsubsec:regEPT}

We briefly review main results for the regularized EPT and distance $d_{\alpha}$ in~\citep{le2021ept} for probability measures on tree $\calT$.

\paragraph{Regularized set of critic functions~\citep{le2021ept}.} Let $\mathbb L_\alpha$ be a collection of all functions $f$ of the form
 \[
  f(x)= s +  \int_{[r,x]} g(y) \omega(dy),
 \]
where $r$ is the tree root, and $s$ is a constant in the interval $\Big[  -w_2(r)- \frac{b\lambda}{2}+\alpha, w_1(r) + \frac{b\lambda}{2} -\alpha\Big]$ and with  $\|g\|_{L^\infty(\calT)}\leq b$.

\paragraph{Regularized EPT~\citep{le2021ept}.} For unbalanced measures $\mu, \nu$ supported on tree $\calT$, the regularized EPT is defined as follows:
\begin{eqnarray}\label{equ:tildeETlambda}
\widetilde{\mathrm{ET}}_\lambda^\alpha(\mu,\nu) := \sup \left\{ \int_\calT f (\mu - \nu):\, f\in \mathbb{L}_\alpha  \right\}  - \frac{b\lambda}{2}\big[ \mu(\calT) +  \nu(\calT)\big].
\end{eqnarray}

Following~\citep[Proposition 3.8]{le2021ept}, we have
\begin{eqnarray*}
\widetilde{\mathrm{ET}}_\lambda^\alpha(\mu,\nu) = \int_{\calT} | \mu(\Lambda(x)) -  \nu(\Lambda(x))| \, \omega(dx) - \frac{b\lambda}{2}\big[ \mu(\calT) +  \nu(\calT)\big]  +   \big[w_i(r) +\frac{b\lambda}{2} -\alpha\big] |\mu(\calT)-\nu(\calT)| 
\end{eqnarray*}
with
$i :=1$ if $ \mu(\calT)\geq \nu(\calT)$ and $i:=2$ if $ \mu(\calT)< \nu(\calT)$. 

\paragraph{Distance $d_{\alpha}$~\citep{le2021ept}.} We briefly review the definition of distance $d_{\alpha}$ in~\citep{le2021ept}
\begin{equation}\label{equ:dAlpha}
d_\alpha(\mu,\nu) := \widetilde{\mathrm{ET}}_\lambda^\alpha(\mu,\nu)  +\frac{b\lambda}{2}\big[ \mu(\calT) +  \nu(\calT)\big].
\end{equation}

Following~\citep[Proposition 3.10]{le2021ept}, the distance $d_{\alpha}$ is a metric.

\subsection{Further Discussions}\label{appsubsec:further_discussion}

We give further discussions and details for various aspects in our work. For completeness, we recall important discussions on the graph in~\citep{le2022st} (for Sobolev transport for probability measures), since these discussions and results are also applied and/or easily adapted for our proposed approaches.

\paragraph{Further details for the computation of Orlicz-EPT and OST.} We describe further details for the computation for Orlicz-EPT and OST.

\textbf{$\bullet$ For Orlicz-EPT.} Following the theoretical ground derived for the computation of the entropic regularized Orlicz-EPT, i.e., the objective function is monotone non-increasing (Proposition~\ref{prop:monotonicity_regOT}), and the lower and upper limits for the objective functions (Proposition~\ref{prop:limits_regOT}), one can compute the entropic regularized Orlicz-EPT by a binary search algorithmic approach. For completeness, we straightforwardly describe the pseudo-code for it in Algorithm~\ref{alg:bs_OrliczEPT}. Additionally, notice that we can leverage Proposition~\ref{rm:upperlimit_regOT} in \S\ref{app:subsec:upper_bound_over_entropicOT} to alternatively set the initial value for $t_{\ell}$ in Algorithm~\ref{alg:bs_OrliczEPT} (line 3).

\newcommand\mycommfont[1]{\footnotesize\ttfamily\textcolor{blue}{#1}}
\SetCommentSty{mycommfont}

\begin{algorithm*}\label{alg:bs_OrliczEPT}

\DontPrintSemicolon
\caption{Compute entropic regularized Orlicz-EPT $\calOE_{\Phi,\varepsilon}$}  \label{alg:bs_OrliczEPT} 


\textbf{Input:} Input measures $\mu, \nu$, function $\Phi$, graph $\G$, parameters $b, \lambda, \varepsilon$, and stopping threshold $\bar{\varepsilon}$
 
\textbf{Output:} entropic regularized Orlicz-EPT $\calOE_{\Phi,\varepsilon}(\mu, \nu)$

\nl Construct $\hat\G = \G \cup \{\hat s\}$ and corresponding nonnegative cost function $\hat c$ (\S\ref{sec:OrliczEPT})

\nl Construct corresponding probability measures $\hat\mu = \frac{\mu +\nu(\G) \delta_{\hat s}}{\mu(\G) + \nu(\G)}$ and $\hat\nu = \frac{\nu +\mu(\G) \delta_{\hat s}}{\mu(\G) + \nu(\G)}$.


\nl Set $t_{r} = \frac{L_{\hat \mu, \hat \nu}}{\Phi^{-1}(1 + \varepsilon)}$ and $t_{\ell} = \frac{\calW_{\hat c}(\hat \mu, \hat \nu)}{\Phi^{-1}(1 + \varepsilon\left[ H(\hat \mu) + H(\hat \nu) - 1 \right])}$ 



\nl \While{$t_r - t_{\ell} > \bar{\varepsilon}$} 	
{
    \nl Set $t_m = \frac{t_{\ell} + t_r}{2}$
    
    \nl Compute $f_m = \mathcal{A}_{\varepsilon} \! \left(t_m; \hat \mu, \hat \nu\right)$

    \nl \If{$f_m \le 1$} 	
    {
        \nl Set $t_r = t_m$ 
        
        \nl \If{$f_m == 1$}
        {
            \nl Break
        }
    }
    \Else
    {
        \nl Set $t_{\ell} = t_m$ 
    }
}

\nl Return $\calOE_{\Phi,\varepsilon}(\mu, \nu) = \left(\mu(\G) + \nu(\G)\right)(t_r - b\lambda)$

\end{algorithm*}

\textbf{$\bullet$ For OST.} For popular $N$-function, it is easy to derive its gradient and Hessian for the objective function of the univariate optimization problem. Therefore, in our experiments, we leverage the \texttt{fmincon} MATLAB solver with the \emph{trust-region-reflective} algorithm to solve the univariate optimization problem for OST computation.

\paragraph{Further discussion on Orlicz-EPT and OST.} Following Proposition~\ref{prop:relation_OST_GST}, we can view OST as an extension of GST~\citep{le2024generalized} for handling unbalanced measures, and Orlicz-EPT as an extension of OW~\citep{sturm2011generalized} for unbalanced measures. Notably, when the input measures have equal mass (i.e., $\mu(\G) = \nu(\G)$), and $b=1$ 
, Orlicz-EPT reduces to OW. Furthermore, Orlicz-EPT shares the same computational complexity as OW, as shown in Equation~\eqref{eq:OrliczEPT}. It is worth noting that GST is a scalable variant of OW for balanced measures, in the same sense, OST can be seen as a scalable variant of Orlicz-EPT.

As a result, Orlicz-EPT is applicable for all applications of OW and extends OW to handle unbalanced input measures, whereas OST may not reserve all the properties of OW.

Like OW, Orlicz-EPT involves a two-level optimization problem, limiting its applicability to small-scale domains. In contrast, OST, similar to GST, is scalable and can be applied to large-scale domains.

\begin{figure}
  \begin{center}
    \includegraphics[width=0.4\textwidth]{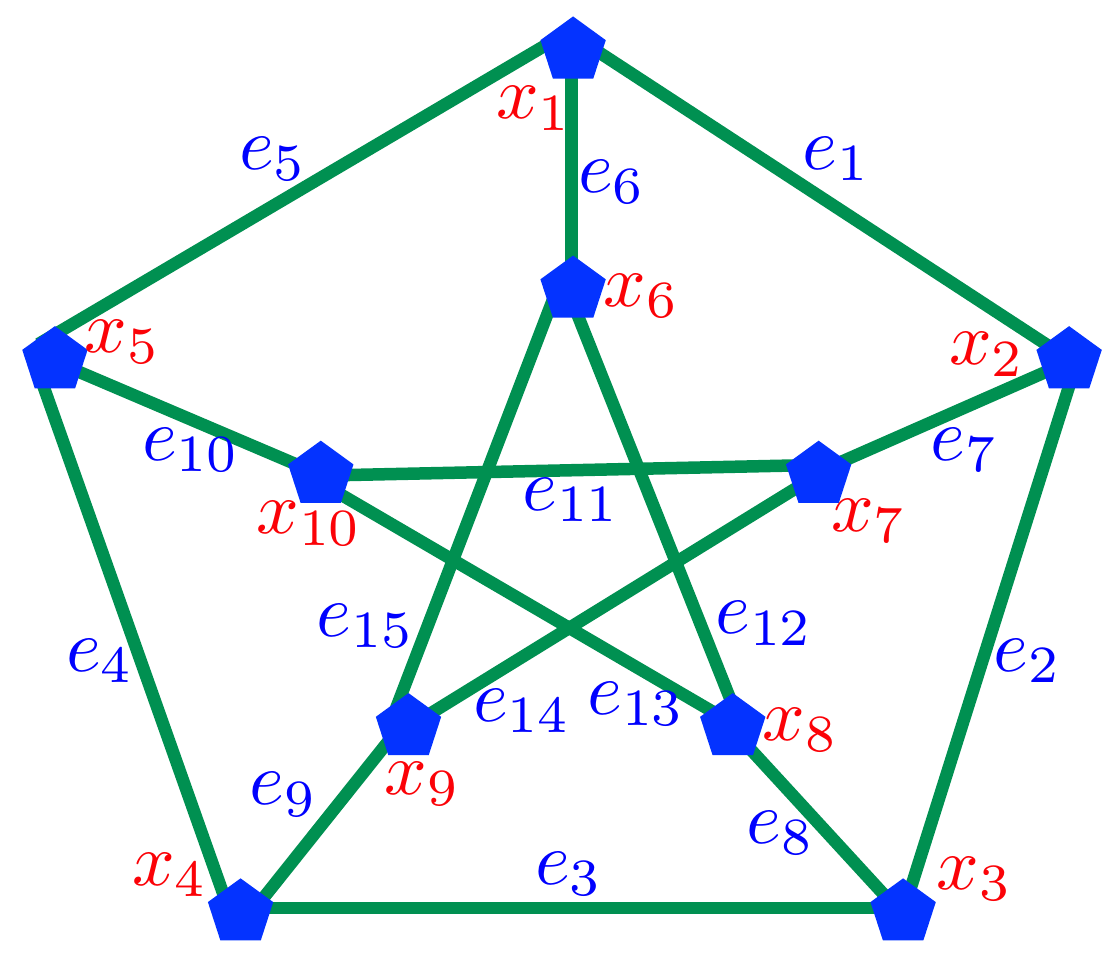}
  \end{center}
  \vspace{-6pt}
  \caption{A geodetic graph illustration. The set of nodes $V$ has $10$ nodes, i.e., $V = \left\{x_1, x_2, \dotsc, x_{10}\right\}$. The set of edges $E$ has $15$ edges, i.e., $E = \left\{e_1, e_2, \dotsc, e_{15}\right\}$ where each edge weight/length is set to one, i.e., $w_{e_j}=1$, for $1 \le j \le 15$. For any $x_i, x_j$, there is a unique shortest path between them, with a length $2$. Let $x_1$ be the unique-path root node (i.e., $z_0 = x_1$) and $\widetilde{\G}$ be a subgraph containing 3 nodes $\left\{x_6, x_8, x_9\right\}$ and 2 edges $\left\{e_{12}, e_{15}\right\}$, then we have $\Lambda(x_6) = \gamma(e_6) = \widetilde{\G}$.}
  \label{fg:GeodeticGraph}
\end{figure}

\paragraph{Illustrations for notations on a graph.} We illustrate notions on a graph in our work in Figure~\ref{fg:GeodeticGraph}.

\paragraph{Geodetic graph.} Recall that for a graph, for every pairs of nodes, the shortest path between them is unique, then it is called geodetic graph~\citep{le2022st}. Therefore, geodetic graphs are special examples satisfying the uniqueness property of the shortest paths (\S\ref{sec:preliminaries}). We give an illustration of geodetic graph in Figure~\ref{fg:GeodeticGraph}. 


\paragraph{Path length.} As discussed in~\citep{le2022st}, we can compute a path length connecting any two points  $x, y \in \G$, where the points $x, y$ are not necessary to be nodes in $V$ (but even for any point on an edge as well). Indeed, for the points $x, y \in \R^n$ belonging to the same edge $e= \langle u, v\rangle$, connecting two nodes $u, v \in V$. We have 
\begin{eqnarray*}
& x = (1-s) u + s v, \\
& y = (1-t)u + t v,
\end{eqnarray*}
for some scalars $t,s\in [0,1]$. Thus, the length of the path connecting $x, y$ along the edge $e$ (i.e., the line segment $\langle x, y\rangle$) can be computed by $|t-s| w_e$. 

Consequently, the length for an arbitrary path in $\G$ can be computed by breaking down into pieces over edges, and then summing over their corresponding lengths~\citep{le2022st}.

\paragraph{Extension to measures supported on $\G$.} For the discrete case, similar to ST~\citep{le2022st}, OST~\eqref{eq:OST_1d_opt_discrete} is extendable for measures with finite supports on $\G$, i.e., measures which may have supports on edges, by following the strategy to compute a path length for points in $\G$. Precisely, we break down edges containing supports into pieces, and then sum over their corresponding values, instead of the sum over edges as in~\eqref{eq:OST_1d_opt_discrete} for OST.

\textbf{About the uniqueness property of the shortest paths on $\G$.} As discussed in \citep{le2022st}, note that edge length is a real nonnegative scale, $w_e \in \R_{+}$ for any edge $e \in E$ in $\G$., it is almost surely that every node in $V$ can be regarded as unique-path root node since with a high probability, lengths of paths connecting any two nodes in graph $\G$ are different.

Moreover, for some special graph, e.g., a grid of nodes, there is \emph{no} unique-path root node for such graphs. However, by perturbing each node (or also perturbing edge lengths in case $\G$ is a non-physical graph) with a small deviation, the perturbed graph will satisfy the unique-path root node assumption.

For continuous case, when measures are extended to support on $\G$, and input measures are fully supported, then for some finite special nodes where there are multiple shortest paths, we fix one of the shortest paths to those points, and treat it as the chosen shortest path for those special points. However, for practical applications, the number of supports are finite, which bypasses the mentioned issue.

\paragraph{For the given graph setting.} As in~\citep{le2022st}, we assume that the graph metric space is given. The question to learn an optimal graph metric structure from data is not considered in this work and leave for future investigation.

\paragraph{Measures on a graph.} In this work, we consider OT problem for \emph{two input unbalanced measures} supported on the \emph{same} graph. See~\citep{le2023scalable} for the same problem setting. 

The proposed approaches, Orlicz-EPT and OST, are for \emph{input unbalanced measures}, i.e., to compute the distance between two unbalanced measures, on the \emph{same} graph. We distinguish our considered problem to the following related problems: 


\textbf{$\bullet$ Compute distances/discrepancies between two (different) input graphs.} For examples,~\citep{petric2019got, dong2020copt} compute distance/discrepancy between \emph{two (different) input graphs}. They are essentially different to our considered problem which computes distance between \emph{two input probability measures} supported on the \emph{same} graph. In particular, Le et al.~\citep{le2021fba} consider a variant of Gromov-Wasserstein problem for \emph{two input probability measures}, but possibly supported on \emph{different} tree metric spaces.

\textbf{$\bullet$ Compute kernels between two (different) input graphs.} Graph kernels are kernel functions between two input (different) graphs to assess their similarity. See~\citep{borgwardt2020graph, kriege2020survey, nikolentzos2021graph} for comprehensive reviews on graph kernels.  Essentially, it is different to our proposed approaches to compute distances between \emph{two input unbalanced measures} on the \emph{same} graph.

\textbf{For (variational) OT problems on a graph.} OT problems for measures supported on a graph metric space have been explored in previous studies~\citep{le2022st, le2023scalable, le2024generalized}. Additionally, Le et al.~\citep{le2025scalable} have recently studied a variational OT problem, e.g., Sobolev IPM where the critic function is constrained within a unit ball of Sobolev norm involving both the critic function and its gradient, for graph-based measures. Notably, graph metrics extend tree metrics, which are utilized in scalable optimal transport methods like tree-sliced Wasserstein (TSW)~\citep{LYFC, tran2025distancebased, tran2025nonlinear, tran2025geometric, tran2025spherical}. TSW, in turn, generalizes the popular sliced-Wasserstein (SW) approach~\citep{rabin2011wasserstein, bonneel2015sliced, nguyen2024randompathSW}. The graph structure offers greater flexibility and degrees of freedom compared to the tree structure in TSW and the line structure in SW. For a more in-depth discussion on the motivation for OT on a graph, please refer to~\citep{le2022st}.

\paragraph{Persistence diagrams for TDA.} Persistence diagrams (PD) are multisets of data points in $\R^2$, containing the birth and death time respectively of topological features (e.g., connected component, ring, or cavity), extracted by algebraic topology methods (e.g., persistence homology)~\citep{edelsbrunner2008persistent}.

\paragraph{Graphs $\G_{\text{Log}}$ and $\G_{\text{Sqrt}}$~\citep{le2022st}.} For completeness, we review the construction for graphs $\G_{\text{Log}}$ and $\G_{\text{Sqrt}}$ in ~\citep{le2022st}. We utilize the farthest-point clustering method to cluster supports of measures into at most $M$ clusters. Then, let the vertex set $V$ be the collection of centroids of these clusters, i.e., graph vertices. For edges, we randomly select $(M\log{M})$ edges, and $M^{3/2}$ edges for graphs  $\G_{\text{Log}}$, and $\G_{\text{Sqrt}}$ respectively. Let $\tilde{E}$ be the set of those randomly sampled edges. For each edge $e$, its edge length/weight $w_e$ is computed by Euclidean distance between the two corresponding end points (i.e., corresponding nodes of edge $e$). Let $n_c$ be the number of connected components in $\tilde{\G}(V, \tilde{E})$. We randomly add $(n_c - 1)$ more edges between these $n_c$ connected components to form a connected graph $\G$ from $\tilde{\G}$. Let $E_c$ be the set of these $(n_c - 1)$ added edges, and denote set $E = \tilde{E} \cup E_c$, then $\G(V, E)$ is the constructed graph.

\paragraph{Further discussion on graphs in experiments.} For all datasets, except \texttt{MPEG7} dataset, $\G_{\text{Sqrt}}$ consists of 10K nodes and 1 million edges, while $\G_{\text{Log}}$ comprises 10K nodes and 100K edges. Due to the smaller size of the \texttt{MPEG7} dataset, we constructed $\G_{\text{Sqrt}}$ with 1K nodes and 32K edges, and $\G_{\text{Log}}$  with 1K nodes and 7K edges.

\paragraph{Datasets and Computational Devices.} For the datasets in our experiments, one can contact the authors of Sobolev transport~\citep{le2022st} to access to them. Additionally, all of our experiments are run on commodity hardware.

In our experiments, we demonstrate the effectiveness of our approach in document classification on four real-world datasets and topological data analysis (TDA), including orbit recognition for linked twist maps, i.e., a discrete dynamical system modeling flows in DNA microarrays~\citep{hertzsch2007dna}, and object shape recognition in \texttt{MPEG7}. These evaluations on document classification and TDA are often used for tasks involving comparing measures on a graph, see~\citep{le2022st, le2024generalized}. We believe that such experimental coverage is rich and diverse enough.

\paragraph{Hyperparameter validation.} We use the same validation as in~\citep{le2023scalable}. Precisely, we further randomly split \emph{the training set} into $70\%/30\%$ for validation-training and validation with $10$ repeats to choose hyper-parameters for the experiments.

\paragraph{Further discussion on hyperparameters.} The performance of OST/Orlicz-EPT typically depends on the choice of the $N$-function $\Phi$, similar to how kernel functions impact performance in kernel-dependent machine learning frameworks. In our experiments with N-functions $\Phi_1$ and $\Phi_2$ (\S\ref{sec:experiments}), we observed slightly different performances. Similar findings have been reported for GST~\citep{le2024generalized}.

Determining the optimal $N$-function $\Phi$ for OST/Orlicz-EPT in a given task is an open problem that warrants further investigation. We leave it for future work. As an interim solution, cross-validation can be used to select $\Phi$ from a set of candidate functions.

Regarding the regularization weight $b$, we used $b=1$ in our experiments based on the results in Propositions~\ref{prop:relation_OST_GST} and~\ref{prop:relation_OST_ST}, as well as for simplicity. This choice of $b$ is supported by theoretical results on EPT on a graph~\citep[Lemma A.6, Remark 4.8]{le2023scalable} and has been used in previous experiments for EPT~\citep{le2021ept, le2023scalable}.

For $\alpha$, from the result in Proposition~\ref{prop:limit_OST_OrliczEPT} and for simplicity, we use $\alpha=0$ for experiments. Such value for $\alpha$ is also supported by theoretical results of EPT on a graph~\citep[Lemma 4.4, Proposition 5.2]{le2023scalable} and used in experiments for EPT~\citep{le2021ept, le2023scalable}. Additionally, recall that $\mathcal{I}_{\alpha}$ is the largest interval when $\alpha=0$.

\paragraph{The number of pairs for kernel SVM~\citep{le2024generalized}.} Denote $N_{tr}, N_{te}$ for the number of measures used for training and test respectively. For the kernel SVM training, the number of pairs for computing the distances is $(N_{tr}-1) \times \frac{N_{tr}}{2}$. For the test, the number of pairs for computing the distances is $N_{tr} \times N_{te}$. Thus, for each run, the number of pairs for computing the distances for both training and test is totally $N_{tr} \times (\frac{N_{tr}-1}{2} +  N_{te})$. In Table~\ref{tb:numpairs}, we summarize the number of pairs which we need to evaluate distances/discrepancies for SVM in each run to illustrate the experimental scale, e.g., more than $29$M pairs for \texttt{AMAZON}.

\begin{table}[]
\caption{The number of pairs for SVM.}
\label{tb:numpairs}
    \centering
    \renewcommand*{\arraystretch}{1}
  \resizebox{0.26\textwidth}{!}{%
\begin{tabular}{|l|c|}
\hline
Datasets & \#pairs \\ \hline
\texttt{TWITTER}  & 4394432 \\ \hline
\texttt{RECIPE}   & 8687560      \\ \hline
\texttt{CLASSIC}  & 22890777       \\ \hline
\texttt{AMAZON}   & 29117200      \\ \hline
\texttt{Orbit}    & 11373250   \\ \hline
\texttt{MPEG7}    & 18130     \\ \hline
\end{tabular}
} 
\vspace{-6pt}
\end{table}


\paragraph{Debiased approaches for (Sinkhorn-based) UOT~\citep{sejourne2019sinkhorn, sejourne2023unbalanced}.} We review the discussion and evaluation in~\citep{le2023scalable} for the debiased approaches for (Sinkhorn-based) UOT in~\citep{sejourne2019sinkhorn, sejourne2023unbalanced}, with the same setup as in our experiments. As noted in the main text, the debiased version for Sinkhorn-based approach for UOT~\citep{frogner2015learning, sejourne2019sinkhorn} which may be helpful for applications, and the debiased version is empirically indefinite. Both the UOT and its debiased version have the same computational complexity.

Empirical results for the debiased version~\citep{sejourne2019sinkhorn, sejourne2023unbalanced} are given in~\citep[Figures 41--44]{le2023scalable}. As in~\citep{le2023scalable}, the debiased version improve performances of UOT in some datasets, especially for datasets in TDA tasks (\texttt{Orbit} and $\texttt{MPEG7}$). For document datasets, performances of the debiased version and UOT are comparative, i.e., the role of debias property is not clear for advantages in applications.

\paragraph{Broader impacts.} In this work, we propose novel approaches to extend OW/GST for unbalanced measures on a graph. The proposed Orlicz-EPT is directly derived from standard OT, bypassing challenges raised from unbalanced measures, as in OW. Additionally, we formulate a novel regularization approach, resulting in the proposed OST, which is efficient in computation. Therefore, our proposals pave the ways to use OT approach endowed with Orlicz geometric structure for applications with unbalanced measures, which is common in real-world scenarios. To our knowledge, there is no foresee negative social impacts for our research.

\subsection{Further Empirical Results}\label{appsubsec:further_empirical_results}


\paragraph{Further empirical results on graph ~$\G_{\text{Log}}$.} We provide corresponding results as in \S\ref{sec:experiments} for graph~$\G_{\text{Log}}$.

\begin{figure}[h]
  \begin{center}
    \includegraphics[width=0.6\textwidth]{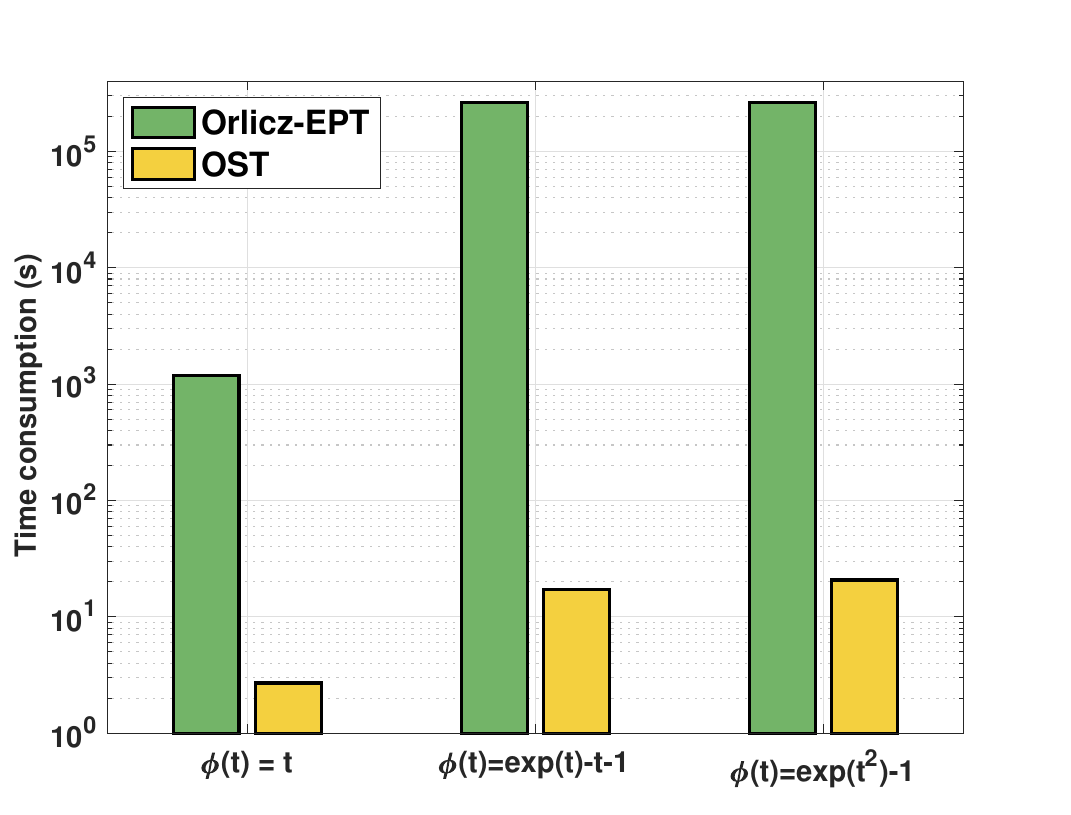}
  \end{center}
  \vspace{-6pt}
  \caption{Time consumption on graph $\G_{\text{Log}}$.}
  \label{fg:Time_OST_OrliczEPT_10K_LLE}
\end{figure}

\begin{figure*}[ht]
  \begin{center}
    \includegraphics[width=\textwidth]{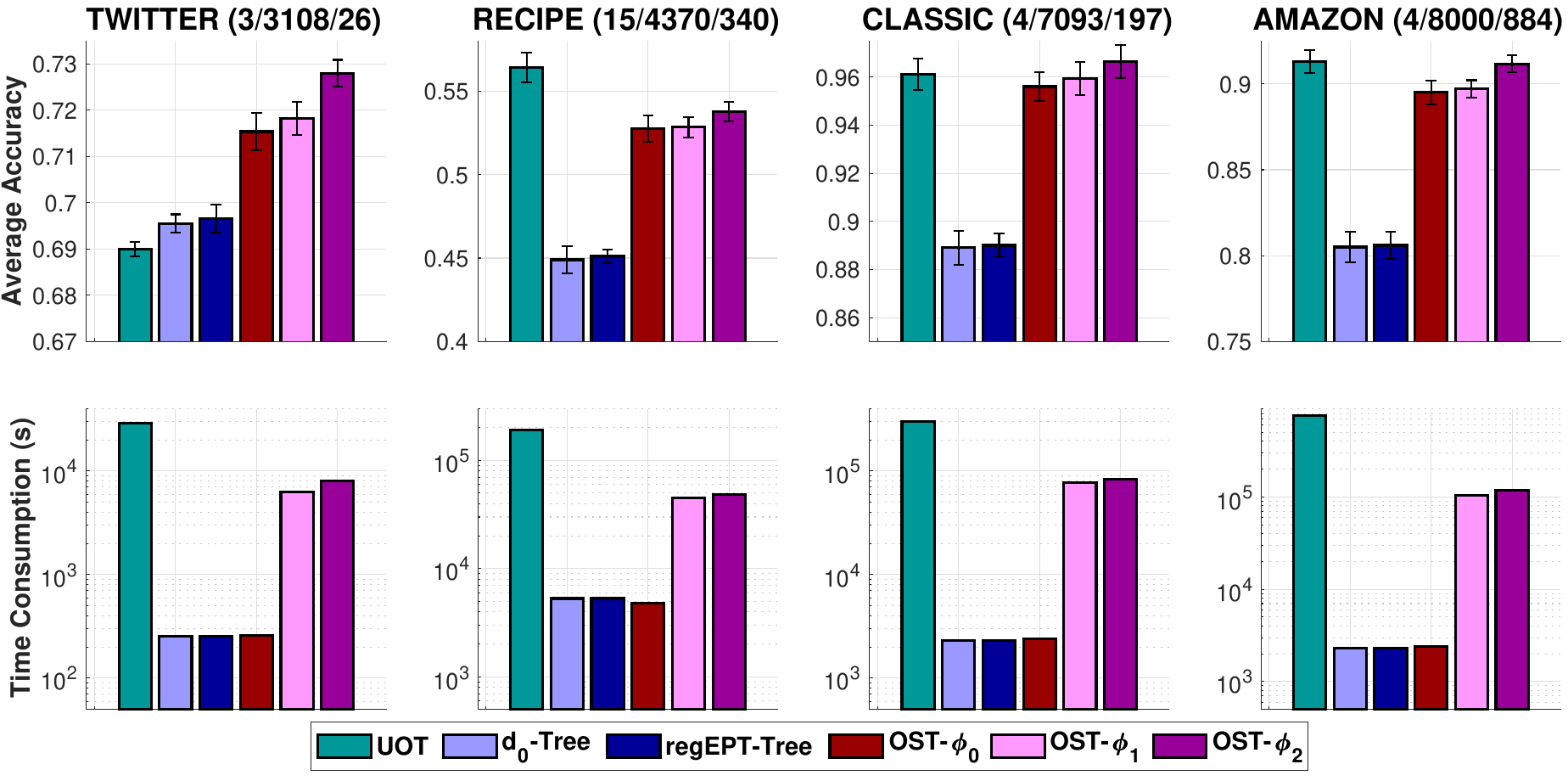}
  \end{center}
  \vspace{-6pt}
  \caption{Document classification on graph $\G_{\text{Log}}$.}
  \label{fg:DOC_LLE_10K}
 \vspace{-6pt}
\end{figure*}

\begin{figure}[h]
  \begin{center}
    \includegraphics[width=0.65\textwidth]{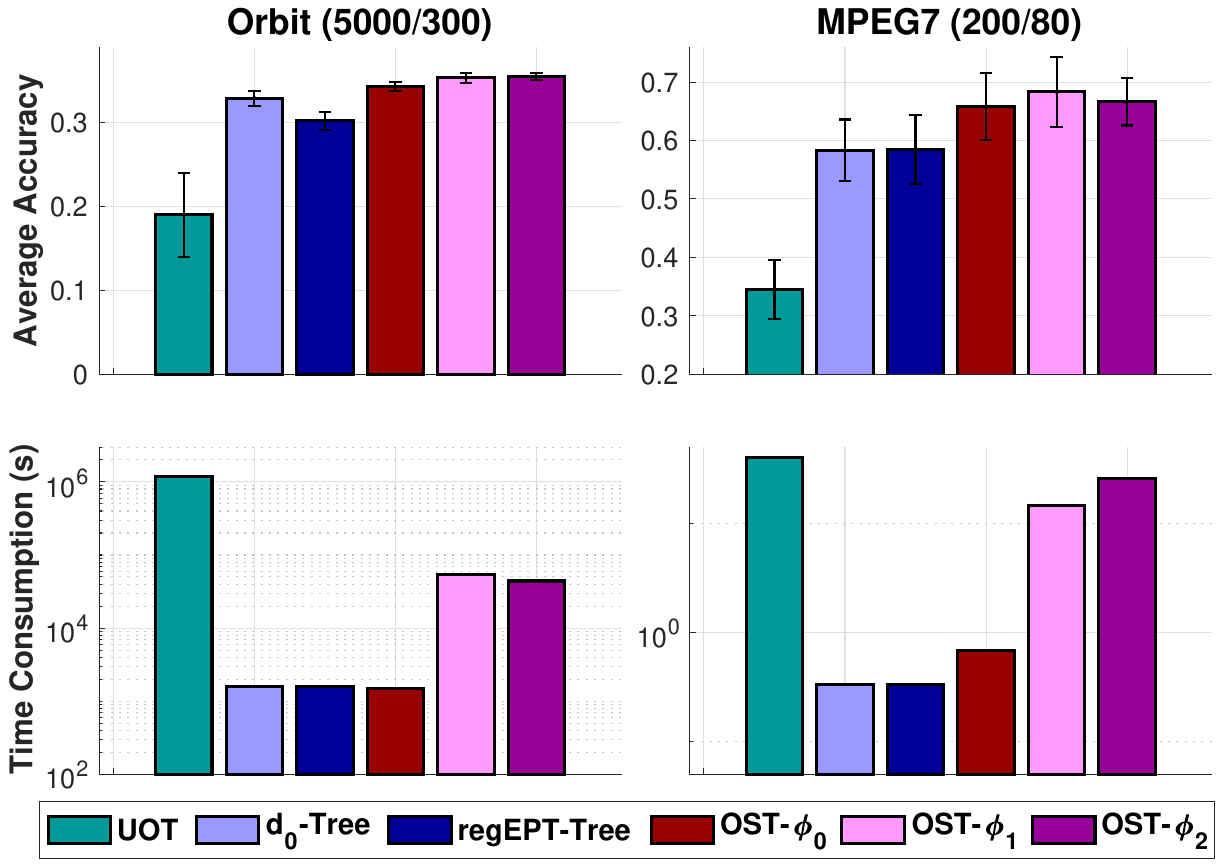}
  \end{center}
  \vspace{-6pt}
  \caption{TDA on graph $\G_{\text{Log}}$.}
  \label{fg:TDA_LLE_10K1K}
 \vspace{-6pt}
\end{figure}

\begin{itemize}
    \item In Figure~\ref{fg:Time_OST_OrliczEPT_10K_LLE}, we compare the time consumption of OST and Orlicz-EPT with $\Phi_0, \Phi_1, \Phi_2$ on graph~$\G_{\text{Log}}$.

    \item In Figure~\ref{fg:DOC_LLE_10K}, we illustrate the SVM results and time consumptions of kernels on document classification with graph~$\G_{\text{Log}}$.

    \item In Figure~\ref{fg:TDA_LLE_10K1K}, we illustrate the SVM results and time consumptions of kernels on TDA with graph~$\G_{\text{Log}}$.
\end{itemize}

\paragraph{Further empirical results for different hyperparameters $b$ and $\alpha$.} We carry out more additional experiments for different $b, \alpha$. The results for average accuracy, and the standard deviation (in the parentheses) are as follows:

\textbf{$\bullet$ For document classification.}

\begin{itemize}
    \item We present further SVM results for document classification on \texttt{TWITTER} dataset with graph $\G_{\text{Sqrt}}$ and different hyperparameters $b$ and $\alpha$ in Tables~\ref{tb:doc_sqrt_b} and~\ref{tb:doc_sqrt_alpha} respectively.

    \item We present further SVM results for document classification on \texttt{TWITTER} dataset with graph $\G_{\text{Log}}$ and different hyperparameters $b$ and $\alpha$ in Tables~\ref{tb:doc_log_b} and~\ref{tb:doc_log_alpha} respectively.
\end{itemize}


\begin{table}[]
\caption{SVM results for document classification on \texttt{TWITTER} dataset with graph $\G_{\text{Sqrt}}$ and different hyperparameter $b$.}
\centering
\label{tb:doc_sqrt_b}
\begin{tabular}{l|l|l|l|}
\cline{2-4}
                              & \multicolumn{1}{c|}{$\Phi_0$} & \multicolumn{1}{c|}{$\Phi_1$} & \multicolumn{1}{c|}{$\Phi_2$} \\ \hline
\multicolumn{1}{|l|}{$b=0.5$} & $71.66  \pm 0.63$             & $72.03 \pm 0.56$              & $72.12 \pm  0.57$             \\ \hline
\multicolumn{1}{|l|}{$b=2$}   & $71.54 \pm  0.49$             & $72.15 \pm 0.42$              & $72.28 \pm 0.34$              \\ \hline
\end{tabular}
\end{table}

\begin{table}[]
\caption{SVM results for document classification on \texttt{TWITTER} dataset with graph $\G_{\text{Sqrt}}$ and different hyperparameter $\alpha$.}
\centering
\label{tb:doc_sqrt_alpha}
\begin{tabular}{l|l|l|l|}
\cline{2-4}
                                   & \multicolumn{1}{c|}{$\Phi_0$} & \multicolumn{1}{c|}{$\Phi_1$} & \multicolumn{1}{c|}{$\Phi_2$} \\ \hline
\multicolumn{1}{|l|}{$\alpha=0.1$} & $71.48 \pm 0.43$              & $72.34 \pm 0.55$              & $72.64 \pm 0.50$              \\ \hline
\multicolumn{1}{|l|}{$\alpha=0.2$} & $71.64 \pm 0.50$              & $72.12 \pm 0.36$              & $72.27 \pm 0.30$              \\ \hline
\end{tabular}
\end{table}


\begin{table}[]
\caption{SVM results for document classification on \texttt{TWITTER} dataset with graph $\G_{\text{Log}}$ and different hyperparameter $b$.}
\centering
\label{tb:doc_log_b}
\begin{tabular}{l|l|l|l|}
\cline{2-4}
                              & \multicolumn{1}{c|}{$\Phi_0$} & \multicolumn{1}{c|}{$\Phi_1$} & \multicolumn{1}{c|}{$\Phi_2$} \\ \hline
\multicolumn{1}{|l|}{$b=0.5$} & $70.97 \pm 0.29$              & $71.74 \pm 0.33$              & $71.78 \pm 0.41$              \\ \hline
\multicolumn{1}{|l|}{$b=2$}   & $71.58 \pm 0.41$              & $71.92 \pm 0.16$              & $71.96 \pm 0.22$              \\ \hline
\end{tabular}
\end{table}

\begin{table}[]
\caption{SVM results for document classification on \texttt{TWITTER} dataset with graph $\G_{\text{Log}}$ and different hyperparameter $\alpha$.}
\centering
\label{tb:doc_log_alpha}
\begin{tabular}{l|l|l|l|}
\cline{2-4}
                                   & \multicolumn{1}{c|}{$\Phi_0$} & \multicolumn{1}{c|}{$\Phi_1$} & \multicolumn{1}{c|}{$\Phi_2$} \\ \hline
\multicolumn{1}{|l|}{$\alpha=0.1$} & $71.57 \pm 0.23$              & $71.87 \pm 0.36$              & $72.04 \pm 0.28$              \\ \hline
\multicolumn{1}{|l|}{$\alpha=0.2$} & $71.59 \pm 0.35$              & $71.74 \pm 0.22$              & $71.95 \pm 0.19$              \\ \hline
\end{tabular}
\end{table}

\textbf{$\bullet$ For TDA.}

\begin{itemize}
    \item We present further SVM results for TDA on \texttt{MPEG7} dataset with graph $\G_{\text{Sqrt}}$ and different hyperparameters $b$ and $\alpha$ in Tables~\ref{tb:tda_sqrt_b} and~\ref{tb:tda_sqrt_alpha} respectively.

    \item We present further SVM results for TDA on \texttt{MPEG7} dataset with graph $\G_{\text{Log}}$ and different hyperparameters $b$ and $\alpha$ in Tables~\ref{tb:tda_log_b} and~\ref{tb:tda_log_alpha} respectively.
    
\end{itemize}


\begin{table}[]
\caption{SVM results for TDA on \texttt{MPEG7} dataset with graph $\G_{\text{Sqrt}}$ and different hyperparameter $b$.}
\centering
\label{tb:tda_sqrt_b}
\begin{tabular}{l|l|l|l|}
\cline{2-4}
                              & \multicolumn{1}{c|}{$\Phi_0$} & \multicolumn{1}{c|}{$\Phi_1$} & \multicolumn{1}{c|}{$\Phi_2$} \\ \hline
\multicolumn{1}{|l|}{$b=0.5$} & $66.67 \pm 3.45$              & $71.67 \pm 5.40$              & $68.83 \pm 3.09$              \\ \hline
\multicolumn{1}{|l|}{$b=2$}   & $65.00 \pm 3.58$              & $69.13 \pm 3.87$              & $67.33 \pm 3.55$              \\ \hline
\end{tabular}
\end{table}

\begin{table}[]
\caption{SVM results for TDA on \texttt{MPEG7} dataset with graph $\G_{\text{Sqrt}}$ and different hyperparameter $\alpha$.}
\centering
\label{tb:tda_sqrt_alpha}
\begin{tabular}{l|l|l|l|}
\cline{2-4}
                                   & \multicolumn{1}{c|}{$\Phi_0$} & \multicolumn{1}{c|}{$\Phi_1$} & \multicolumn{1}{c|}{$\Phi_2$} \\ \hline
\multicolumn{1}{|l|}{$\alpha=0.1$} & $66.43 \pm 4.81$              & $70.47 \pm 4.51$              & $68.50 \pm 4.54$              \\ \hline
\multicolumn{1}{|l|}{$\alpha=0.2$} & $66.82 \pm 4.01$              & $68.43 \pm 4.63$              & $69.17 \pm 3.60$              \\ \hline
\end{tabular}
\end{table}


\begin{table}[]
\caption{SVM results for TDA on \texttt{MPEG7} dataset with graph $\G_{\text{Log}}$ and different hyperparameter $b$.}
\centering
\label{tb:tda_log_b}
\begin{tabular}{l|l|l|l|}
\cline{2-4}
                              & \multicolumn{1}{c|}{$\Phi_0$} & \multicolumn{1}{c|}{$\Phi_1$} & \multicolumn{1}{c|}{$\Phi_2$} \\ \hline
\multicolumn{1}{|l|}{$b=0.5$} & $65.87 \pm 5.68$              & $67.00 \pm 5.60$              & $68.67 \pm 5.61$              \\ \hline
\multicolumn{1}{|l|}{$b=2$}   & $65.93 \pm 5.11$              & $68.91 \pm 6.00$              & $66.17 \pm 4.60$              \\ \hline
\end{tabular}
\end{table}

\begin{table}[]
\caption{SVM results for TDA on \texttt{MPEG7} dataset with graph $\G_{\text{Log}}$ and different hyperparameter $\alpha$.}
\centering
\label{tb:tda_log_alpha}
\begin{tabular}{l|l|l|l|}
\cline{2-4}
                                   & \multicolumn{1}{c|}{$\Phi_0$} & \multicolumn{1}{c|}{$\Phi_1$} & \multicolumn{1}{c|}{$\Phi_2$} \\ \hline
\multicolumn{1}{|l|}{$\alpha=0.1$} & $65.94 \pm 4.19$              & $67.83 \pm 5.83$              & $66.46 \pm 4.58$              \\ \hline
\multicolumn{1}{|l|}{$\alpha=0.2$} & $65.84 \pm 5.45$              & $69.61 \pm 5.56$              & $69.86 \pm 5.22$              \\ \hline
\end{tabular}
\end{table}
We observe that turning these hyperparameters $b$ and $\alpha$, e.g., via cross-validation, may help to improve the performances further.


\clearpage
\section*{NeurIPS Paper Checklist}

\begin{enumerate}

\item {\bf Claims}
    \item[] Question: Do the main claims made in the abstract and introduction accurately reflect the paper's contributions and scope?
    \item[] Answer: \answerYes{} 
    \item[] Justification: The proposed Orlicz-EPT is presented in \S\ref{sec:OrliczEPT}, where we revisit the EPT problem, leverage~\citet{CM}'s insights to reformulate EPT as a corresponding standard OT. By carefully calibrating the ground cost of the corresponding standard OT, we guarantee that its ground cost is nonnegative, which is essential to develop Orlicz-EPT, a variant of EPT with Orlicz geometric structure, since the $N$-function is only defined on the nonnegative domain. Additionally, we provide theoretical background to solve Orlicz-EPT by a binary search approach.

    Additionally, we present the proposed Orlicz-Sobolev transport (OST) in \S\ref{sec:OST}. By leveraging the dual EPT and the underlying graph structure, we derive a novel regularization for the critic function, to develop the proposed OST. We prove that OST can be computed by simply solving a univariate optimization problem.

    Furthermore, in \S\ref{sec:properties_OrliczSobolevTransport}, we derive geometric structures for OST, and show its connections to other transport distances. In \S\ref{sec:experiments}, we empirically illustrate that OST is several-order faster than Orlicz-EPT. We also show evidences on the advantages of OST for unbalanced measures on a graph for document classification and TDA.
    \item[] Guidelines:
    \begin{itemize}
        \item The answer NA means that the abstract and introduction do not include the claims made in the paper.
        \item The abstract and/or introduction should clearly state the claims made, including the contributions made in the paper and important assumptions and limitations. A No or NA answer to this question will not be perceived well by the reviewers. 
        \item The claims made should match theoretical and experimental results, and reflect how much the results can be expected to generalize to other settings. 
        \item It is fine to include aspirational goals as motivation as long as it is clear that these goals are not attained by the paper. 
    \end{itemize}

\item {\bf Limitations}
    \item[] Question: Does the paper discuss the limitations of the work performed by the authors?
    \item[] Answer: \answerYes{} 
    \item[] Justification: We consider OT problem for unbalanced measures supported on a graph metric space.
    
    For Orlicz-EPT, we show that it is directly derived from a standard OT, with a guarantee on the nonnegativity for its ground cost via calibration, similar to OW approach (derived from corresponding standard OT with nonnegative ground cost~\citep{sturm2011generalized}). Therefore, we can bypass all challenges raised from unbalanced measures. We develop theoretical backgrounds to solve it by a binary search approach. However, Orlicz-EPT is still a two-level optimization problem, it has a high computational cost, illustrated in \S\ref{sec:experiments}.

    For OST, it is efficient in computation by simply solving a univariate optimization problem (\S\ref{sec:OST}). Although OST is a regularized approach, we theoretically show that OST generalize GST for unbalanced measures on a graph (Proposition~\ref{prop:relation_OST_GST}).

    \item[] Guidelines:
    \begin{itemize}
        \item The answer NA means that the paper has no limitation while the answer No means that the paper has limitations, but those are not discussed in the paper. 
        \item The authors are encouraged to create a separate "Limitations" section in their paper.
        \item The paper should point out any strong assumptions and how robust the results are to violations of these assumptions (e.g., independence assumptions, noiseless settings, model well-specification, asymptotic approximations only holding locally). The authors should reflect on how these assumptions might be violated in practice and what the implications would be.
        \item The authors should reflect on the scope of the claims made, e.g., if the approach was only tested on a few datasets or with a few runs. In general, empirical results often depend on implicit assumptions, which should be articulated.
        \item The authors should reflect on the factors that influence the performance of the approach. For example, a facial recognition algorithm may perform poorly when image resolution is low or images are taken in low lighting. Or a speech-to-text system might not be used reliably to provide closed captions for online lectures because it fails to handle technical jargon.
        \item The authors should discuss the computational efficiency of the proposed algorithms and how they scale with dataset size.
        \item If applicable, the authors should discuss possible limitations of their approach to address problems of privacy and fairness.
        \item While the authors might fear that complete honesty about limitations might be used by reviewers as grounds for rejection, a worse outcome might be that reviewers discover limitations that aren't acknowledged in the paper. The authors should use their best judgment and recognize that individual actions in favor of transparency play an important role in developing norms that preserve the integrity of the community. Reviewers will be specifically instructed to not penalize honesty concerning limitations.
    \end{itemize}

\item {\bf Theory assumptions and proofs}
    \item[] Question: For each theoretical result, does the paper provide the full set of assumptions and a complete (and correct) proof?
    \item[] Answer: \answerYes{} 
    \item[] Justification: Detailed proofs of theoretical results are placed in Appendix \S A.2. 
    \item[] Guidelines:
    \begin{itemize}
        \item The answer NA means that the paper does not include theoretical results. 
        \item All the theorems, formulas, and proofs in the paper should be numbered and cross-referenced.
        \item All assumptions should be clearly stated or referenced in the statement of any theorems.
        \item The proofs can either appear in the main paper or the supplemental material, but if they appear in the supplemental material, the authors are encouraged to provide a short proof sketch to provide intuition. 
        \item Inversely, any informal proof provided in the core of the paper should be complemented by formal proofs provided in appendix or supplemental material.
        \item Theorems and Lemmas that the proof relies upon should be properly referenced. 
    \end{itemize}

    \item {\bf Experimental result reproducibility}
    \item[] Question: Does the paper fully disclose all the information needed to reproduce the main experimental results of the paper to the extent that it affects the main claims and/or conclusions of the paper (regardless of whether the code and data are provided or not)?
    \item[] Answer: \answerYes{} 
    \item[] Justification: we provide detailed setup for our experiments in \S\ref{sec:experiments}. We discuss further details in Appendix B.2. We also submit the code together with our submission. 
    \item[] Guidelines:
    \begin{itemize}
        \item The answer NA means that the paper does not include experiments.
        \item If the paper includes experiments, a No answer to this question will not be perceived well by the reviewers: Making the paper reproducible is important, regardless of whether the code and data are provided or not.
        \item If the contribution is a dataset and/or model, the authors should describe the steps taken to make their results reproducible or verifiable. 
        \item Depending on the contribution, reproducibility can be accomplished in various ways. For example, if the contribution is a novel architecture, describing the architecture fully might suffice, or if the contribution is a specific model and empirical evaluation, it may be necessary to either make it possible for others to replicate the model with the same dataset, or provide access to the model. In general. releasing code and data is often one good way to accomplish this, but reproducibility can also be provided via detailed instructions for how to replicate the results, access to a hosted model (e.g., in the case of a large language model), releasing of a model checkpoint, or other means that are appropriate to the research performed.
        \item While NeurIPS does not require releasing code, the conference does require all submissions to provide some reasonable avenue for reproducibility, which may depend on the nature of the contribution. For example
        \begin{enumerate}
            \item If the contribution is primarily a new algorithm, the paper should make it clear how to reproduce that algorithm.
            \item If the contribution is primarily a new model architecture, the paper should describe the architecture clearly and fully.
            \item If the contribution is a new model (e.g., a large language model), then there should either be a way to access this model for reproducing the results or a way to reproduce the model (e.g., with an open-source dataset or instructions for how to construct the dataset).
            \item We recognize that reproducibility may be tricky in some cases, in which case authors are welcome to describe the particular way they provide for reproducibility. In the case of closed-source models, it may be that access to the model is limited in some way (e.g., to registered users), but it should be possible for other researchers to have some path to reproducing or verifying the results.
        \end{enumerate}
    \end{itemize}

\item {\bf Open access to data and code}
    \item[] Question: Does the paper provide open access to the data and code, with sufficient instructions to faithfully reproduce the main experimental results, as described in supplemental material?
    \item[] Answer: \answerYes{}
    \item[] Justification: Details of the experiments are given in \S\ref{sec:experiments}. We use public datasets for our experiments. Code is submitted together with the submission. We also discuss further details in Appendix B.2.
    \item[] Guidelines:
    \begin{itemize}
        \item The answer NA means that paper does not include experiments requiring code.
        \item Please see the NeurIPS code and data submission guidelines (\url{https://nips.cc/public/guides/CodeSubmissionPolicy}) for more details.
        \item While we encourage the release of code and data, we understand that this might not be possible, so “No” is an acceptable answer. Papers cannot be rejected simply for not including code, unless this is central to the contribution (e.g., for a new open-source benchmark).
        \item The instructions should contain the exact command and environment needed to run to reproduce the results. See the NeurIPS code and data submission guidelines (\url{https://nips.cc/public/guides/CodeSubmissionPolicy}) for more details.
        \item The authors should provide instructions on data access and preparation, including how to access the raw data, preprocessed data, intermediate data, and generated data, etc.
        \item The authors should provide scripts to reproduce all experimental results for the new proposed method and baselines. If only a subset of experiments are reproducible, they should state which ones are omitted from the script and why.
        \item At submission time, to preserve anonymity, the authors should release anonymized versions (if applicable).
        \item Providing as much information as possible in supplemental material (appended to the paper) is recommended, but including URLs to data and code is permitted.
    \end{itemize}

\item {\bf Experimental setting/details}
    \item[] Question: Does the paper specify all the training and test details (e.g., data splits, hyperparameters, how they were chosen, type of optimizer, etc.) necessary to understand the results?
    \item[] Answer: \answerYes{} 
    \item[] Justification: Please see \S\ref{sec:experiments}, Appendix B.2. 
    \item[] Guidelines:
    \begin{itemize}
        \item The answer NA means that the paper does not include experiments.
        \item The experimental setting should be presented in the core of the paper to a level of detail that is necessary to appreciate the results and make sense of them.
        \item The full details can be provided either with the code, in appendix, or as supplemental material.
    \end{itemize}

\item {\bf Experiment statistical significance}
    \item[] Question: Does the paper report error bars suitably and correctly defined or other appropriate information about the statistical significance of the experiments?
    \item[] Answer: \answerYes{}
    \item[] Justification: Please see reported empirical results. 
    \item[] Guidelines:
    \begin{itemize}
        \item The answer NA means that the paper does not include experiments.
        \item The authors should answer "Yes" if the results are accompanied by error bars, confidence intervals, or statistical significance tests, at least for the experiments that support the main claims of the paper.
        \item The factors of variability that the error bars are capturing should be clearly stated (for example, train/test split, initialization, random drawing of some parameter, or overall run with given experimental conditions).
        \item The method for calculating the error bars should be explained (closed form formula, call to a library function, bootstrap, etc.)
        \item The assumptions made should be given (e.g., Normally distributed errors).
        \item It should be clear whether the error bar is the standard deviation or the standard error of the mean.
        \item It is OK to report 1-sigma error bars, but one should state it. The authors should preferably report a 2-sigma error bar than state that they have a 96\% CI, if the hypothesis of Normality of errors is not verified.
        \item For asymmetric distributions, the authors should be careful not to show in tables or figures symmetric error bars that would yield results that are out of range (e.g. negative error rates).
        \item If error bars are reported in tables or plots, The authors should explain in the text how they were calculated and reference the corresponding figures or tables in the text.
    \end{itemize}

\item {\bf Experiments compute resources}
    \item[] Question: For each experiment, does the paper provide sufficient information on the computer resources (type of compute workers, memory, time of execution) needed to reproduce the experiments?
    \item[] Answer: \answerYes{}
    \item[] Justification: Please see Appendix \S B.2, where we describe that all experiments are carried on commodity hardware.
    \item[] Guidelines:
    \begin{itemize}
        \item The answer NA means that the paper does not include experiments.
        \item The paper should indicate the type of compute workers CPU or GPU, internal cluster, or cloud provider, including relevant memory and storage.
        \item The paper should provide the amount of compute required for each of the individual experimental runs as well as estimate the total compute. 
        \item The paper should disclose whether the full research project required more compute than the experiments reported in the paper (e.g., preliminary or failed experiments that didn't make it into the paper). 
    \end{itemize}
    
\item {\bf Code of ethics}
    \item[] Question: Does the research conducted in the paper conform, in every respect, with the NeurIPS Code of Ethics \url{https://neurips.cc/public/EthicsGuidelines}?
    \item[] Answer: \answerYes{}
    \item[] Justification: The research in our submission is to advance the machine learning field. To our knowledge, there is no foresee harmful consequences. 
    \item[] Guidelines:
    \begin{itemize}
        \item The answer NA means that the authors have not reviewed the NeurIPS Code of Ethics.
        \item If the authors answer No, they should explain the special circumstances that require a deviation from the Code of Ethics.
        \item The authors should make sure to preserve anonymity (e.g., if there is a special consideration due to laws or regulations in their jurisdiction).
    \end{itemize}

\item {\bf Broader impacts}
    \item[] Question: Does the paper discuss both potential positive societal impacts and negative societal impacts of the work performed?
    \item[] Answer: \answerYes{} 
    \item[] Justification: Please see Appendix \S B.2, where we describe that our research aims to advance to the machine learning field. The proposed OST has an efficient computation, which may help to reduce the computational cost. To our knowledge, there is no foresee negative social impacts for our research. 
    \item[] Guidelines:
    \begin{itemize}
        \item The answer NA means that there is no societal impact of the work performed.
        \item If the authors answer NA or No, they should explain why their work has no societal impact or why the paper does not address societal impact.
        \item Examples of negative societal impacts include potential malicious or unintended uses (e.g., disinformation, generating fake profiles, surveillance), fairness considerations (e.g., deployment of technologies that could make decisions that unfairly impact specific groups), privacy considerations, and security considerations.
        \item The conference expects that many papers will be foundational research and not tied to particular applications, let alone deployments. However, if there is a direct path to any negative applications, the authors should point it out. For example, it is legitimate to point out that an improvement in the quality of generative models could be used to generate deepfakes for disinformation. On the other hand, it is not needed to point out that a generic algorithm for optimizing neural networks could enable people to train models that generate Deepfakes faster.
        \item The authors should consider possible harms that could arise when the technology is being used as intended and functioning correctly, harms that could arise when the technology is being used as intended but gives incorrect results, and harms following from (intentional or unintentional) misuse of the technology.
        \item If there are negative societal impacts, the authors could also discuss possible mitigation strategies (e.g., gated release of models, providing defenses in addition to attacks, mechanisms for monitoring misuse, mechanisms to monitor how a system learns from feedback over time, improving the efficiency and accessibility of ML).
    \end{itemize}
    
\item {\bf Safeguards}
    \item[] Question: Does the paper describe safeguards that have been put in place for responsible release of data or models that have a high risk for misuse (e.g., pretrained language models, image generators, or scraped datasets)?
    \item[] Answer: \answerNA{}
    \item[] Justification: Our research poses no such risks.
    \item[] Guidelines:
    \begin{itemize}
        \item The answer NA means that the paper poses no such risks.
        \item Released models that have a high risk for misuse or dual-use should be released with necessary safeguards to allow for controlled use of the model, for example by requiring that users adhere to usage guidelines or restrictions to access the model or implementing safety filters. 
        \item Datasets that have been scraped from the Internet could pose safety risks. The authors should describe how they avoided releasing unsafe images.
        \item We recognize that providing effective safeguards is challenging, and many papers do not require this, but we encourage authors to take this into account and make a best faith effort.
    \end{itemize}

\item {\bf Licenses for existing assets}
    \item[] Question: Are the creators or original owners of assets (e.g., code, data, models), used in the paper, properly credited and are the license and terms of use explicitly mentioned and properly respected?
    \item[] Answer: \answerYes{} 
    \item[] Justification: Please see \S\ref{sec:experiments}, where we give proper credit/citation to the original owners.
    \item[] Guidelines:
    \begin{itemize}
        \item The answer NA means that the paper does not use existing assets.
        \item The authors should cite the original paper that produced the code package or dataset.
        \item The authors should state which version of the asset is used and, if possible, include a URL.
        \item The name of the license (e.g., CC-BY 4.0) should be included for each asset.
        \item For scraped data from a particular source (e.g., website), the copyright and terms of service of that source should be provided.
        \item If assets are released, the license, copyright information, and terms of use in the package should be provided. For popular datasets, \url{paperswithcode.com/datasets} has curated licenses for some datasets. Their licensing guide can help determine the license of a dataset.
        \item For existing datasets that are re-packaged, both the original license and the license of the derived asset (if it has changed) should be provided.
        \item If this information is not available online, the authors are encouraged to reach out to the asset's creators.
    \end{itemize}

\item {\bf New assets}
    \item[] Question: Are new assets introduced in the paper well documented and is the documentation provided alongside the assets?
    \item[] Answer: \answerNA{} 
    \item[] Justification: NA
    \item[] Guidelines:
    \begin{itemize}
        \item The answer NA means that the paper does not release new assets.
        \item Researchers should communicate the details of the dataset/code/model as part of their submissions via structured templates. This includes details about training, license, limitations, etc. 
        \item The paper should discuss whether and how consent was obtained from people whose asset is used.
        \item At submission time, remember to anonymize your assets (if applicable). You can either create an anonymized URL or include an anonymized zip file.
    \end{itemize}

\item {\bf Crowdsourcing and research with human subjects}
    \item[] Question: For crowdsourcing experiments and research with human subjects, does the paper include the full text of instructions given to participants and screenshots, if applicable, as well as details about compensation (if any)? 
    \item[] Answer: \answerNA{}
    \item[] Justification: NA 
    \item[] Guidelines:
    \begin{itemize}
        \item The answer NA means that the paper does not involve crowdsourcing nor research with human subjects.
        \item Including this information in the supplemental material is fine, but if the main contribution of the paper involves human subjects, then as much detail as possible should be included in the main paper. 
        \item According to the NeurIPS Code of Ethics, workers involved in data collection, curation, or other labor should be paid at least the minimum wage in the country of the data collector. 
    \end{itemize}

\item {\bf Institutional review board (IRB) approvals or equivalent for research with human subjects}
    \item[] Question: Does the paper describe potential risks incurred by study participants, whether such risks were disclosed to the subjects, and whether Institutional Review Board (IRB) approvals (or an equivalent approval/review based on the requirements of your country or institution) were obtained?
    \item[] Answer: \answerNA{} 
    \item[] Justification: NA 
    \item[] Guidelines:
    \begin{itemize}
        \item The answer NA means that the paper does not involve crowdsourcing nor research with human subjects.
        \item Depending on the country in which research is conducted, IRB approval (or equivalent) may be required for any human subjects research. If you obtained IRB approval, you should clearly state this in the paper. 
        \item We recognize that the procedures for this may vary significantly between institutions and locations, and we expect authors to adhere to the NeurIPS Code of Ethics and the guidelines for their institution. 
        \item For initial submissions, do not include any information that would break anonymity (if applicable), such as the institution conducting the review.
    \end{itemize}

\item {\bf Declaration of LLM usage}
    \item[] Question: Does the paper describe the usage of LLMs if it is an important, original, or non-standard component of the core methods in this research? Note that if the LLM is used only for writing, editing, or formatting purposes and does not impact the core methodology, scientific rigorousness, or originality of the research, declaration is not required.
    \item[] Answer: \answerNA{} 
    \item[] Justification: NA 
    \item[] Guidelines:
    \begin{itemize}
        \item The answer NA means that the core method development in this research does not involve LLMs as any important, original, or non-standard components.
        \item Please refer to our LLM policy (\url{https://neurips.cc/Conferences/2025/LLM}) for what should or should not be described.
    \end{itemize}

\end{enumerate}

\end{document}